\documentclass{article}

    \usepackage[final, dandb]{neurips_2025}

\usepackage[utf8]{inputenc} 
\usepackage[T1]{fontenc}    
\usepackage{hyperref}       
\usepackage{url}            
\usepackage{booktabs}       
\usepackage{amsfonts}       
\usepackage{nicefrac}       
\usepackage{xcolor}         
\usepackage{multirow}
\usepackage{array}
\usepackage{balance}
\usepackage{amssymb}
\usepackage{verbatim}
\usepackage{booktabs}
\usepackage{bbding}
\usepackage{pifont}
\usepackage{makecell}
\usepackage{soul, color}
\usepackage{setspace}
\usepackage{colortbl}
\usepackage{amsmath}
\usepackage{graphicx}
\usepackage{stfloats}
\usepackage{diagbox}
\usepackage{subcaption}
\usepackage{wrapfig}
\usepackage{titletoc}
\usepackage{dirtree}
\usepackage{listings}
\usepackage{amsmath,amssymb,amsfonts}
\usepackage{algorithmic}
\usepackage[ruled,vlined]{algorithm2e}
\usepackage[final,nopatch=footnote]{microtype}

\usepackage{tcolorbox}
\tcbuselibrary{skins,xparse,breakable}
\newtcolorbox{promptbox}{
    colback=gray!10,    
    colframe=gray!80,   
    boxrule=0.5pt,      
    arc=3pt,            
    left=6pt,           
    right=6pt,          
    top=6pt,            
    bottom=6pt,         
    boxsep=0pt,         
    breakable
}
\newtcolorbox{promptbox2}{
    colback=orange!10,   
    colframe=orange!80,   
    boxrule=0.5pt,      
    arc=3pt,            
    left=6pt,           
    right=6pt,          
    top=6pt,            
    bottom=6pt,         
    boxsep=0pt,         
    breakable
}

\definecolor{lightgoldenrodyellow}{rgb}{0.98, 0.98, 0.82}
\definecolor{lightcornflowerblue}{rgb}{0.6, 0.81, 0.93}
\definecolor{lightgreen}{rgb}{0.56, 0.93, 0.56}
\definecolor{lightcoral}{rgb}{0.94, 0.5, 0.5}
\definecolor{cadmiumorange}{rgb}{0.93, 0.53, 0.18}

\newcommand{\bq}[1]{\textcolor{purple}{(bq:#1)}}
\newcommand{\zm}[1]{\textcolor{blue}{(zm:#1)}}

\newcommand{\supp}[1]{{{#1Supplementary Material}}}
\newcommand{\redbox}[1]{{\fboxrule=0.4pt\fboxsep=1pt\color{red}\fbox{\color{black}#1}}}

\newcommand{\hlgray}[1]{{\setlength{\fboxsep}{1.5pt}\colorbox{gray!50}{#1}}} 
\newcommand{\hllightgray}[1]{{\setlength{\fboxsep}{1.5pt}\colorbox{gray!20}{#1}}} 
\newcommand{\hldarkorange}[1]{{\setlength{\fboxsep}{1.5pt}\colorbox{cadmiumorange!70}{#1}}} 
\newcommand{\hlorange}[1]{{\setlength{\fboxsep}{1.5pt}\colorbox{cadmiumorange!40}{#1}}} 
\newcommand{\hllightorange}[1]{{\setlength{\fboxsep}{1.5pt}\colorbox{cadmiumorange!20}{#1}}}

\title{MineAnyBuild: Benchmarking Spatial Planning\\ for Open-world AI Agents}
\author{%
  Ziming Wei\textsuperscript{1}\thanks{Equal Contribution. $^{\dagger}$Corresponding Author.}\ \ ,~~Bingqian Lin\textsuperscript{2}$^{*}$,~~Zijian Jiao\textsuperscript{1}$^{*}$,~~Yunshuang Nie\textsuperscript{1},~~Liang Ma\textsuperscript{3}\\\textbf{Yuecheng Liu\textsuperscript{4},~~Yuzheng Zhuang\textsuperscript{4},~~Xiaodan Liang\textsuperscript{1}$^{\dagger}$}\vspace{0.1cm}\\
  \textsuperscript{1}Shenzhen Campus of Sun Yat-sen University\quad\textsuperscript{2}Shanghai Jiao Tong University\\
  \textsuperscript{3}Mohamed bin Zayed University of Artificial Intelligence\quad\textsuperscript{4}Huawei Noah’s Ark Lab\vspace{0.15cm}\\
Project Website: \url{https://mineanybuild.github.io/}}

\begin{document}

\maketitle

\begin{abstract}
Spatial Planning is a crucial part in the field of spatial intelligence, which requires the understanding and planning about object arrangements in space perspective. AI agents with the spatial planning ability can better adapt to various real-world applications, including robotic manipulation, automatic assembly, urban planning
{\it etc}.  Recent works have attempted to construct benchmarks for evaluating the spatial intelligence of Multimodal Large Language Models (MLLMs). Nevertheless, these benchmarks primarily focus on spatial reasoning based on typical Visual Question-Answering (VQA) forms, which suffers from the gap between abstract spatial understanding and concrete task execution. In this work, we take a step further to build a comprehensive benchmark called \textbf{MineAnyBuild}, aiming to evaluate the spatial planning ability of open-world AI agents in the \textit{Minecraft} game. Specifically, MineAnyBuild requires an
agent to generate {\it executable architecture building plans} based on the given multi-modal human instructions. It involves 4,000 curated 
tasks and
provides a paradigm for infinitely expandable data collection by utilizing rich player-generated content.
MineAnyBuild evaluates spatial planning through four core supporting dimensions:
spatial understanding, spatial reasoning, creativity, and spatial commonsense. Based on MineAnyBuild, we perform a comprehensive evaluation for existing MLLM-based agents, revealing the severe limitations but enormous potential in their spatial planning abilities. We believe our MineAnyBuild
will open new avenues for the evaluation of spatial intelligence and help promote further development for open-world AI agents capable of spatial planning. 

\end{abstract}

\section{Introduction}
\label{intro}
Spatial intelligence, an emerging research field gradually attracting the attention of AI researchers, requires AI agents to understand, reason and memorize the visual-spatial relationships between objects and spaces~\cite{song2024robospatial,ray2024sat,li2024seeground,zhang2024spartun3d}. 
Spatial planning is a pivotal capability regarding spatial intelligence, which requires agents to not only perform spatial perception and cognition, but also generate executable planning in 3D space. Spatial planning is widely needed in various human-centric real-world applications,  including automatic assembly, architectural design, environmental urban planning, {\it etc}.

AI Agents integrated with  Multi-modal Large Language Models (MLLMs) have demonstrated astonishing capabilities in various tasks in text (1D) and image (2D) domains~\cite{liu2023visual,openai2023gpt4v-system,team2023gemini}. 
To investigate how existing MLLM-based agents can handle space dimension tasks, several benchmarks designed for evaluating the spatial intelligence have been proposed recently~\cite{yang2024thinking,chen2024spatialvlm,guo2024drivemllm,tang2025lego}.  These benchmarks reveal that although AI agents perform well in tasks of text and image domains, they still present relatively poor performance in spatial dimension tasks. However, current benchmarks have critical constraints in evaluating spatial intelligence.
They mainly focus on metric-level spatial understanding tasks and predominantly employ Visual Question-Answering (VQA) pairs, requiring AI agents to answer geometric attributes (e.g., distance, positional coordinates, or spatial relations of objects in 3D space), while neglecting the gap between abstract spatial understanding and concrete task execution.

In this work, we propose \textbf{MineAnyBuild}, which is an innovative benchmark designed to evaluate an important yet unexplored aspect of spatial intelligence, i.e.,  spatial planning, for open-world AI agents. In our MineAnyBuild, the agents need to generate {\it executable spatial plans} following human instructions for constructing a building or indoor decoration, which requires both spatial reasoning and task execution.
We build our benchmark on the popular \textit{Minecraft} game, where a player journeys through a 3D world with diverse biomes to explore, tools to craft, and architectures to build. Compared to benchmarks focusing on skills learning or tech-tree tasks~\cite{fan2022minedojo, wang2023voyager}, 
architecture building
has always been a vital attraction to millions of players to present the openness and freedom of Minecraft. Unlike most other games, Minecraft defines go-as-you-please goals, making it well suited for developing open-ended tasks for AI agent research. 

\begin{figure}[t]
  \centering
\includegraphics[width=\textwidth]{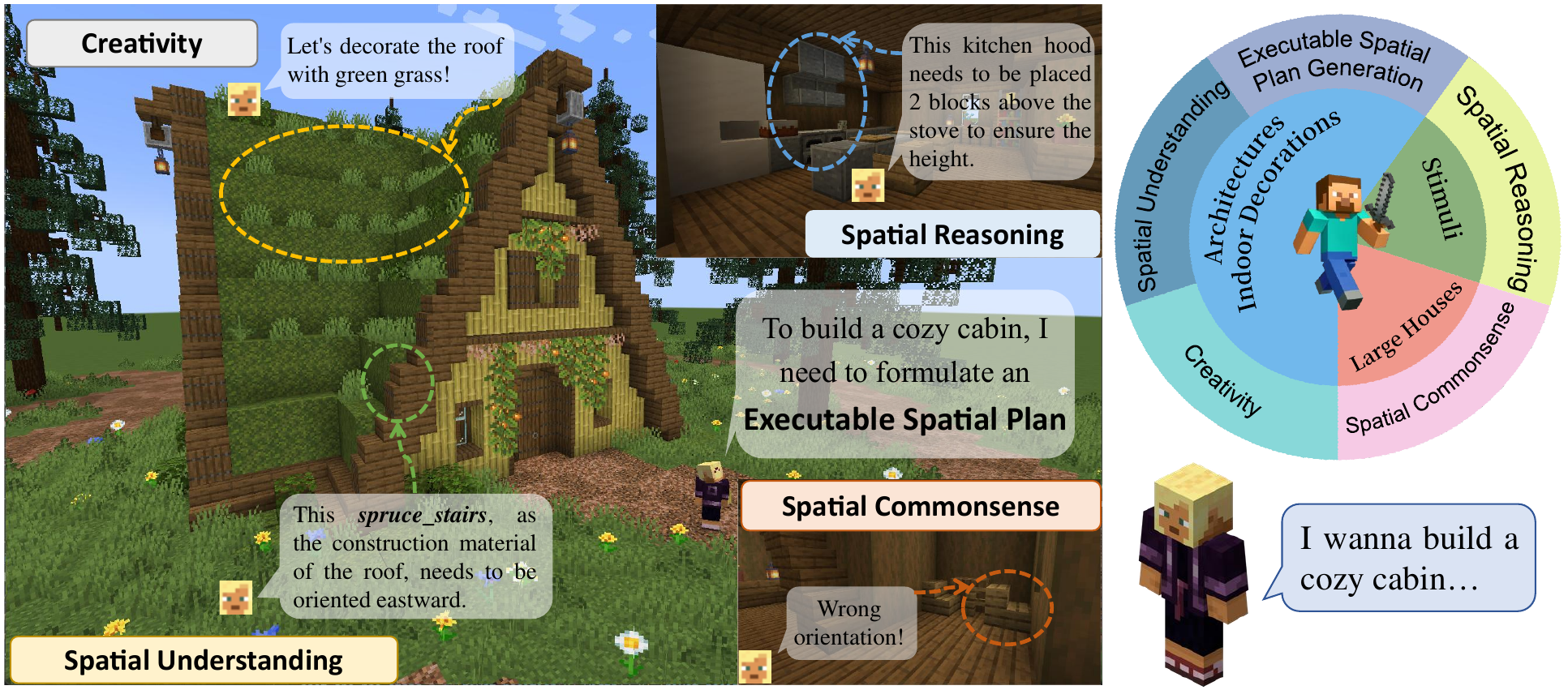}
  \caption{Overview of \textbf{MineAnyBuild}. Our MineAnyBuild is a novel benchmark built on the Minecraft game, which aims to evaluate the spatial planning capabilities of open-world AI agents. In MineAnyBuild, the agent needs to generate {\it executable spatial plans} to construct a building or indoor decoration following given multi-modal human instructions. We introduce four core  dimensions, including spatial understanding, creativity, spatial reasoning, and spatial commonsense to fulfill a comprehensive assessment for spatial planning.}
  \label{fig:overview}
  \vspace{-0.3cm}
\end{figure}

Our MineAnyBuild benchmark consists of 4,000 curated tasks where four core evaluation dimensions are introduced. As shown in Figure~\ref{fig:overview}, given a multimodal human instruction, agents are requested to perform \textbf{spatial understanding} to abstract a pivotal basic structure according to specific or brief demands, where agents emulate architects in our real world, and plan the composition of each basic units. 
Agents also need to reason and think about whether the units of the architecture from different perspectives conforms to spatial rules by \textbf{spatial reasoning}. For the 
overall appearance of the architecture, agents exert their \textbf{creativity} and imagination to make it more aesthetically unique, or to simulate some well-designed or delicate structural designs in the real world through the combination of fixed-shaped blocks, e.g., using a variety of stairs and slabs to design unique Chinese-style or castle-style roofs. For some architectures like modern houses, agents implement \textbf{spatial commonsense} to judge the rationality of each designs inside the buildings.

We design and construct different tasks based on multiple aspects that a human player would consider, to evaluate several capabilities of AI agents. Agents are supposed to response with concrete layout sequences of expected architectures to present their spatial planning. For some tasks that are not easy to evaluate directly like spatial reasoning, we customize tasks inspired by classical mental rotation experiments~\cite{shepard1971mental,shepard1988mental}
to test agents. For creativity, we score and vote on the overall aesthetics and structural strategy by human evaluation or critique-based MLLMs. We also propose an infinitely expandable paradigm to utilize Minecraft data on the Internet, where millions of active players provide their creation and shares, to build our tasks. Through our data curation pipeline, we can collect endless tasks evaluating spatial intelligence for open-world agents, making further contributions to promoting AI agents research.

We test the tasks on several state-of-the-art MLLM-based AI agents and observe that even the most powerful MLLMs like 
GPT-4o and Claude-3.7-Sonnet
demonstrate significant limitations in most tasks, where GPT-4o obtains an overall score of 41.02, far lower than the maximum score of 100. 
Open-source models generally have poor capabilities to generate executable spatial plan, reflecting a serious deficiency in their understanding of spatial data.
These results reveal the foresight of our MineAnyBuild for AI evaluation.

To summarize, the main contributions of this work are as follows:
\begin{itemize}

\item We propose MineAnyBuild, which benchmarks the spatial planning evaluation for open-world AI agents in the Minecraft game. MineAnyBuild covers diverse evaluation dimensions, including spatial reasoning, creativity, spatial commonsense, {\it etc}. Through requiring the agent to generate executable architecture building plans, our MineAnyBuild significantly mitigate the gap between abstract spatial understanding and concrete task execution.    

\item We test various existing MLLM-based AI agents for spatial planning in multiple perspectives and difficulties, which exposes the insufficiency of the existing AI agents' capabilities in spatial planning.
We provide the visualization results on executable planning outputs and failure cases, revealing that current AI agents are still facing tough issues such as spatial misunderstanding and implementation gap to be handled.

\item We propose an infinitely expandable data curation pipeline to scale our benchmark and datasets, 
where we can collect endless player-generated content on the Internet and automatically convert it into processable data. Our pipeline well utilize the abundant creations made by millions of players to benefit the training and evaluation of open-world AI agents. 


\end{itemize}

\section{Benchmark and Task Suite}
\label{bench}
In this section, we describe our MineAnyBuild benchmark in detail. Specifically, we first present the overview of our benchmark in Section~\ref{Benchmark Overview}. Then, we define various spatial planning tasks in MineAnyBuild in Section~\ref{Tasks}. Finally, we introduce our data curation pipeline in Section~\ref{Data Curation}.


\subsection{Benchmark Overview}
\label{Benchmark Overview}
MineAnyBuild is designed
to evaluate an AI agent's capabilities in spatial planning to conduct infinite architecture creations in Minecraft game. 
Spatial planning is a critical capability, aiming to examine agents' understanding and disassembly of combinations in 3D space, and to construct each sub-units and judge the rationality of them by reasoning or commonsense. Our benchmark examines various MLLM-based agents to conduct planning on architectures, which requires them to generate executable architectural construction plans according to different forms of instructions or visual inputs.
Evaluating the creativity of agents becomes essential in our particular architectural construction tasks, as it reflects human-centric assessments of aesthetic value across spatial planning and conception designing domains.
For some evaluating dimensions that are not easily presented directly in spatial planning, such as spatial reasoning and spatial commonsense, we design series of visual question-answering pairs to indirectly reflect the manifestation of these two capabilities of agents.
The next sections detail the specific tasks and the process of our data curation.

\subsection{Tasks}
\label{Tasks}
\begin{figure*}[t]
\vspace{-0.2cm}
\centering
\includegraphics[width=0.95\linewidth]{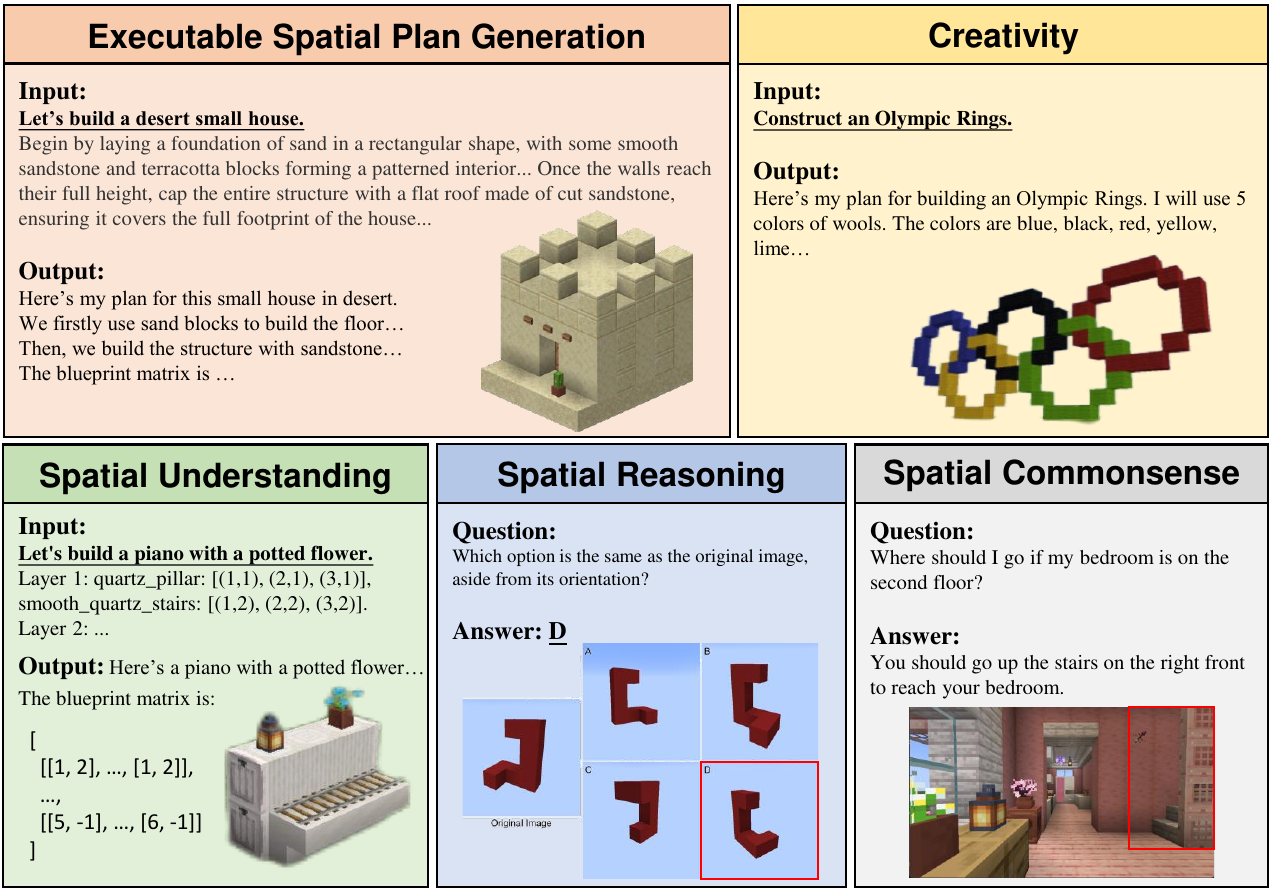}
\caption{Task examples of \textbf{MineAnyBuild}. We present five task examples with specific inputs (questions) and outputs (answers). Some of 
them are simplified to illustrate the core presentations.}
\label{fig:task example}
\vspace{-0.6cm}
\end{figure*}

Our MineAnyBuild involves approximately  4,000 spatial planning tasks with 500+ buildings/indoor decoration assets.
These tasks, including Executable Spatial Plan Generation, Spatial Understanding, Creativity, Spatial Reasoning, and Spatial Commonsense, correspond to diverse evaluation dimensions, thereby conducting a comprehensive assessment of AI agents' spatial planning capabilities.
In Executable Spatial Plan Generation, Spatial Understanding, and Creativity tasks, the agent needs to generate executable spatial plans for building an architecture according to the given instruction. While in Spatial Reasoning and Spatial Commonsense tasks, 
we introduce $\sim$2,000 VQA pairs, where we ask the agent to answer the given questions accompanied by the related images.
In the following, we define each task in detail, and present the corresponding task examples in Figure~\ref{fig:task example}. 
\noindent\textbf{Executable Spatial Plan Generation.}
To construct an architecture, an agent first needs to design the layout of the architecture and accordingly generate the executable spatial plans based on its spatial perception, spatial understanding, and abundant knowledge. Based on this motivation, we propose the Executable Spatial Plan Generation task, 
which evaluates agents' abilities to perform
Spatial Planning.
The task input is an abstract architecture building instruction accompanied by precise explanations. Under the given task input, the agents are required to think on the decomposition of architecture substructures and corresponding connections to generate executable spatial plans for architecture building, just like completing a jigsaw puzzle.


For example, in this task instruction ``\textit{Build an apple...The apple also needs to have a stem, which we can use black\_terracotta to make it.}'' for architecture building, we lead the agent to think on the \textit{stem} substructure in \textit{apple} architecture and how to connect it with other substructures. If the agent could understand and plan in spatial perspective, the result should be better than planning in an abstract perspective.
For the instruction regarding the indoor decoration, the agents are challenged to make more delicate and exquisite design and planning.
More details of the Executable Spatial Plan Generation task are provided in the \supp{}.

\noindent\textbf{Spatial Understanding.} 
Inspired by the popular instruction following tasks~\cite{zeng2023evaluating,zhou2023instruction,lou2024large} which are widely developed for evaluating MLLM-based agents, we introduce a Spatial Understanding task, where the agent needs to build the architecture according to the step-by-step instruction containing the positions of each building block through abstract spatial understanding. Specifically, we label the parts of our data with ground-truth annotations and generate the instructions with a mapping table of relative coordinate corresponding to the pivot position, for instance, \textit{Layer 2: "red\_wool": [(0,0),(1,0)]...}, where the block types and relative positions are provided. The agents are required to translate it into 
a blueprint matrix, 
which reflects the cognitive transition and integration of relative and holistic spatial understanding, simulating human-like cognitive mapping mechanisms that dynamically 
balance egocentric (body-centered) and allocentric (world-centered) perspectives.


\noindent\textbf{Creativity.}
Architecture constructing and indoor decoration designing are attracting evaluation tasks than the previous tasks~\cite{fan2022minedojo,hu20243d,barthet2022open,chen2024apt,earle2024dreamcraft}, like textual reasoning and coding.
In our MineAnyBuild, we introduce a novel Creativity task for evaluating architecture building, where
agents receive an instruction and are required to brainstorm block combinations for different parts of the architecture and outline a rough structure layout, to find ways to maximize creativity and the dynamic range of possible builds.


It is worth elaborating that although this evaluation dimension is different from other traditional ones, creativity is an aesthetic and humanistic criterion that is more in line with human intuition and consensus, akin to the aesthetic assessment of image generation tasks.
Creativity is a crucial component of
scientific thinking that reflects the process of new things replacing old ones, which is precisely the spirit that scientific researchers have always pursued. Therefore, creativity 
not only reflects the cognitive depth of future AI systems, but also emerges as a novel and important criterion for agents towards Artificial General Intelligence (AGI). 
We test creativity through state-of-the-art MLLM-based critic models and human evaluations.
Despite the challenges in standardizing evaluation criteria and achieving inter-rater reliability, 
the majority of recognized favorable comments or votes indicate that agents exhibit measurable creativity.

\noindent\textbf{Spatial Reasoning.}
Spatial reasoning is the ability to imagine, visualize and differentiate objects in 3D space~\cite{liu2023visual,byrne1989spatial,clements1992geometry}. 
Inspired by the
classic experiments in psychology named \textit{mental rotation}~\cite{hegarty2010components, shepard1988mental, vandenberg1978mental}, 
we construct 48 geometric objects made of blocks, denoted as \textit{stimuli}, and generated 1,900 tasks for evaluation of spatial reasoning.
As shown in Figure~\ref{fig:task example}, for example, the agent needs to analyze the geometric structure of the given stimuli and determine whether others are the same or not as the given one.
We conduct these tasks with Visual Question-Answer pairs, which is easier and more accurate for evaluation.

\noindent\textbf{Spatial Commonsense.}
Spatial commonsense is the critical intuitive comprehension in our daily life that humans possess about the spatial attributes of objects in the physical world, including location, orientation, distance, shape, {\it etc.}~\cite{davis2015commonsense,liu2022things,collell2018acquiring}.
Spatial commonsense is generally reflected in several aspects: 1) navigation and sense of direction: human can roughly orient the bedroom without a map. 2) rationality of object placement: a refrigerator cannot be placed in a bathroom. 
We evaluate these tasks on agents and 
we place the complete commonsense tasks tested in the \supp{}.

\subsection{Data Curation}
\label{Data Curation}

\begin{figure*}[t]
\vspace{-0.2cm}
\centering
\includegraphics[width=\linewidth]{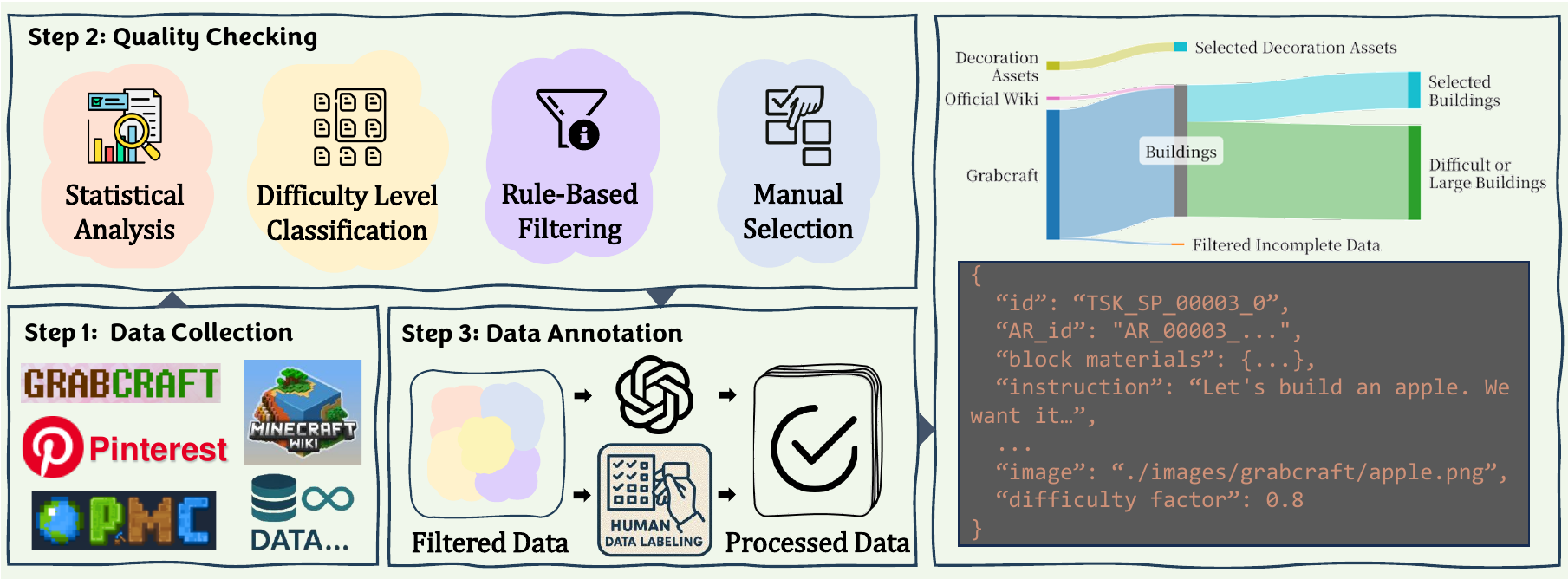}
\caption{Data curation pipeline of \textbf{MineAnyBuild}. We conduct three core steps to curate our datasets: data collection, quality checking, and data annotation. On the right side, a Sankey diagram showing our data processing flow is presented, along with an example of simplified format of processed data.}
\label{fig:data_pipe}
\vspace{-0.3cm}
\end{figure*}

MineAnyBuild is a comprehensive benchmark with diverse architectures and indoor decorations, aligned with various instructions and visual reference images. We build our benchmark based on the following steps: 1) data collection, 
2) quality checking, and 3) data annotation. 
Figure~\ref{fig:data_pipe} presents our data curation pipeline for constructing MineAnyBuild.

\noindent\textbf{Data Collection.}
Benefiting from
the abundant and creative player-generated content on the Internet, we first collect $\sim$7000 architectures from several websites, e.g., GrabCraft~\cite{grabcraft} and Minecraft Official Wiki~\cite{official_wiki,fandom_wiki}, and collect $\sim$500 indoor decoration assets from sharing platforms by Minecraft creators.
For some player-uploaded data containing potential issues, we filter out these problematic data with some quality standards. 
For spatial reasoning tasks, we collect 48 stimuli referencing the classic mental rotation experiments~\cite{shepard1971mental} and generate three groups of chiral
stimuli symmetrical about the X/Y/Z coordinate axes.
We design three major types of questions to construct VQA data for the spatial reasoning task based on these generated stimuli.
These questions involve having agents select the only one among the four options that differs from or is the same as the stimulus in reference image, or to determine whether the stimuli in the two images are consistent.
We utilize the data of large-scale buildings incorporating interior decorations to generate VQA pairs that complies with spatial commonsense, and further request agents to plan how to construct these decoration assets.


\noindent\textbf{Quality Checking and Data Annotation.}
We implement some codes to filter the problematic data, and
then we conduct a human review process to maintain high quality for data annotation. 
We annotate the instructions  of tasks by human or state-of-the-art MLLMs. Specifically, we
first carefully design some instructions that guide the agents to think about the decomposition and construction of the architectures, thereby more closely aligning with the motivation of spatial planning.
For spatial commonsense tasks, we manually design the VQA pairs that well-fit with questions in the real world.



\noindent\textbf{Infinitely Expandable Paradigm.}
As shown in Figure \ref{fig:data_pipe}, we provide an infinitely expandable paradigm for data curation, facilitating the subsequent development of training and evaluation 
resources to 
advance AI agents research for spatial planning.
Through our infinitely expandable paradigm, we can collect the majority of the existing player-generated content on the Internet and import it into the Minecraft game. Specifically,
we manually mark the starting block (the minimum values on the X/Y/Z coordinates) and the ending block (the maximum values on the X/Y/Z coordinates) of the 3D coordinates as the three-dimensional coordinate box of the entire building, and obtain all the block information corresponding to each position through \textit{mineflayer} simulator~\cite{mineflayer}.
After filtering the \textit{``air''} blocks, 
corresponding \textit{three\_d\_info}, \textit{blueprint} and \textit{block\_materials} can be acquired, with which
we can generate this building by calling high-level commands in a blank Minecraft environment and obtain the corresponding visual images through manual screenshot for MLLM or manual annotation.
The data that finally generate follows the requirements of our datasheet in terms of format, ultimately ensuring that all data has its unified format.

\section{Experiments}
\label{exp}
In this section, we describe the agents evaluated on MineAnyBuild and corresponding evaluation metrics, followed by the results and analysis of performance of agents on MineAnyBuild. 

\subsection{Agents}
We mainly conduct our evaluation on MLLM-based agents that suitable to address the spatial planning task in our benchmark.
To adapt MLLM-based agents to our spatial planning task, we ask the agents to directly output the executable blueprint matrices. Then, 
the matrices are subsequently utilized by \textit{mineflayer} simulator~\cite{mineflayer} to automatically generate corresponding architectures in Minecraft environment.
We evaluate 13 MLLMs for our MineAnyBuild.
For proprietary models, we evaluate Claude-3.5-Sonnet, Claude-3.7-Sonnet~\cite{claude}, Gemini-1.5-Flash, Gemini-1.5-Pro, Gemini-2.0-Flash~\cite{team2023gemini}, GPT-4o, GPT-4o-mini~\cite{gpt4o}.
For open-source models, we evaluate InternVL2.5-[2B/4B/8B]~\cite{chen2024expanding}, Qwen2.5VL-[3B/7B]~\cite{Qwen2.5-VL}, LLava-Onevision-7B~\cite{li2024llava}.

All evaluations are conducted in a zero-shot manner for a fair comparison.
We also provide RL-based agents for future research and adaptation. We place the detailed information of them
and compute resources for agents in the \supp{}.

\begin{table*}[t]
	\fontsize{12}{12}\selectfont
 \centering
\caption{Evaluation results of AI agents on \textbf{MineAnyBuild}. \hlgray{Gray} indicates the best performance of each evaluation dimension among all agents and \hllightgray{Light Gray} indicates the second best results. We also highlight the top three agents based on their overall performance with \hldarkorange{Dark Orange}, \hlorange{Orange}, \hllightorange{Light Orange}, respectively.}
	\label{tab1_metrics}
	\resizebox{\linewidth}{!}{
	{\renewcommand{\arraystretch}{1.3}
		\begin{tabular}{l||ccccc|c}

			\specialrule{.1em}{.05em}{.05em}
    \multirow{2}{*}{Models}&\cellcolor{lightcornflowerblue!40}Executable Spatial Plan Generation&\cellcolor{lightgreen!40}Spatial Understanding&\cellcolor{lightcornflowerblue!40}Spatial Reasoning&\cellcolor{lightgreen!40}Creativity&\cellcolor{lightcornflowerblue!40}Spatial Commonsense
    &\multirow{2}{*}{Overall}
    \cr\cline{2-6}
    &Score $\uparrow$&Score $\uparrow$&Accuracy $\uparrow$&Score $\uparrow$&Score $\uparrow$&\cr\hline
		\rowcolor{lightgoldenrodyellow}\multicolumn{7}{l}{\textbf{\textit{Proprietary}}}\\

Claude-3.5-Sonnet&3.21&4.63&19.8&\cellcolor{gray!50}3.24&6.90&39.92\\
Claude-3.7-Sonnet&\cellcolor{gray!20}3.48&\cellcolor{gray!50}5.07&17.6&\cellcolor{gray!20}3.10&6.94&\cellcolor{cadmiumorange!40}40.70\\
Gemini-1.5-Flash&2.87&2.49&\cellcolor{gray!20}25.8&2.71&7.12&35.54\\
Gemini-1.5-Pro&\cellcolor{gray!50}3.53&\cellcolor{gray!20}4.80&16.9&2.73&\cellcolor{gray!50}7.52&\cellcolor{cadmiumorange!20}40.54\\
Gemini-2.0-Flash&2.63&4.19&16.0&2.44&6.82&35.36\\
GPT-4o&3.27&4.75&24.4&2.73&\cellcolor{gray!20}7.32&\cellcolor{cadmiumorange!70}41.02\\
GPT-4o-mini&2.08&2.52&\cellcolor{gray!50}26.7&2.38&7.14&33.58\\

\hline
\rowcolor{lightgoldenrodyellow}\multicolumn{7}{l}{\textbf{\textit{Open-source}}}\\
InternVL2.5-2B&0.24&0.34&19.8&0.28&4.94&15.56\\
InternVL2.5-4B&0.32&0.42&20.0&0.63&5.66&18.06\\
InternVL2.5-8B&0.68&0.62&20.4&0.66&5.62&19.24\\
Qwen2.5VL-3B&0.63&0.61&17.0&0.54&5.46&17.88\\
Qwen2.5VL-7B&1.29&1.12&16.0&1.34&6.30&23.30\\
LLava-Onevision-7B&0.73&0.92&19.6&0.98&5.54&20.26\\

        

 \specialrule{.1em}{.05em}{.05em}

			\end{tabular}}}
	\vspace{-0.4cm}
\end{table*}

\subsection{Evaluation Metrics}
We introduce diverse metrics for evaluating different tasks in our benchmark.
For the Executable Spatial Plan Generation, Creativity, and Spatial Commonsense tasks, as their results do not have a definitely correct or perfect answer, we use the state-of-the-art MLLM (GPT-4.1~\cite{gpt41}) as the critic model to score the planning. Specifically, 
we query GPT-4.1 to score separately based on different evaluation sub-dimensions and calculate a weighted ``Evaluation Score''.
For different tasks, we obtain a comprehensive score based on the score, 
denoted as \textbf{``Score'' (out of 10)} shown in Table \ref{tab1_metrics}, through corresponding weighting to indicate the performance of agents in each task. 
For some cases where the plans generated by agents are not executable, we directly set the scores of these cases to 0, showing that agents have failed in these cases.
For the Spatial Reasoning task, we directly calculate the \textbf{Accuracy(\%)} of the agent's responses as our results.
More details about the evaluation metrics and 
the weighted formulas of scores
are given in the \supp{}.

\subsection{Results Analysis}
\begin{figure*}[t]
\vspace{-0.2cm}
\centering

\includegraphics[width=0.9\linewidth]{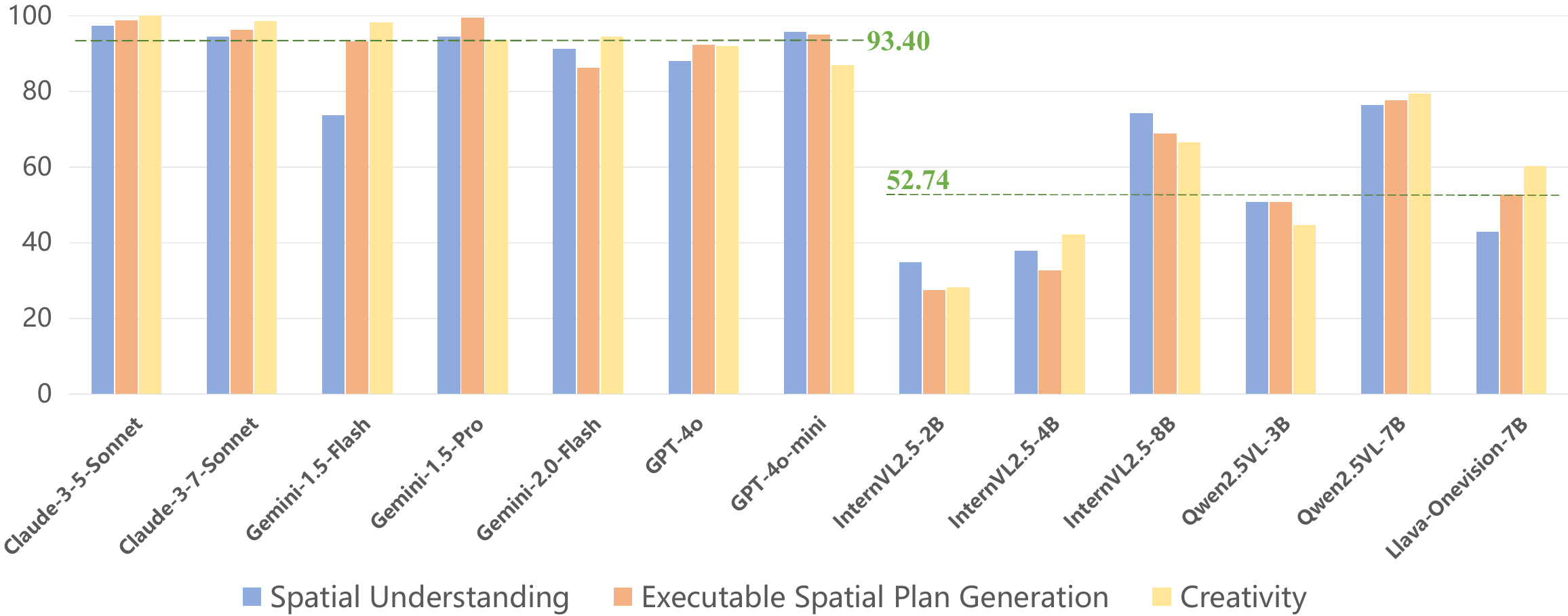}
\caption{Bar chart of Output Success Rate (OSR) for MLLMs. Two green dotted lines indicate the average OSR of proprietary models and open-source models, respectively.}
\label{fig:bar}
\vspace{-0.3cm}
\end{figure*}

\begin{figure*}[t]
\centering
\vspace{-0.2cm}
\includegraphics[width=\textwidth]{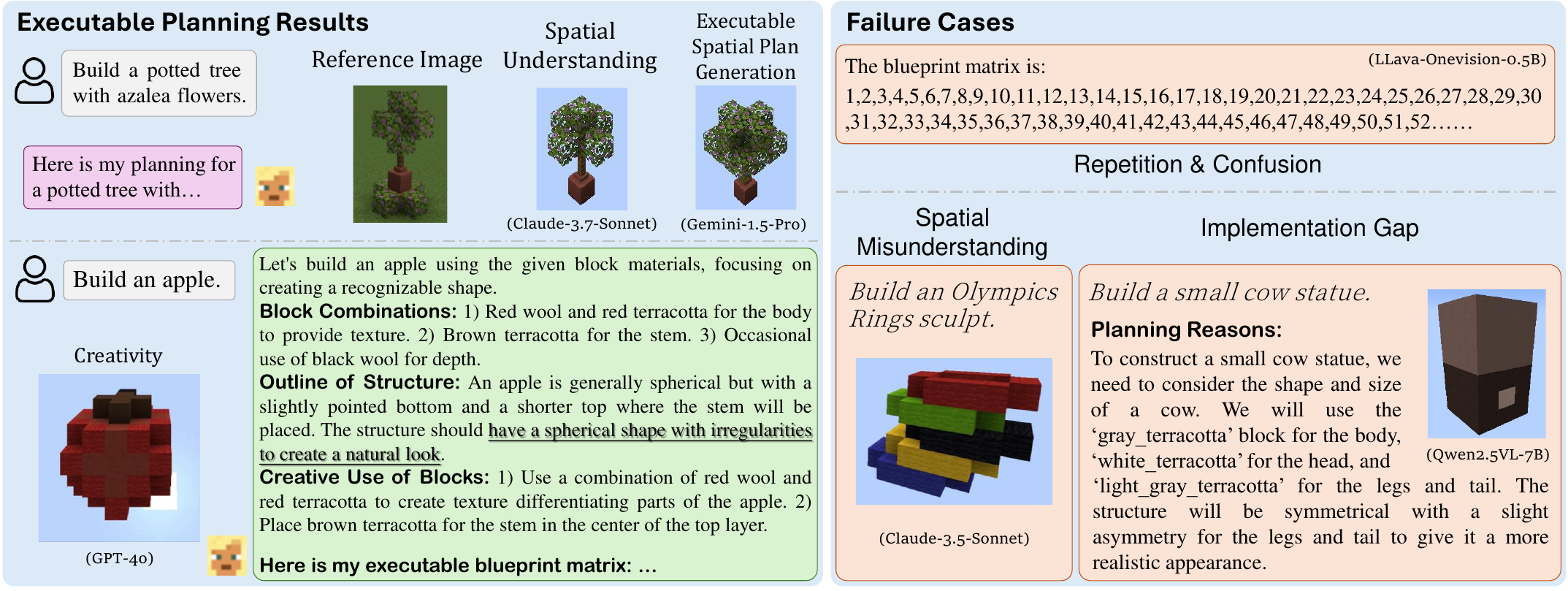}
\caption{Visualization of executable planning results (left) and failure cases (right).}
\label{fig:visual}
\vspace{-0.2cm}
\end{figure*}

We include evaluation results of tested agents and Output Success Rate (OSR) in Table \ref{tab1_metrics} and Figure \ref{fig:bar}, respectively. We also provide some output results and failure cases in Figure \ref{fig:visual} for specific analyses.

\noindent \textbf{Task Performance Results.}
We evaluate 13 MLLM-based agents on our MineAnyBuild benchmark, including 7 proprietary models and 6 open-source models. From Table~\ref{tab1_metrics}, we can see that for most proprietary models, the performances are much better than those of open-source models. 
However, these proprietary models still perform relatively poorly in terms of the average absolute scores,
e.g., even the GPT-4o with the highest overall score of 41.02 achieves less than half of the full score of 100.
We analyze the task-specific findings as follows:

(1) \textbf{Executable Spatial Plan Generation}:
For some low-parameter MLLM-based agents (e.g. InternVL2.5-2B/4B), they tend to understand the basic elements in the given instruction or image, and offer a simple or detailed plan for constructing. However, they often encounter difficulties when generating the executable spatial plan and cannot convert their planning into an executable 3D matrix, thus leading to scores under 0.4 as shown in Table~\ref{tab1_metrics}.
For some large-parameter models, they can usually understand the block materials partly compared to low-parameter ones, and can actively select some diverse blocks to build structures. Nevertheless, they often fail to understand the correlations between various combinations of block materials, resulting in a faulty completion of the final building and thus poor scores (from 0.63 to 1.29 in Table~\ref{tab1_metrics}) by critic model. Proprietary models often achieve a relatively good balance in this aspect, but their planning is generally limited to a boxy or conservative design, and therefore their creations are not highly appreciated by the critic model with the average score of 3.01 ultimately.


(2) \textbf{Creativity}: 
Most proprietary MLLM-based agents can leverage their imagination to construct relatively novel designs, but their 3D architectural capabilities are weak, leading to the difficulty in outputting the executable plans corresponding to their planning and design.
Conversely, open-source MLLMs receive lower scores frequently due to their invalid output results rather than the creative plans they generate, yielding a maximum score of 1.34 as quantified in Table~\ref{tab1_metrics}.

(3) \textbf{Spatial Understanding}:
In Table~\ref{tab1_metrics}, the majority of proprietary models achieve solid results, while Gemini-1.5-Flash frequently generates matrices with more than three dimensions resulting execution errors, which suggests a limited grasp of structural understanding. For open-source models, they struggle with the building structures and mainly respond with repeated or increasing matrix results shown in Figure~\ref{fig:visual}, which points to unclear interpretations of task goals.

(4) \textbf{Spatial Reasoning}:
Spatial reasoning tasks, i.e., mental rotation experiments, require agents to simulate how humans' brain recognizes and moves the stimuli by rotating 3D objects in the mental representation based on the reference stimulus. The distractors are generally mirror-reversed geometries of the stimuli with extra rotations to increase task difficulty. 
From Table~\ref{tab1_metrics}, we can observe that most MLLM-based agents perform poorly on this task, where even the top-performing model, GPT-4o-mini, obtains merely 26.7\% accuracy. Some models that are more capable in general AI tasks, e.g. GPT-4o, perform worse than GPT-4o-mini,
which indicates that our spatial reasoning task still remains challenging for most MLLM-based agents.


(5) \textbf{Spatial Commonsense}:
We evaluated these agents on spatial commonsense tasks relevant to humans' daily life, including the rationality of location, {\it etc}. The results in Table~\ref{tab1_metrics} reveal that most proprietary models have abundant spatial commonsense, 
achieving comparable responses to human-annotated answers.
Open-source MLLM-based agents also show competent performances though with marginally lower consistency scores.


\noindent \textbf{Output Success Rate.}
We statistically calculate the proportion of these MLLM-based agents that successfully respond with executable plans. In the Figure~\ref{fig:bar}, we can observe that for the majority of proprietary models, their instruction-following capability and comprehension of 3D data are relatively strong, thus they can generate the corresponding executable blueprints according to their spatial planning.
Gemini-1.5-Flash scores 73.81 on OSR which is below the mean line of 93.40 due to its incorrect understanding of the output dimension of the executable plan.
The average line of open-source MLLM-based agents is quite lower than that of proprietary models, revealing that these agents are only effective for basic visual or textual understanding, while further training is still required for these 3D spatial tasks, such as spatial planning.
Full metrics of Figure~\ref{fig:bar} are provided in the \supp{}.

\noindent \textbf{Planning Output Visualization.} We visualize some planning results in Figure~\ref{fig:visual}.
We can find that for some relatively easier tasks, most agents with strong capabilities can achieve great performance similar to the structure in the reference image. For example, as shown in Figure~\ref{fig:visual}, agents are required to {\it build a potted tree with azalea flowers}, and Claude-3.7-Sonnet and Gemini-1.5-Pro show effective results under the tasks of spatial understanding and executable
spatial plan generation, respectively. For the creativity task, the agent accessing GPT-4o can analyze and plan what blocks should be utilized and in what form to combine sub-structures into an integral whole. Moreover, the agent tend to consider how to increase the diversity and creativity of the overall structure and appearance, revealing its capabilities of spatial intelligence.
More visualization results of all tasks are provided in the \supp{}.

\noindent \textbf{Failure Cases Analysis.}
We provide some failure cases in Figure~\ref{fig:visual}. We can observe that there are several causes of failure, which leads to agents being unable to generate the executable results based on their planning. 
For some low-parameter open-source MLLM-based agents, they have difficulty handling the 3D executable structures well, generating repetitive or confusing blueprints, leading to compilation failure. Some powerful proprietary models can understand some requirements and build the  substructures, but there is still a spatial misunderstanding in their planning of combining them. For example, Claude-3.5-Sonnet wrongly overlaps the five rings of the Olympics Rings instead of laying them flat on the same plane, which is not in line with commonsense. For most MLLM-based agents, there is usually a severe implementation gap, i.e., they can not convert their textural planning into a spatial structure, which is precisely their huge defect in spatial planning.
More visualization results of failure cases are provided in the \supp{}.
Moreover, we provide deep analyses of the failure reasons for the cases with a summarization as follows: 

\textbf{(1) Spatial Misunderstanding:}
Agents frequently misinterpret 3D positional relationships or fail to maintain the correct spatial arrangements, which highlights a persistent weakness in spatial grounding and planning.

\textbf{(2) Implementation Gap:}
The agents have a central issue that they can not transform high-level textual plans into precise and executable blueprint matrices. The integration of substructures often fails due to incorrect block indexing, orientation errors or inconsistent spatial logic, leading to blueprint parsing or execution failures. This is essentially because the model's understanding of data structures such as codes or DSL and 3D matrices from a numerical perspective is still limited. If these models strengthen the training of spatial data, it may enhance the capabilities of these agents.

\textbf{(3) Structural Degeneration under Complexity:}
When the tasks demand non-cubic, asymmetric or creative designs, the agents tend to collapse into simple and box-like outputs or disorganized results. This indicates that their limited ability to scale from basic patterns to more abstract and complex architectural concepts.

These failure modes reflect deeper limitations in MLLM's capabilities to perform hierarchical spatial planning, maintain geometric consistency and ground language into manipulable 3D structures. They also provide more research directions for MLLMs, e.g., to improve multi-modal spatial understanding, align linguistic abstraction with executable plans or enhance agent’s ability for structural composition in open-ended 3D environments.

\section{Related Works}
\label{related_works}

\noindent \textbf{Spatial Intelligence.}
Spatial intelligence involves thinking about the shapes and arrangements of objects in space and about spatial processes, such as the deformation of objects, and the movement of objects and other entities through space.
Current works mainly focus on spatial understanding and spatial reasoning~\cite{yang2024thinking,chen2024spatialvlm,guo2024drivemllm,tang2025lego,cheng2024spatialrgpt,li2025sti,du2024embspatial,zhang2024vlabench,liu2025spatialcot,qu2025spatialvla}.
VSI-Bench~\cite{yang2024thinking} first introduces the definition of visual-spatial intelligence and proposes a benchmark for it.
SpatialVLM~\cite{chen2024spatialvlm} presents an automatic framework generating millions of VQA samples of spatial reasoning for VLMs' evaluation.
Lego-Puzzles~\cite{tang2025lego} introduces a scalable benchmark with several VQA samples including tasks in multi-step spatial reasoning.
However, these benchmarks suffer from the gap between abstract spatial understanding and concrete task execution. In this paper, we introduce an innovative benchmark concentrating on spatial planning, where the open-world AI agents need to generate executable spatial plans based on its spatial perception and cognition for architecture and indoor decorations. We also introduce diverse evaluation dimensions such as creativity and spatial commonsense to realize a comprehensive assessment for spatial planning capabilities.

\noindent \textbf{Minecraft for AI Research.} 
Minecraft is a 3D world sandbox video game with diverse game mechanics supporting various tasks and activities. Benefiting from its open-ended property, the training and evaluation of autonomous agents built on Minecraft are quite inspiring for the research in the field of artificial intelligence and embodied AI. There are several related works~\cite{fan2022minedojo,guo2025mineworld,baker2022video,wang2023voyager,li2024optimus,li2025optimus,cai2025rocket,long2024teamcraft,cai2024rocket} contributing to the development
in recent years. VPT~\cite{baker2022video}  utilizes Youtube videos for agents' large-scale pretraining. MineDojo~\cite{fan2022minedojo} features a massive database collected automatically from the Internet and learns a 
MineCLIP model
by watching thousands of Youtube videos. 
Voyager~\cite{wang2023voyager} imitate behavior by pseudo-labeling actions by plugging GPT-4 while
Optimus-2~\cite{li2025optimus} learned a VLA-based model with MLLMs for high-level planning.
These works are merely confined to traditional embodied planning tasks like skill learning or tech-tree goals.
Compared to these works, we propose a new benchmark MineAnyBuild to evaluate 
AI agents in spatial intelligence, which is an emerging research field regarding the ability of AI agents to reason about 3D space.

\section{Conclusion}
\label{conclu}
We introduce MineAnyBuild, an innovative benchmark designed to evaluate spatial planning for open-world AI agents. Our MineAnyBuild consists of 4,000 curated tasks with 500+ buildings and decoration assets for evaluating spatial planning, and approximately 2,000 VQA pairs for spatial reasoning and commonsense evaluation. 
Extensive experiments on 13 advanced MLLM-based agents reflects that there is still a great growth space for spatial intelligence of agents.
We believe that our MineAnyBuild will pioneer a novel paradigm to evaluate spatial intelligence, while advancing the development of open-world AI agents with spatial planning capabilities.

\section{Acknowledgments}
This work is supported by National Key Research and Development Program of China (2024YFE0203100), National Natural Science Foundation of China (NSFC) under Grants No.62476293, National Postdoctoral Program for Innovative Talents under Grant Number BX20250379, China Postdoctoral Science Foundation under Grant Number 2025M771521, and General Embodied AI Center of Sun Yat-sen University.

\newpage
\bibliographystyle{unsrt}
\bibliography{nips}

\begin{thebibliography}{10}

\bibitem{song2024robospatial}
Chan~Hee Song, Valts Blukis, Jonathan Tremblay, Stephen Tyree, Yu~Su, and Stan Birchfield.
\newblock Robospatial: Teaching spatial understanding to 2d and 3d vision-language models for robotics.
\newblock {\em arXiv preprint arXiv:2411.16537}, 2024.

\bibitem{ray2024sat}
Arijit Ray, Jiafei Duan, Reuben Tan, Dina Bashkirova, Rose Hendrix, Kiana Ehsani, Aniruddha Kembhavi, Bryan~A Plummer, Ranjay Krishna, Kuo-Hao Zeng, et~al.
\newblock Sat: Spatial aptitude training for multimodal language models.
\newblock {\em arXiv preprint arXiv:2412.07755}, 2024.

\bibitem{li2024seeground}
Rong Li, Shijie Li, Lingdong Kong, Xulei Yang, and Junwei Liang.
\newblock Seeground: See and ground for zero-shot open-vocabulary 3d visual grounding.
\newblock {\em arXiv preprint arXiv:2412.04383}, 2024.

\bibitem{zhang2024spartun3d}
Yue Zhang, Zhiyang Xu, Ying Shen, Parisa Kordjamshidi, and Lifu Huang.
\newblock Spartun3d: Situated spatial understanding of 3d world in large language models.
\newblock {\em arXiv preprint arXiv:2410.03878}, 2024.

\bibitem{liu2023visual}
Haotian Liu, Chunyuan Li, Qingyang Wu, and Yong~Jae Lee.
\newblock Visual instruction tuning.
\newblock {\em Advances in neural information processing systems}, 36:34892--34916, 2023.

\bibitem{openai2023gpt4v-system}
OpenAI.
\newblock Gpt-4v(ision) system card, 2023.

\bibitem{team2023gemini}
Gemini Team, Rohan Anil, Sebastian Borgeaud, Jean-Baptiste Alayrac, Jiahui Yu, Radu Soricut, Johan Schalkwyk, Andrew~M Dai, Anja Hauth, Katie Millican, et~al.
\newblock Gemini: a family of highly capable multimodal models.
\newblock {\em arXiv preprint arXiv:2312.11805}, 2023.

\bibitem{yang2024thinking}
Jihan Yang, Shusheng Yang, Anjali~W Gupta, Rilyn Han, Li~Fei-Fei, and Saining Xie.
\newblock Thinking in space: How multimodal large language models see, remember, and recall spaces.
\newblock {\em arXiv preprint arXiv:2412.14171}, 2024.

\bibitem{chen2024spatialvlm}
Boyuan Chen, Zhuo Xu, Sean Kirmani, Brain Ichter, Dorsa Sadigh, Leonidas Guibas, and Fei Xia.
\newblock Spatialvlm: Endowing vision-language models with spatial reasoning capabilities.
\newblock In {\em Proceedings of the IEEE/CVF Conference on Computer Vision and Pattern Recognition}, pages 14455--14465, 2024.

\bibitem{guo2024drivemllm}
Xianda Guo, Ruijun Zhang, Yiqun Duan, Yuhang He, Chenming Zhang, Shuai Liu, and Long Chen.
\newblock Drivemllm: A benchmark for spatial understanding with multimodal large language models in autonomous driving.
\newblock {\em arXiv preprint arXiv:2411.13112}, 2024.

\bibitem{tang2025lego}
Kexian Tang, Junyao Gao, Yanhong Zeng, Haodong Duan, Yanan Sun, Zhening Xing, Wenran Liu, Kaifeng Lyu, and Kai Chen.
\newblock Lego-puzzles: How good are mllms at multi-step spatial reasoning?
\newblock {\em arXiv preprint arXiv:2503.19990}, 2025.

\bibitem{fan2022minedojo}
Linxi Fan, Guanzhi Wang, Yunfan Jiang, Ajay Mandlekar, Yuncong Yang, Haoyi Zhu, Andrew Tang, De-An Huang, Yuke Zhu, and Anima Anandkumar.
\newblock Minedojo: Building open-ended embodied agents with internet-scale knowledge.
\newblock {\em Advances in Neural Information Processing Systems}, 35:18343--18362, 2022.

\bibitem{wang2023voyager}
Guanzhi Wang, Yuqi Xie, Yunfan Jiang, Ajay Mandlekar, Chaowei Xiao, Yuke Zhu, Linxi Fan, and Anima Anandkumar.
\newblock Voyager: An open-ended embodied agent with large language models.
\newblock {\em arXiv preprint arXiv:2305.16291}, 2023.

\bibitem{shepard1971mental}
Roger~N Shepard and Jacqueline Metzler.
\newblock Mental rotation of three-dimensional objects.
\newblock {\em Science}, 171(3972):701--703, 1971.

\bibitem{shepard1988mental}
Shenna Shepard and Douglas Metzler.
\newblock Mental rotation: effects of dimensionality of objects and type of task.
\newblock {\em Journal of experimental psychology: Human perception and performance}, 14(1):3, 1988.

\bibitem{zeng2023evaluating}
Zhiyuan Zeng, Jiatong Yu, Tianyu Gao, Yu~Meng, Tanya Goyal, and Danqi Chen.
\newblock Evaluating large language models at evaluating instruction following.
\newblock {\em arXiv preprint arXiv:2310.07641}, 2023.

\bibitem{zhou2023instruction}
Jeffrey Zhou, Tianjian Lu, Swaroop Mishra, Siddhartha Brahma, Sujoy Basu, Yi~Luan, Denny Zhou, and Le~Hou.
\newblock Instruction-following evaluation for large language models.
\newblock {\em arXiv preprint arXiv:2311.07911}, 2023.

\bibitem{lou2024large}
Renze Lou, Kai Zhang, and Wenpeng Yin.
\newblock Large language model instruction following: A survey of progresses and challenges.
\newblock {\em Computational Linguistics}, 50(3):1053--1095, 2024.

\bibitem{hu20243d}
Shiying Hu, Zengrong Huang, Chengpeng Hu, and Jialin Liu.
\newblock 3d building generation in minecraft via large language models.
\newblock In {\em 2024 IEEE Conference on Games (CoG)}, pages 1--4. IEEE, 2024.

\bibitem{barthet2022open}
Matthew Barthet, Antonios Liapis, and Georgios~N Yannakakis.
\newblock Open-ended evolution for minecraft building generation.
\newblock {\em IEEE Transactions on Games}, 15(4):603--612, 2022.

\bibitem{chen2024apt}
Jun~Yu Chen and Tao Gao.
\newblock Apt: Architectural planning and text-to-blueprint construction using large language models for open-world agents.
\newblock {\em arXiv preprint arXiv:2411.17255}, 2024.

\bibitem{earle2024dreamcraft}
Sam Earle, Filippos Kokkinos, Yuhe Nie, Julian Togelius, and Roberta Raileanu.
\newblock Dreamcraft: Text-guided generation of functional 3d environments in minecraft.
\newblock In {\em Proceedings of the 19th International Conference on the Foundations of Digital Games}, pages 1--15, 2024.

\bibitem{byrne1989spatial}
Ruth~MJ Byrne and Philip~N Johnson-Laird.
\newblock Spatial reasoning.
\newblock {\em Journal of memory and language}, 28(5):564--575, 1989.

\bibitem{clements1992geometry}
Douglas~H Clements and Michael~T Battista.
\newblock Geometry and spatial reasoning.
\newblock {\em Handbook of research on mathematics teaching and learning: A project of the National Council of Teachers of Mathematics}, pages 420--464, 1992.

\bibitem{hegarty2010components}
Mary Hegarty.
\newblock Components of spatial intelligence.
\newblock In {\em Psychology of learning and motivation}, volume~52, pages 265--297. Elsevier, 2010.

\bibitem{vandenberg1978mental}
Steven~G Vandenberg and Allan~R Kuse.
\newblock Mental rotations, a group test of three-dimensional spatial visualization.
\newblock {\em Perceptual and motor skills}, 47(2):599--604, 1978.

\bibitem{davis2015commonsense}
Ernest Davis and Gary Marcus.
\newblock Commonsense reasoning and commonsense knowledge in artificial intelligence.
\newblock {\em Communications of the ACM}, 58(9):92--103, 2015.

\bibitem{liu2022things}
Xiao Liu, Da~Yin, Yansong Feng, and Dongyan Zhao.
\newblock Things not written in text: Exploring spatial commonsense from visual signals.
\newblock {\em arXiv preprint arXiv:2203.08075}, 2022.

\bibitem{collell2018acquiring}
Guillem Collell, Luc Van~Gool, and Marie-Francine Moens.
\newblock Acquiring common sense spatial knowledge through implicit spatial templates.
\newblock In {\em Proceedings of the AAAI conference on artificial intelligence}, 2018.

\bibitem{grabcraft}
GrabCraft LLC.
\newblock Grabcraft - the biggest library of minecraft objects, models, floor plans, ideas, and blueprints.
\newblock \url{https://www.grabcraft.com/}.

\bibitem{official_wiki}
Citricsquid.
\newblock Minecraft wiki.
\newblock \url{https://minecraft.wiki/}.

\bibitem{fandom_wiki}
Fandom.
\newblock Fandom wiki for minecraft.
\newblock \url{https://minecraft.fandom.com/wiki/Minecraft_Wiki}.

\bibitem{mineflayer}
PrismarineJS.
\newblock Mineflayer.
\newblock \url{https://github.com/PrismarineJS/mineflayer}.

\bibitem{claude}
Anthropic.
\newblock The claude 3 model family: Opus, sonnet, haiku, 2024.

\bibitem{gpt4o}
OpenAI:Josh Achiam, Steven Adler, Sandhini Agarwal, Lama Ahmad, Ilge Akkaya, FlorenciaLeoni Aleman, Diogo Almeida, Janko Altenschmidt, Sam Altman, Shyamal Anadkat, Red Avila, Igor Babuschkin, Suchir Balaji, Valerie Balcom, Paul Baltescu, Haiming Bao, Mo~Bavarian, Jeff Belgum, Irwan Bello, Jake Berdine, Gabriel Bernadett-Shapiro, Christopher Berner, Lenny Bogdonoff, Oleg Boiko, Madelaine Boyd, Anna-Luisa Brakman, Greg Brockman, Tim Brooks, Miles Brundage, Kevin Button, Trevor Cai, Rosie Campbell, Andrew Cann, Brittany Carey, Chelsea Carlson, Rory Carmichael, Brooke Chan, Che Chang, Fotis Chantzis, Derek Chen, Sully Chen, Ruby Chen, Jason Chen, Mark Chen, Ben Chess, Chester Cho, Casey Chu, HyungWon Chung, Dave Cummings, and Jeremiah Currier.
\newblock Gpt-4 technical report.
\newblock {\em arXiv preprint arXiv:2303.08774}, Dec 2023.

\bibitem{chen2024expanding}
Zhe Chen, Weiyun Wang, Yue Cao, Yangzhou Liu, Zhangwei Gao, Erfei Cui, Jinguo Zhu, Shenglong Ye, Hao Tian, Zhaoyang Liu, et~al.
\newblock Expanding performance boundaries of open-source multimodal models with model, data, and test-time scaling.
\newblock {\em arXiv preprint arXiv:2412.05271}, 2024.

\bibitem{Qwen2.5-VL}
Shuai Bai, Keqin Chen, Xuejing Liu, Jialin Wang, Wenbin Ge, Sibo Song, Kai Dang, Peng Wang, Shijie Wang, Jun Tang, Humen Zhong, Yuanzhi Zhu, Mingkun Yang, Zhaohai Li, Jianqiang Wan, Pengfei Wang, Wei Ding, Zheren Fu, Yiheng Xu, Jiabo Ye, Xi~Zhang, Tianbao Xie, Zesen Cheng, Hang Zhang, Zhibo Yang, Haiyang Xu, and Junyang Lin.
\newblock Qwen2.5-vl technical report.
\newblock {\em arXiv preprint arXiv:2502.13923}, 2025.

\bibitem{li2024llava}
Bo~Li, Yuanhan Zhang, Dong Guo, Renrui Zhang, Feng Li, Hao Zhang, Kaichen Zhang, Yanwei Li, Ziwei Liu, and Chunyuan Li.
\newblock Llava-onevision: Easy visual task transfer.
\newblock {\em arXiv preprint arXiv:2408.03326}, 2024.

\bibitem{gpt41}
OpenAI.
\newblock Introducing gpt-4.1 in the api.
\newblock \url{https://openai.com/index/gpt-4-1/}, 2025.

\bibitem{cheng2024spatialrgpt}
An-Chieh Cheng, Hongxu Yin, Yang Fu, Qiushan Guo, Ruihan Yang, Jan Kautz, Xiaolong Wang, and Sifei Liu.
\newblock Spatialrgpt: Grounded spatial reasoning in vision language models.
\newblock {\em arXiv preprint arXiv:2406.01584}, 2024.

\bibitem{li2025sti}
Yun Li, Yiming Zhang, Tao Lin, XiangRui Liu, Wenxiao Cai, Zheng Liu, and Bo~Zhao.
\newblock Sti-bench: Are mllms ready for precise spatial-temporal world understanding?
\newblock {\em arXiv preprint arXiv:2503.23765}, 2025.

\bibitem{du2024embspatial}
Mengfei Du, Binhao Wu, Zejun Li, Xuanjing Huang, and Zhongyu Wei.
\newblock Embspatial-bench: Benchmarking spatial understanding for embodied tasks with large vision-language models.
\newblock {\em arXiv preprint arXiv:2406.05756}, 2024.

\bibitem{zhang2024vlabench}
Shiduo Zhang, Zhe Xu, Peiju Liu, Xiaopeng Yu, Yuan Li, Qinghui Gao, Zhaoye Fei, Zhangyue Yin, Zuxuan Wu, Yu-Gang Jiang, et~al.
\newblock Vlabench: A large-scale benchmark for language-conditioned robotics manipulation with long-horizon reasoning tasks.
\newblock {\em arXiv preprint arXiv:2412.18194}, 2024.

\bibitem{liu2025spatialcot}
Yuecheng Liu, Dafeng Chi, Shiguang Wu, Zhanguang Zhang, Yaochen Hu, Lingfeng Zhang, Yingxue Zhang, Shuang Wu, Tongtong Cao, Guowei Huang, et~al.
\newblock Spatialcot: Advancing spatial reasoning through coordinate alignment and chain-of-thought for embodied task planning.
\newblock {\em arXiv preprint arXiv:2501.10074}, 2025.

\bibitem{qu2025spatialvla}
Delin Qu, Haoming Song, Qizhi Chen, Yuanqi Yao, Xinyi Ye, Yan Ding, Zhigang Wang, JiaYuan Gu, Bin Zhao, Dong Wang, et~al.
\newblock Spatialvla: Exploring spatial representations for visual-language-action model.
\newblock {\em arXiv preprint arXiv:2501.15830}, 2025.

\bibitem{guo2025mineworld}
Junliang Guo, Yang Ye, Tianyu He, Haoyu Wu, Yushu Jiang, Tim Pearce, and Jiang Bian.
\newblock Mineworld: a real-time and open-source interactive world model on minecraft.
\newblock {\em arXiv preprint arXiv:2504.08388}, 2025.

\bibitem{baker2022video}
Bowen Baker, Ilge Akkaya, Peter Zhokov, Joost Huizinga, Jie Tang, Adrien Ecoffet, Brandon Houghton, Raul Sampedro, and Jeff Clune.
\newblock Video pretraining (vpt): Learning to act by watching unlabeled online videos.
\newblock {\em Advances in Neural Information Processing Systems}, 35:24639--24654, 2022.

\bibitem{li2024optimus}
Zaijing Li, Yuquan Xie, Rui Shao, Gongwei Chen, Dongmei Jiang, and Liqiang Nie.
\newblock Optimus-1: Hybrid multimodal memory empowered agents excel in long-horizon tasks.
\newblock {\em arXiv preprint arXiv:2408.03615}, 2024.

\bibitem{li2025optimus}
Zaijing Li, Yuquan Xie, Rui Shao, Gongwei Chen, Dongmei Jiang, and Liqiang Nie.
\newblock Optimus-2: Multimodal minecraft agent with goal-observation-action conditioned policy.
\newblock {\em arXiv preprint arXiv:2502.19902}, 2025.

\bibitem{cai2025rocket}
Shaofei Cai, Zhancun Mu, Anji Liu, and Yitao Liang.
\newblock Rocket-2: Steering visuomotor policy via cross-view goal alignment.
\newblock {\em arXiv preprint arXiv:2503.02505}, 2025.

\bibitem{long2024teamcraft}
Qian Long, Zhi Li, Ran Gong, Ying~Nian Wu, Demetri Terzopoulos, and Xiaofeng Gao.
\newblock Teamcraft: A benchmark for multi-modal multi-agent systems in minecraft.
\newblock {\em arXiv preprint arXiv:2412.05255}, 2024.

\bibitem{cai2024rocket}
Shaofei Cai, Zihao Wang, Kewei Lian, Zhancun Mu, Xiaojian Ma, Anji Liu, and Yitao Liang.
\newblock Rocket-1: Mastering open-world interaction with visual-temporal context prompting.
\newblock {\em arXiv preprint arXiv:2410.17856}, 2024.

\bibitem{faraboschi2023artificial}
Paolo Faraboschi, Eitan Frachtenberg, Phil Laplante, Dejan Milojicic, and Roberto Saracco.
\newblock Artificial general intelligence: Humanity’s downturn or unlimited prosperity.
\newblock {\em Computer}, 56(10):93--101, 2023.

\bibitem{heilman2003creative}
Kenneth~M Heilman, Stephen~E Nadeau, and David~O Beversdorf.
\newblock Creative innovation: possible brain mechanisms.
\newblock {\em Neurocase}, 9(5):369--379, 2003.

\bibitem{deitke2023objaverse}
Matt Deitke, Dustin Schwenk, Jordi Salvador, Luca Weihs, Oscar Michel, Eli VanderBilt, Ludwig Schmidt, Kiana Ehsani, Aniruddha Kembhavi, and Ali Farhadi.
\newblock Objaverse: A universe of annotated 3d objects.
\newblock In {\em Proceedings of the IEEE/CVF conference on computer vision and pattern recognition}, pages 13142--13153, 2023.

\bibitem{guss2019minerl}
William~H Guss, Brandon Houghton, Nicholay Topin, Phillip Wang, Cayden Codel, Manuela Veloso, and Ruslan Salakhutdinov.
\newblock Minerl: A large-scale dataset of minecraft demonstrations.
\newblock {\em arXiv preprint arXiv:1907.13440}, 2019.

\bibitem{cai2024minsstudio}
Shaofei Cai, Zhancun Mu, Kaichen He, Bowei Zhang, Xinyue Zheng, Anji Liu, and Yitao Liang.
\newblock Minsstudio: A streamlined package for minecraft ai agent development.
\newblock {\em arXiv preprint arXiv:2412.18293}, 2024.

\bibitem{lifshitz2023steve}
Shalev Lifshitz, Keiran Paster, Harris Chan, Jimmy Ba, and Sheila McIlraith.
\newblock Steve-1: A generative model for text-to-behavior in minecraft.
\newblock {\em Advances in Neural Information Processing Systems}, 36:69900--69929, 2023.

\bibitem{wang2024jarvis}
Zihao Wang, Shaofei Cai, Anji Liu, Yonggang Jin, Jinbing Hou, Bowei Zhang, Haowei Lin, Zhaofeng He, Zilong Zheng, Yaodong Yang, et~al.
\newblock Jarvis-1: Open-world multi-task agents with memory-augmented multimodal language models.
\newblock {\em IEEE Transactions on Pattern Analysis and Machine Intelligence}, 2024.

\bibitem{cai2024groot}
Shaofei Cai, Bowei Zhang, Zihao Wang, Haowei Lin, Xiaojian Ma, Anji Liu, and Yitao Liang.
\newblock Groot-2: Weakly supervised multi-modal instruction following agents.
\newblock {\em arXiv preprint arXiv:2412.10410}, 2024.

\bibitem{wang2023describe}
Zihao Wang, Shaofei Cai, Guanzhou Chen, Anji Liu, Xiaojian Ma, and Yitao Liang.
\newblock Describe, explain, plan and select: Interactive planning with large language models enables open-world multi-task agents.
\newblock {\em arXiv preprint arXiv:2302.01560}, 2023.

\bibitem{qin2024mp5}
Yiran Qin, Enshen Zhou, Qichang Liu, Zhenfei Yin, Lu~Sheng, Ruimao Zhang, Yu~Qiao, and Jing Shao.
\newblock Mp5: A multi-modal open-ended embodied system in minecraft via active perception.
\newblock In {\em 2024 IEEE/CVF Conference on Computer Vision and Pattern Recognition (CVPR)}, pages 16307--16316. IEEE, 2024.

\bibitem{wei2022chain}
Jason Wei, Xuezhi Wang, Dale Schuurmans, Maarten Bosma, Fei Xia, Ed~Chi, Quoc~V Le, Denny Zhou, et~al.
\newblock Chain-of-thought prompting elicits reasoning in large language models.
\newblock {\em Advances in neural information processing systems}, 35:24824--24837, 2022.

\bibitem{csato2017ranking}
L{\'a}szl{\'o} Csat{\'o}.
\newblock On the ranking of a swiss system chess team tournament.
\newblock {\em Annals of Operations Research}, 254(1):17--36, 2017.

\bibitem{bersier2025cognitive}
Nadia~M Bersier, Eleonora Fornari, Raffaella~I Rumiati, and Silvio Ionta.
\newblock Cognitive traits shape the brain activity associated with mental rotation.
\newblock {\em Cerebral Cortex}, 35(4):bhaf069, 2025.

\bibitem{jiang2025behavior}
Yunfan Jiang, Ruohan Zhang, Josiah Wong, Chen Wang, Yanjie Ze, Hang Yin, Cem Gokmen, Shuran Song, Jiajun Wu, and Li~Fei-Fei.
\newblock Behavior robot suite: Streamlining real-world whole-body manipulation for everyday household activities.
\newblock {\em arXiv preprint arXiv:2503.05652}, 2025.

\end{thebibliography}

\newpage
\section*{NeurIPS Paper Checklist}

\begin{enumerate}

\item {\bf Claims}
    \item[] Question: Do the main claims made in the abstract and introduction accurately reflect the paper's contributions and scope?
    \item[] Answer: \answerYes{}
    \item[] Justification: 
    We clearly state our contributions and problem scope in Section \ref{intro} of our paper.

\item {\bf Limitations}
    \item[] Question: Does the paper discuss the limitations of the work performed by the authors?
    \item[] Answer: \answerYes{}
    \item[] Justification:
    We discuss our limitations in Section.\redbox{A} of the \supp{}.

\item {\bf Theory assumptions and proofs}
    \item[] Question: For each theoretical result, does the paper provide the full set of assumptions and a complete (and correct) proof?
    \item[] Answer: \answerNA{}
    \item[] Justification: 
    Our paper does not include theoretical results such as theory assumptions and proofs.

    \item {\bf Experimental result reproducibility}
    \item[] Question: Does the paper fully disclose all the information needed to reproduce the main experimental results of the paper to the extent that it affects the main claims and/or conclusions of the paper (regardless of whether the code and data are provided or not)?
    \item[] Answer: \answerYes{}
    \item[] Justification: 
    We provide the information, code and datasets in Section \redbox{D} of the \supp{} for reproduction. We will also make our datasets and codes public in the future.

\item {\bf Open access to data and code}
    \item[] Question: Does the paper provide open access to the data and code, with sufficient instructions to faithfully reproduce the main experimental results, as described in supplemental material?
    \item[] Answer: \answerYes{}
    \item[] Justification: 
    Our paper in this submission track provides Dataset URL and Code URL in order to facilitate the review process.
    We will make our datasets and codes public with sufficient instructions in the future.

\item {\bf Experimental setting/details}
    \item[] Question: Does the paper specify all the training and test details (e.g., data splits, hyperparameters, how they were chosen, type of optimizer, etc.) necessary to understand the results?
    \item[] Answer: \answerYes{}
    \item[] Justification: 
    We provide all the experimental details in Section \ref{exp} of our paper and additional results in Section \redbox{F} of the \supp{}.

\item {\bf Experiment statistical significance}
    \item[] Question: Does the paper report error bars suitably and correctly defined or other appropriate information about the statistical significance of the experiments?
    \item[] Answer: \answerYes{}{}
    \item[] Justification: 
     We provide some analyses on experimental results, such as scoring by critic models, in Section \redbox{F} of the \supp{}.

\item {\bf Experiments compute resources}
    \item[] Question: For each experiment, does the paper provide sufficient information on the computer resources (type of compute workers, memory, time of execution) needed to reproduce the experiments?
    \item[] Answer: \answerYes{}
    \item[] Justification: 
    We provide the information of our experimental compute resources in Section \redbox{F} of the \supp{}.
    
\item {\bf Code of ethics}
    \item[] Question: Does the research conducted in the paper conform, in every respect, with the NeurIPS Code of Ethics \url{https://neurips.cc/public/EthicsGuidelines}?
    \item[] Answer: \answerYes{}
    \item[] Justification: 
    Our research conducts in every respect with the NeurIPS Code of Ethics.

\item {\bf Broader impacts}
    \item[] Question: Does the paper discuss both potential positive societal impacts and negative societal impacts of the work performed?
    \item[] Answer: \answerYes{}
    \item[] Justification:
    We discuss the broader impacts of our work in Section \redbox{A} of the \supp{}.
    
\item {\bf Safeguards}
    \item[] Question: Does the paper describe safeguards that have been put in place for responsible release of data or models that have a high risk for misuse (e.g., pretrained language models, image generators, or scraped datasets)?
    \item[] Answer: \answerYes{}
    \item[] Justification:
    We conduct manual reviews on our data obtained from the Internet from containing unsafe images and curate our data for safety as safeguards.

\item {\bf Licenses for existing assets}
    \item[] Question: Are the creators or original owners of assets (e.g., code, data, models), used in the paper, properly credited and are the license and terms of use explicitly mentioned and properly respected?
    \item[] Answer: \answerYes{}
    \item[] Justification: 
    All datasets and models are properly cited in our main text and our \supp{}. 

\item {\bf New assets}
    \item[] Question: Are new assets introduced in the paper well documented and is the documentation provided alongside the assets?
    \item[] Answer: \answerYes{}
    \item[] Justification: 
    We provide the information and document of our datasets in Section \redbox{D} of the \supp{}.

\item {\bf Crowdsourcing and research with human subjects}
    \item[] Question: For crowdsourcing experiments and research with human subjects, does the paper include the full text of instructions given to participants and screenshots, if applicable, as well as details about compensation (if any)? 
    \item[] Answer: \answerNA{}
    \item[] Justification: 
     Our paper does not involve crowdsourcing nor research with human subjects.

\item {\bf Institutional review board (IRB) approvals or equivalent for research with human subjects}
    \item[] Question: Does the paper describe potential risks incurred by study participants, whether such risks were disclosed to the subjects, and whether Institutional Review Board (IRB) approvals (or an equivalent approval/review based on the requirements of your country or institution) were obtained?
    \item[] Answer: \answerNA{}
    \item[] Justification: 
    Our paper does not involve crowdsourcing nor research with human subjects.

\item {\bf Declaration of LLM usage}
    \item[] Question: Does the paper describe the usage of LLMs if it is an important, original, or non-standard component of the core methods in this research? Note that if the LLM is used only for writing, editing, or formatting purposes and does not impact the core methodology, scientific rigorousness, or originality of the research, declaration is not required.
    \item[] Answer: \answerYes{}
    \item[] Justification: 
    We provide the declaration of our LLM usage in Section \redbox{H} in our Supplementary Material.
    However, the usage of LLMs in our work does not impact the scientific rigorousness and originality of the research.

\end{enumerate}

\appendix
\clearpage
\begin{center}
    \LARGE\textbf{Supplementary Material}
\end{center}
\startcontents  
\printcontents{}{1}{\section*{Contents}}

\setcounter{section}{0}  
\newpage

Our Supplementary Material for ``MineAnyBuild: Benchmarking Spatial Planning for Open-world AI Agents'' is organized as follows. 
In Section~\ref{sec_A}, we discuss the impacts, limitations and future directions of our work. 
In Section~\ref{sec_B}, we introduce the Minecraft game and its common simulated environments.
In Section~\ref{sec_C}, we present the datasheets of our datasets, including dataset description, data field descriptions, {\it etc.} 
In Section~\ref{sec_D}, we provide the details for our benchmark, tasks, and data curation pipeline.
In Section~\ref{sec_E}, we present the prompts utilized for proprietary models and critic models in our benchmark.
In Section~\ref{sec_F}, we provide more experimental results, information and analyses.
In Section~\ref{sec_G}, we present more visualization results of our evaluation on AI agents.
In Section~\ref{sec_H}, we declare our usage of LLM in our paper and the \supp{}.

\section{Broader Impacts, Limitations, and Future Directions}
\label{sec_A}
\subsection{Boarder Impacts}

Spatial intelligence has become an important research dimension in developing open-world AI agents for handling various downstream applications. AI agents with spatial intelligence can better satisfy the need of spatial perception, understanding, memorization, {\it etc.}, in tasks like automatic assembly, architectural design, and robotic manipulation. Although existing open-world AI agents driven by Multi-modal Large Language Models (MLLMs) have exhibited astonishing capabilities in various text-based (1D) and image-based (2D) tasks, they still struggle to tackle tasks requiring spatial understanding and cognition from a 3D perspective. Moreover, existing benchmarks designed for evaluating spatial intelligence typically ignore the gap between abstract spatial understanding and actual task execution.

In light of this, we propose an innovative benchmark to evaluate a critical part in spatial intelligence, i.e., {\it spatial planning}, to bridge the gap between spatial reasoning and task completion. Our proposed \textbf{MineAnyBuild} benchmark standardizes task definitions, modules, and interfaces, enabling a comprehensive and rigorous evaluation of spatial planning capabilities for open-world AI agents, especially MLLM-based AI agents. Through our introduced infinitely expandable data collection paradigm, our benchmark can also acquire scalable data for facilitating spatial planning training and evaluation. We believe our proposed benchmark can make a significant stride in spatial intelligence research to promote the development of AI agents capable of spatial planning and reasoning, which brings great benefits for a wide range of real-world applications.

Creativity is also an important indicator for evaluating the aesthetics of 3D space tasks.
Just like text creation in the 1D domain and image editing/painting creation in the 2D domain, with the rapid development of generative AI in recent years, the evaluation of aesthetic features and creative assessment under these dimensional data have always been one of the important current evaluation indicators. Similarly, we set up the creativity task to evaluate the agents' assessment of the aesthetics and 3D potential of space tasks. Agents receive an instruction and are required to brainstorm block combinations for different parts of the architecture and outline a rough structure layout, to find ways to maximize creativity and the dynamic range of possible builds. Therefore, creativity 
not only reflects the cognitive depth of future AI systems, but also emerges as a novel and important criterion for agents towards Artificial General Intelligence (AGI)~\cite{faraboschi2023artificial,heilman2003creative}.

Constrained by the limitations and biases of these AI agents, the potential negative impacts should also be considered when developing them to address real-world applications requiring spatial perception and cognition. For example, the uncertainty and hallucination of their outputs may cause unreliability and even safety problems, e.g., damage of public facilities. Therefore, besides the benchmark designed for encouraging AI agents research, it is also important to establish guidelines for accountability and design risk mitigation strategies when deploying these AI agents in real-world applications.

\subsection{Limitations}
\label{limitations}

The main limitations of this work are summarized as follows:

1) Evaluation tools that we choose are relatively slow and require manual operations. We discuss the choice of tools in Section~\ref{eva_tools}.

2) Evaluation based on a scoring system still has drawbacks compared to evaluation based on models or fixed criteria. 
Although we believe that our scoring-based evaluation can better mimic the human evaluation, it is obvious that this approach is somewhat inferior in terms of scientificity, reliability and stability when compared to some AI model-based evaluators. 

3) The output format based on the 3D blueprint matrix is not the best so far. Although this is the best option we have chosen after careful consideration at present, this approach is not optimal due to the output token limits of MLLMs or the action sequence length of embodied agents.
We also discuss it in Section~\ref{discuss_eva}.

\subsection{Future Directions}

Based on the limitations we discuss above, we outline the future  directions to represent a roadmap for advancing the spatial intelligence research, especially the spatial planning, in this section.

\begin{itemize}

    \item \textbf{Better Evaluation Tool:}
    Currently, the existing evaluation tools and frameworks are suitable for the previous tasks, such as skill learning and tech-tree, but they are not well-adapted to the building construction tasks that we proposed and lack a necessary interface for high-quality visualizations. We hope that more developers can cooperate to jointly develop and improve the evaluation tools for our proposed tasks of spatial planning in the future, so as to better promote the development of AI agents in spatial intelligence.
    
    \item \textbf{Advanced Evaluation System:}
    Designing evaluation methods beyond current scoring-based approach to obtain a more advanced evaluation system is crucial to evaluate AI agents and to improve the quality of agents' spatial planning and intelligence.    
    
    \item \textbf{Optimized Output Format:}
    Developing an output format which is easier for agents to understand and learn than the blueprint matrix or code generation is also beneficial, with which agents are able to ``translate'' their planning into executable results more effectively, reducing errors or biases caused by the output.

    \item \textbf{RL-based Agents:}
    On the basis of multiple existing simulator environments in Minecraft, implementing a basic RL-based agent is important at present. Training RL-based agents can greatly advance the development of Embodied AI research. We would like to develop RL-based agents in the future to better adapt to our MineAnyBuild benchmark and datasets.
    
    \item \textbf{Memory and Learning Modules:}
    We believe that the memory and learning modules are very important for the tasks and benchmarks
    we have proposed. Currently, most of our evaluations are based on zero-shot querying of MLLMs, and they usually respond with plans relying on their original knowledge base. If powerful memory modules are integrated, AI agents can gradually learn from small/simple architectures/assets, gain a deeper understanding of spatial planning step by step, and then complete more complex and large-scale buildings finally, akin to curriculum learning techniques, so as to build an ideal and perfect castle just like human players.
    
    \item \textbf{A General Agent Framework Combining MLLMs and RL Training:}
    Existing embodied AI frameworks mainly revolve around an MLLM-based ``brain'' and bottom-level modules for action execution, and such a framework can be extended to the tasks we proposed. In the future, a general agent framework that combines MLLMs with efficient RL training will greatly enhance the capabilities of agents to perform better on these planning tasks.

    \item \textbf{Transferring to Non-game Scenarios:}
    Our benchmark could be migrated to a similar non-game environment, such as Isaac Sim or AI2Thor simulators, for 3D scene generation tasks. This task requires a framework to retrieve assets from an asset library (e.g., Objaverse~\cite{deitke2023objaverse}) and properly place them into an indoor scene. How to ensure that the positions and rotations of all assets are correctly arranged is exactly a form of spatial planning. This task is one of the cutting-edge research directions in recent years, and the motivation of this is similar to that of our benchmark.

    \item \textbf{Transferring to Real-world Scenarios:}
    How to perform sim-to-real transfer for our benchmark is also one of our interests. We actually watched a video\footnote{Video address: \url{https://www.youtube.com/watch?v=UkyEyVoP8Jc}} on YouTube showing physical blocks with same appearance as those in the game (e.g., grass block), which can be spliced together using the built-in magnets, just like toy blocks and LEGO bricks. We plan to build a physical (real-world) version of our benchmark in the future, and study how agents trained in simulation can be deployed to manipulate these blocks in our real world to construct buildings.
    
\end{itemize}

\section{Minecraft}
\label{sec_B}
Minecraft is a highly popular 3D sandbox video game developed by Microsoft's Mojang Studios. The environment is primarily composed of 3D blocks, each distinguished by pixel-art textures on their surfaces that represent a variety of materials, such as dirt, stone, minerals, water, and trees. Minecraft facilitates a high degree of interaction with these blocks for both human players and AI agents, enabled through well-developed user interfaces or APIs. Players are capable of freely navigating the world, collecting these fundamental blocks, and subsequently positioning them on a standardized 3D integer-based coordinate grid to engage in diverse construction activities. Consequently, the diversity in block types, the extensive freedom of interaction, and the standardized 3D coordinate system establish Minecraft as an ideal environment for assessing the spatial reasoning and planning capabilities of large AI models.

Furthermore, Minecraft is supported by a vast and active user community that consistently generates an abundant and expanding collection of user-generated content. This content, including items such as modifications (mods), skins, texture packs, and custom maps, is readily available for download, a factor that underscores the platform's significant extensibility.

\subsection{Environment Details of Minecraft Game}
The Minecraft environment has several versions that are available for players to download and use. 
In our work, we recommend the version 1.16.5 and 1.20.4 for better adaptation to some simulator environments.
We finally choose the version 1.20.4 for {\it mineflayer} simulator.

The Minecraft environment, specifically version 1.20.4, comprises an extensive set of 830+ standard block types. Each block type is uniquely identified by a textual name, which anyone can easily obtain the information on the website: \url{https://minecraft.wiki/w/Block}. 

In Minecraft, players can do anything they can imagine. Players can build structures, craft tools, smelt ore, brew potions, trade with villagers and wandering traders, attack mobs, grow crops, raise animals in
captivity, etc. Players even can use redstone to build a computer. This is a world of freedom and infinite possibilities.

The Minecraft world is divided into different areas called ``biomes''. These biomes contain different blocks and plants and change how the land is shaped. There are 79 biomes in Minecraft 1.16.5, including ocean, plains, forest, desert, etc. Diverse environments have high requirements for the generalization of agents.

Current works~\cite{fan2022minedojo,guo2025mineworld,baker2022video,wang2023voyager,li2024optimus,li2025optimus,cai2025rocket,long2024teamcraft,cai2024rocket} are merely confined to traditional embodied planning tasks like skill learning or tech-tree goals.
Compared to these works, our benchmark MineAnyBuild is proposed to evaluate AI agents in spatial intelligence, which is an emerging research field regarding the ability of AI agents to reason about 3D space.

\subsection{Simulator Environments}
We compile the simulator environments used in some current works in Table~\ref{tab:mc_env}. It can be observed that most current works mainly focus on several categories such as MineRL~\cite{guss2019minerl}, MineStudio~\cite{cai2024minsstudio}, MineDojo~\cite{fan2022minedojo} and Mineflayer~\cite{mineflayer}.

Our benchmark MineAnyBuild is mainly built on the Mineflayer simulator environment. As we mentioned in Section~\ref{sec_A}, we plan to expand and adapt our benchmark to these RL-based simulator environments in Table~\ref{tab:mc_env} as much as possible in the future.

\begin{table}[h]
\centering
	\fontsize{9}{9}\selectfont
	\caption{Various Minecraft agents in different code environments.
	}\label{tab:mc_env}
		\resizebox{\linewidth}{!}{
	{\renewcommand{\arraystretch}{1.5}	\begin{tabular}{c||ccc}
				\specialrule{.1em}{.05em}{.05em}	
			
    Code Environment&Agent&Publication&Task Type\\ \hline

\multirow{4}{*}{MineRL~\cite{guss2019minerl}}
&VPT~\cite{baker2022video}&NeruIPS 2022&Skill Learning\\
&STEVE-1~\cite{lifshitz2023steve}&NeruIPS 2023&Skill Learning\\
&Optimus-1~\cite{li2024optimus}&NeruIPS 2024&Skill Learning\\
&Optimus-2~\cite{li2025optimus}&CVPR 2025&Skill Learning\\
\hline
\multirow{3}{*}{MineStudio~\cite{cai2024minsstudio}}
&Jarvis-1~\cite{wang2024jarvis}&T-PAMI 2024&Skill Learning\\

&GROOT-2~\cite{cai2024groot}&ICLR 2025&Skill Learning\\
&ROCKET-1~\cite{cai2024rocket}&CVPR 2025&Skill Learning\\
\hline
\multirow{3}{*}{MineDojo~\cite{fan2022minedojo}}
&MineDojo~\cite{fan2022minedojo}&NeruIPS 2022 D\&B&Skill Learning \& Architecture Building\\
&DEPS~\cite{wang2023describe}&NeruIPS 2023&Skill Learning\\
&MP5~\cite{qin2024mp5}&CVPR 2024&Skill Learning\\
\hline
\multirow{2}{*}{Mineflayer~\cite{mineflayer}}
&Voyager~\cite{wang2023voyager}&TMLR 2024&Skill Learning \& Tech Tree\\
&APT~\cite{chen2024apt}&AAAI 2025 Workshop&Architecture Building\\


    
	\specialrule{.1em}{.05em}{.05em}	\end{tabular}}}
\vspace{-0.2cm}	
\end{table}

\subsection{Tools for Evaluation and Visualization}
\label{eva_tools}
We introduce two common tools for evaluation on our benchmark, and discuss our choice between them and the corresponding reasons.

Compared with traditional Minecraft benchmarks and works, 
their evaluation is based on whether a task is successful or not, simply by checking data such as whether atomic actions match or environmental information and states match. 
However, our new benchmark MineAnyBuild focuses on evaluating the overall appearance of buildings and the spatial planning process, so it has high requirements for visual data. Therefore, we must rely on existing Minecraft visualization tools to visualize the buildings constructed by agents and then submit them to evaluators (critic models) for evaluation.

Currently, there are mainly two common visualization tools in relevant research works, namely Mineflayer Viewer and Replay Mod.
We summarize the advantages and disadvantages of these two tools in Table~\ref{tab:tools}.
\begin{table}[h]
\centering
	\caption{Advantages and disadvantages of two visulization tools.
    }\label{tab:tools}
		\resizebox{\linewidth}{!}{
	{\renewcommand{\arraystretch}{2}	\begin{tabular}{c||cc}
				\specialrule{.1em}{.05em}{.05em}	
    Tools&Advantages&Disadvantages\\ \hline
Mineflayer Viewer&\makecell{1) Easy to use.\\2) Quick image acquisition speed.}&\makecell{1) Some blocks cannot be rendered, \\e.g., stairs and slabs.\\2) Fixed perspective.\\3) Poor custonmization.}\\
\hline
Replay Mod&\makecell{1) Any blocks can be correctly rendered.\\2) Customizable perspectives.\\3) High degree of freedom.}&\makecell{1) Slow image acquisition speed.\\2) Non-automated.\\3) Requiring manual operations.}\\

	\specialrule{.1em}{.05em}{.05em}	\end{tabular}}}
\vspace{-0.2cm}	
\end{table}

Next, we will specifically introduce the characteristics and general usage of these two tools, and explain the reasons for our final choice (Replay Mod) in Section~\ref{replay_mod}.

\subsubsection{Mineflayer Viewer}

\textbf{Mineflayer prismarine-viewer}
is a tool proposed by PrismarineJS team. This tool provides Viewer and WorldView which make it possible to render a Minecraft world. 
This tool can be well integrated with Mineflayer~\cite{mineflayer} simulator environment to quickly generate the corresponding visual views from the agent's current perspective.
The Github repository of this tool is at \url{https://prismarinejs.github.io/prismarine-viewer/}.

Our initial plan for using this tool is as follows: We set up an observing agent (named {\it bot}) to observe the constructed objects from a specific position and at a specific angle, and then use this tool to take screenshots to obtain images from the corresponding perspectives and positions. This method has almost no delay and can be easily automated, making it highly suitable for evaluating agents.

\begin{figure*}[h]
\vspace{-0.2cm}
\centering
\includegraphics[width=0.65\linewidth]{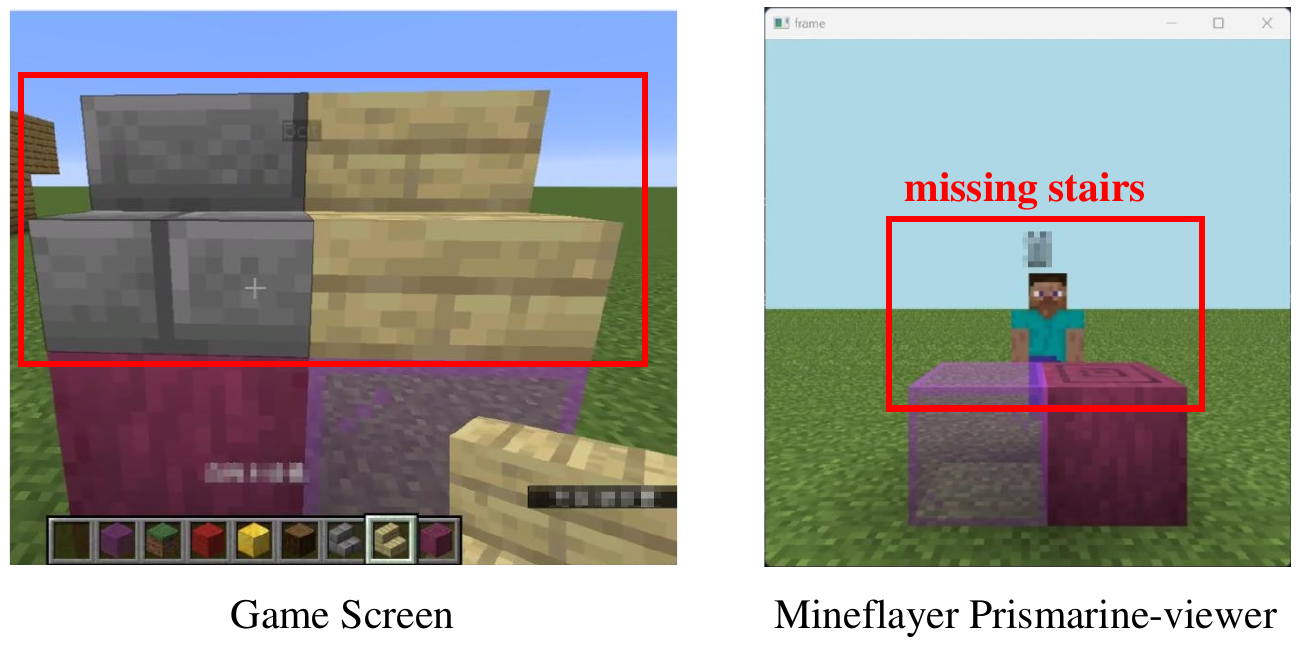}
\caption{Render error in Mineflayer prismarine-viewer.}
\label{fig:viewer_error}
\vspace{-0.2cm}
\end{figure*}

However, we found that prismarine-viewer in Minecraft 1.16.5 and 1.20.4 can not render some important blocks that are frequently appeared in building tasks, such as {\it stairs} and {\it slabs}, as shown in Figure~\ref{fig:viewer_error}. 
This poses a significant problem for building data and tasks involving structures such as houses, as it may result in these houses lacking well-designed roofs. 
Voyager~\cite{wang2023voyager} has answered an issue\footnote{https://github.com/MineDojo/Voyager/issues/19} that they finally didn't choose this tool due to its poor-quality visualization.
Therefore, we have to temporarily abandon this tool, and hoping that this tool can further address this rendering issue in the future.

While writing this Supplementary Material, we find that the project seems to be able to successfully render stairs in the latest version of Minecraft 1.21.4 (at \url{https://github.com/PrismarineJS/prismarine-viewer}). Therefore, we will update our benchmark in our GitHub code repository in the future to adapt to these latest versions, which will facilitate the further evaluation of AI agents.





\subsubsection{Replay Mod}
\label{replay_mod}
\textbf{Replay Mod} is a mod for the popular sandbox game Minecraft which allows players to record, replay and share their gaming experience. It's easy to use, but an incredibly powerful tool to create perfect Minecraft videos within minutes. The website of this tool is at \url{https://www.replaymod.com/}.

This evaluation tool can correctly render the blocks (e.g., {\it stairs} and {\it slabs}) that appear in architectural structures, and the quality of the output images/videos is relatively high. Although this tool requires some manual operations, these operations are relatively easy. Moreover, the generated images are less likely to cause errors that would lead to incorrect judgments of the agent's output by the critic model. 

\begin{figure*}[h]
\vspace{-0.2cm}
\centering
\includegraphics[width=0.95\linewidth]{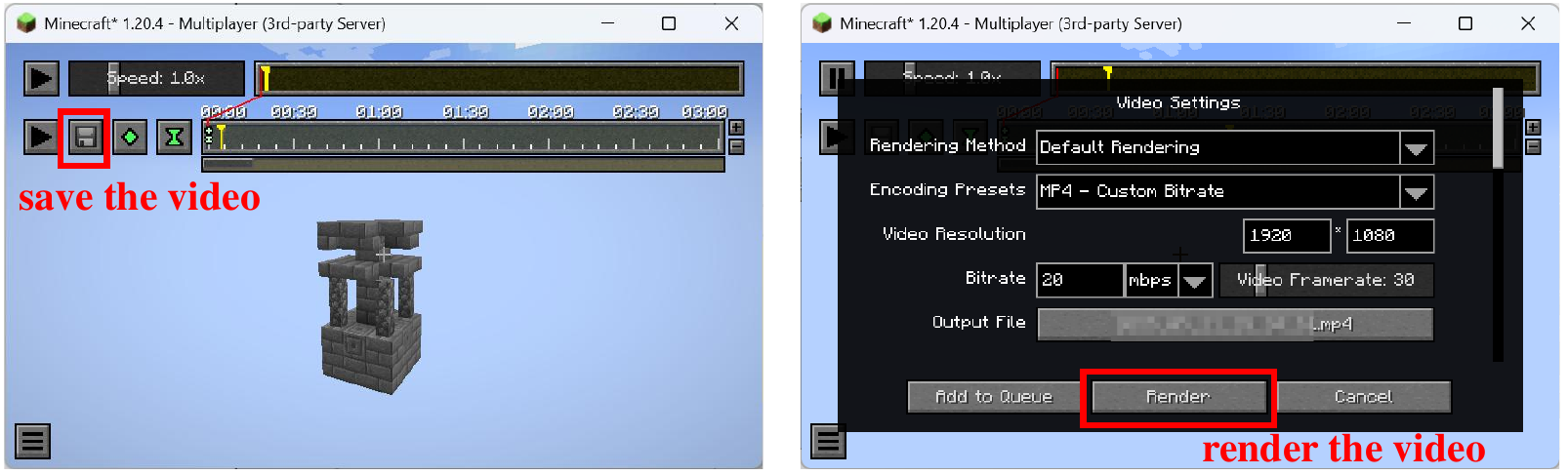}
\caption{Visualization of Replay Mod.}
\label{fig:replaymod}
\vspace{-0.2cm}
\end{figure*}

Therefore, we finally selected this tool as our evaluation tool.
We use this tool to obtain visualized videos, and extract frames from the videos through the \textit{opencv-python} library to select the corresponding images as the input for the critic models, as shown in Figure~\ref{fig:replaymod}. Specific operation tutorials can be referred to at \url{https://www.replaymod.com/docs/}. We will also provide corresponding reference documents and video processing code segments using \textit{opencv-python} in our Github code repository in the future.


\section{Datasheets}
\label{sec_C}
\subsection{Dataset Description}
Our dataset is built on several player-generated content (PGC) on the Internet. We conduct data curation process on this PGC and design various tasks based on the curated data. 
We recommend that any researchers using our dataset should use it for non-commercial AI research purposes only.
We have released our dataset on Hugging Face: \url{https://huggingface.co/datasets/SaDil/MineAnyBuild}.

\subsection{License}
Our MineAnyBuild benchmark is released under the 
\href{https://creativecommons.org/licenses/by-nc-sa/4.0/}{CC BY-NC-SA 4.0} license.
We also list the relevant licenses and terms-of-use of data resources utilized in the process of constructing our dataset as follow:

\textbf{(1) GrabCraft}~\cite{grabcraft}: \href{https://creativecommons.org/licenses/by-nc-sa/4.0/}{CC BY-NC-SA 4.0}
(\href{https://www.grabcraft.com/terms-of-use-privacy-policy/}{Terms of Use}).

\textbf{(2) Minecraft Official Wiki}~\cite{fandom_wiki}: \href{https://creativecommons.org/licenses/by-nc-sa/4.0/}{CC BY-NC-SA 4.0}
(\href{https://www.fandom.com/terms-of-use}{Terms of Use}).

\textbf{(3) Creations from Raekon}:
\href{https://creativecommons.org/licenses/by-nc-sa/4.0/}{CC BY-NC-SA 4.0}
(\href{https://www.patreon.com/policy/legal}{Terms of Use}).
We may partially release our dataset to protect the copyright of this author, Raekon. Anyone who use this dataset can adopt our labeled data or our data curation process to get the same data. \textbf{Commercial use is strictly forbidden}, and we recommend you to subscribe to this author and download the corresponding map environments to obtain the same dataset as ours. Author homepage: \url{https://www.patreon.com/Raekon}.


\subsection{Format}
The folder structure of the \textit{MineAnyBuild.zip} file we provide on Hugging Face (\url{https://huggingface.co/datasets/SaDil/MineAnyBuild}) is organized as follow:


\dirtree{%
.1 MineAnyBuild.zip.
.2 data.
.3 images.
.4 assets\DTcomment{{\it Images of decoration assets from Raekon}}.
.5 1.jpg.
.5 ....
.4 commonsense\DTcomment{{\it Images for Spatial Commonsense Task}}.
.5 SC\_0001\_IMG\_0.jpg.
.5 ....
.4 grabcraft\DTcomment{{\it Images of architectures from GrabCraft}}.
.5 al\_capone\_cadillac.png.
.5 ....
.4 reasoning\DTcomment{{\it Images for Spatial Reasoning Task}}.
.5 TSK\_SR\_1\_0001.jpg.
.5 ....
.4 stimuli\DTcomment{{\it Images of 48 stimuli}}.
.5 MR1\DTcomment{{\it Stimulus No.1}}.
.6 N\_0.png\DTcomment{{\it Image of original stimulus in default +45 viewing angle}}.
.6 RX\_1.png\DTcomment{{\it Image of \textbf{x-axis-mirrored} stimulus in +135 viewing angle}}.
.6 RY\_2.png\DTcomment{{\it Image of \textbf{y-axis-mirrored} stimulus in +225 viewing angle}}.
.6 RZ\_3.png\DTcomment{{\it Image of \textbf{z-axis-mirrored} stimulus in +315 viewing angle}}.
.6 ....
.5 ....
.4 wiki\DTcomment{{\it Images of architectures from Minecraft Official Wiki}}.
.5 desert\_animal\_pen\_1.png.
.5 ....
.3 architectures.json\DTcomment{{\it Data of architectures and decoration assets}}.
.2 task\DTcomment{{\it Data of 5 types of tasks}}.
.3 Task\_Creativity.json.
.3 Task\_Spatial\_Commonsense.json.
.3 Task\_Spatial\_Planning.json.
.3 Task\_Spatial\_Reasoning.json.
.3 Task\_Spatial\_Understanding.json.
}

\subsection{Data Field Descriptions}
Our dataset mainly consists of two major parts: architectures and tasks.
``Architectures'' refer to the buildings/decoration assets collected from the above-mentioned data resources, while ``tasks'' refer to five types of tasks that we designed for these ``Architectures''.
Each row in architectures and tasks data is organized as a structured JSON format.
We provide the specific descriptions for each data field utilized in our dataset as follow. 

\subsubsection{Architectures}
The data field descriptions of architectures data are separated into two parts: \textsc{Global} for field common to all data and \textsc{Optional} for field only available in partial data.

\textsc{Global}:
\begin{itemize}
    \item {\it \textbf{id}}: A unique identifier for an architecture. This identifier follows the format of \{AR\_TDDDD\_{\it hash1}\_{\it hash2}\}, where T indicates the types of data resources, and DDDD indicates the unique number of the architecture. {\it hash1} and {\it hash2} represent the first 16 hexadecimal digits of the hash values generated by the SHA-256 algorithm for information such as {\it \textbf{types}}, {\it \textbf{name}} and {\it \textbf{description}}. These hash strings are utilized as unique identifiers for an architecture/asset.
    (e.g.: AR\_00001\_4025eb7e4ff6ef93\_a9eab824ea6e85d9)
    \item {\it \textbf{name}}: A simple name of this architecture. For decoration assets, we set it as the format of ``decoration\_asset\_\{num\}''.
    \item {\it \textbf{description}}: A textual description for the architecture or asset.
    \item {\it \textbf{data\_resource}}: An identifier selected from ``Grabcraft'', ``Minecraft Official Wiki'' and ``Raekon''. 
    \item {\it \textbf{3d\_info}}: A dictionary that describes the 3D data of a cubic bounding box enclosing an entire 3D architectures/assets, e.g., \{``width'': 5, ``height'': 5, ``depth'': 5\}.
    \item {\it \textbf{difficulty\_factor}}: A difficulty coefficient obtained through comprehensive calculation based on the data distribution (introduced in Section~\ref{diff_factor}).
    \item {\it \textbf{image}}: A relative file address of a visualization image of the architecture/decoration asset.
    \item {\it \textbf{block\_materials}}: A list of all types of blocks utilized in the architecture/decoration asset. The list will be constructed into a mapping table (dictionary) during the construction process, using to associate {\it \textbf{block\_materials}} with the {\it \textbf{blueprint}} matrix.
    \item {\it \textbf{blueprint}}: A 3-dimension list (matrix) representing architectural construction blueprint through integers. Each integer corresponds to the index number of the {\it\textbf{block\_materials}} list plus 1. (e.g., for the list [A, B, C], the integer 2 corresponds to the block B.)

\end{itemize}

\textsc{Optional}:
\begin{itemize}
    \item {\it \textbf{type}}: A field used to describe architecture types, which is not utilized to describe the indoor decoration assets. (e.g., the architecture ``simple\_tree\_house'' belongs to the type of ``houses'')
    \item {\it \textbf{biome}}: A field exclusive to ``Minecraft Official Wiki'' data. Since these data come from the official resources, the Minecraft game has different construction plans and block types for similar building in different biomes (i.e., Minecraft natural environments).
    \item {\it \textbf{image\_urls}}: A field exclusive to ``Grabcraft'', presenting the image urls that can be downloaded by http/https requests.
    \item {\it \textbf{metadata}}: A field to describe the data sources, which can be data urls or architectural metadata.
\end{itemize}

\subsubsection{Tasks}
The data field descriptions of tasks data are separated into two parts: \textsc{Global} for field common to all data and \textsc{Optional} for field only available in partial data.

\textsc{Global}:
\begin{itemize}
    \item {\it \textbf{id}}: A unique identifier for a task. This identifier follows the format of \{TSK\_{\it tasktype}\_{\it uniquestring}\}.
    {\it tasktype} is selected from ``CR'', ``SC'', ``SP'', ``SR'' and ``SU'', representing ``Creativity'', ``Spatial Commonsense'', ``Executable Spatial Plan Generation'', ``Spatial Reasoning'' and ``Spatial Understanding'', respectively. {\it uniquestring} is the unique string of different tasks.
    (e.g.: TSK\_SR\_1\_0084)
    \item {\it \textbf{instruction}}: An instruction for building architectures/assets or a question for Spatial Reasoning task and Spatial Commonsense task.
\end{itemize}

\textsc{Optional}:
\begin{itemize}
    \item {\it \textbf{AR\_id}}: An identifier that is the same as {\it \textbf{AR\_id}} in Architecture data. This field is only set for Executable Spatial Plan Generation, Creativity and Spatial Understanding tasks.
    \item {\it \textbf{difficulty\_factor}}: A float value that is the same as {\it \textbf{difficulty\_factor}} in Architecture data. This field is only set for Executable Spatial Plan Generation, Creativity and Spatial Understanding tasks.
    \item {\it \textbf{block\_materials}}: A list that is the same as {\it \textbf{block\_materials}} in Architecture data. This field is only set for Executable Spatial Plan Generation, Creativity and Spatial Understanding tasks.
    \item {\it \textbf{image}}: One or more images that visualize the architecture/asset or perspective views.
    This field is only set for Executable Spatial Plan Generation, Spatial Understanding and Spatial Commonsense tasks.
    \item {\it \textbf{image\_desp}}: One or more textual descriptions of the {\it \textbf{image}}. This field is only set for Spatial Commonsense task.
    \item {\it \textbf{options\_image}}: An image containing a combination of 4 option images or 1 compared image, and an original image.
    This field is only set for Spatial Reasoning task.
    \item {\it \textbf{options}}: A list consisting of 4 option images.
    This field is only set for Spatial Reasoning task.
    \item {\it \textbf{metadata}}: A string that describes the option and selected stimuli.
    This field is only set for Spatial Reasoning task.
\end{itemize}

\subsection{Data Examples}
In this subsection, we provide more visualization examples in our datasets.

\subsubsection{Architectures}
\begin{figure*}[h]
\vspace{-0.2cm}
\centering
\includegraphics[width=\linewidth]{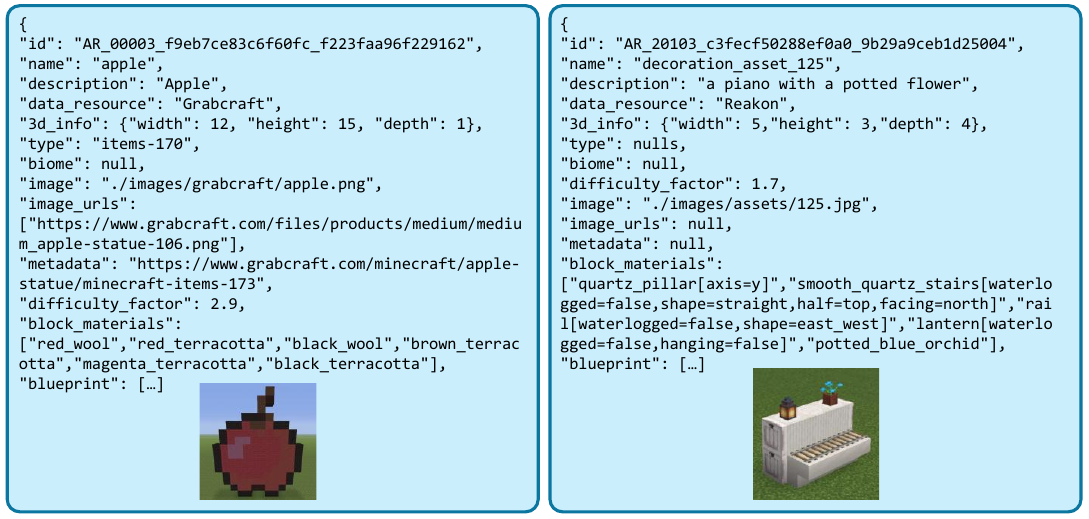}
\caption{Visualization examples of architecture data.}
\label{fig:arch task example}
\vspace{-0.2cm}
\end{figure*}

We provide two visualization examples of architecture data in Figure~\ref{fig:arch task example}.

\subsubsection{Tasks}
\begin{figure*}[h!]
\vspace{-0.2cm}
\centering
\includegraphics[width=0.9\linewidth]{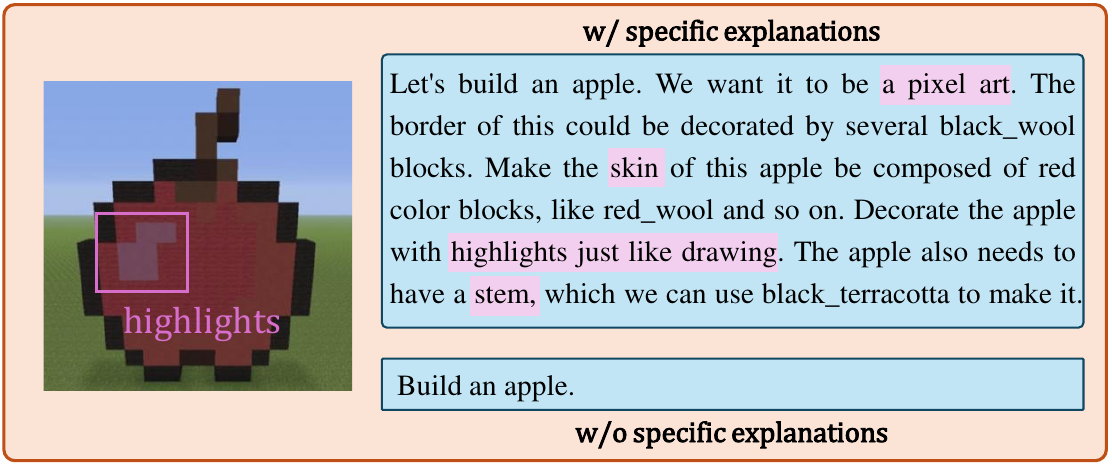}
\caption{Task example visualization of Executable Spatial Plan Generation task.}
\label{fig:task_example_spatial_plan}
\vspace{-0.2cm}
\end{figure*}

\begin{figure*}[h!]
\vspace{-0.2cm}
\centering
\includegraphics[width=0.85\linewidth]{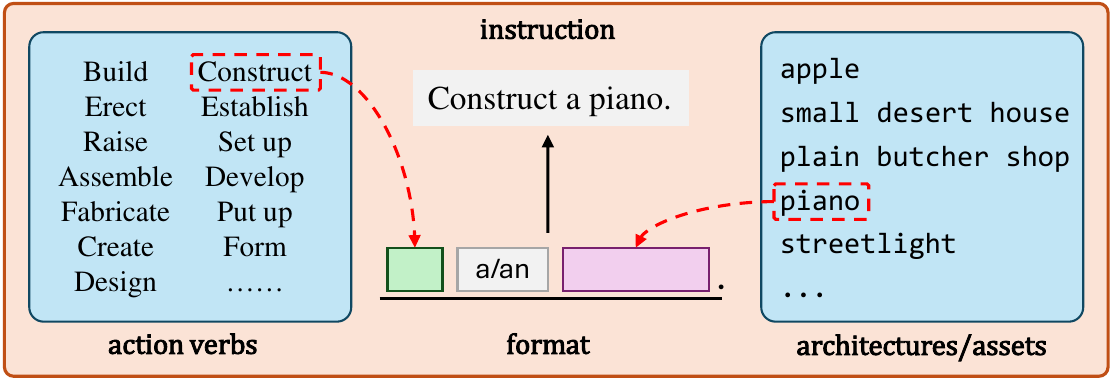}
\caption{Task example visualization of Creativity task.}
\label{fig:task_example_creativity}
\vspace{-0.2cm}
\end{figure*}

\begin{figure*}[h!]
\vspace{-0.1cm}
\centering
\includegraphics[width=0.85\linewidth]{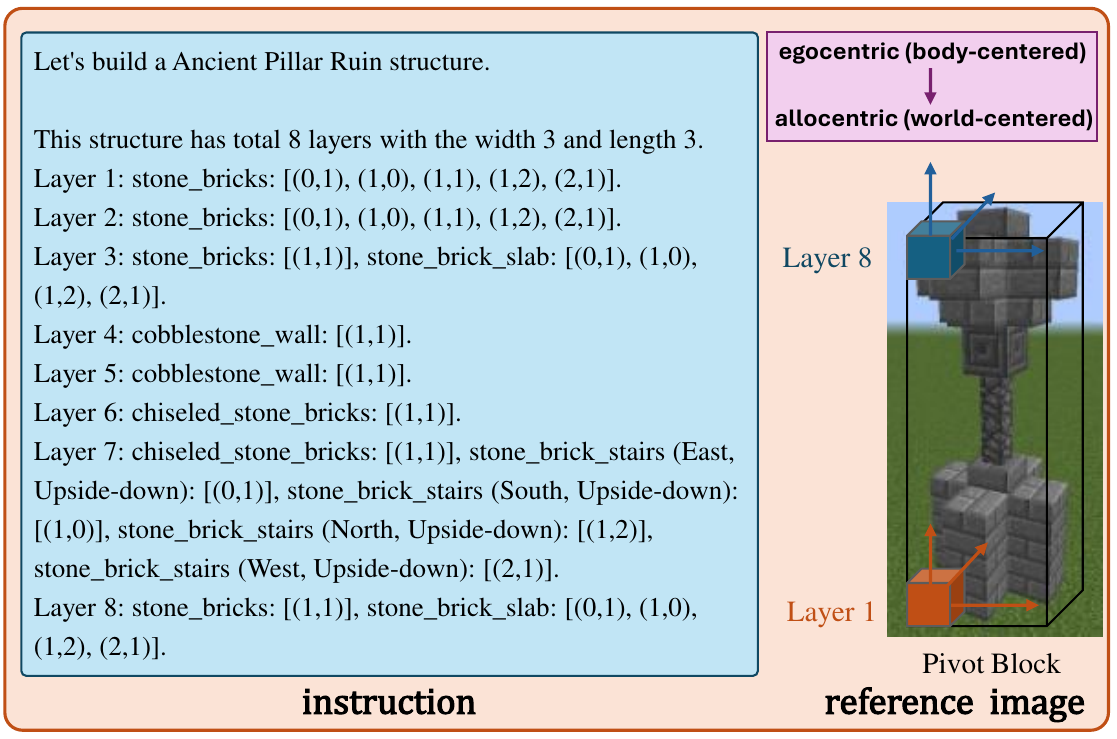}
\caption{Task example visualization of Spatial Understanding task.}
\label{fig:task_example_spatial_ud}
\vspace{-0.1cm}
\end{figure*}

\begin{figure*}[h!]
\vspace{-0.1cm}
\centering
\includegraphics[width=0.8\linewidth]{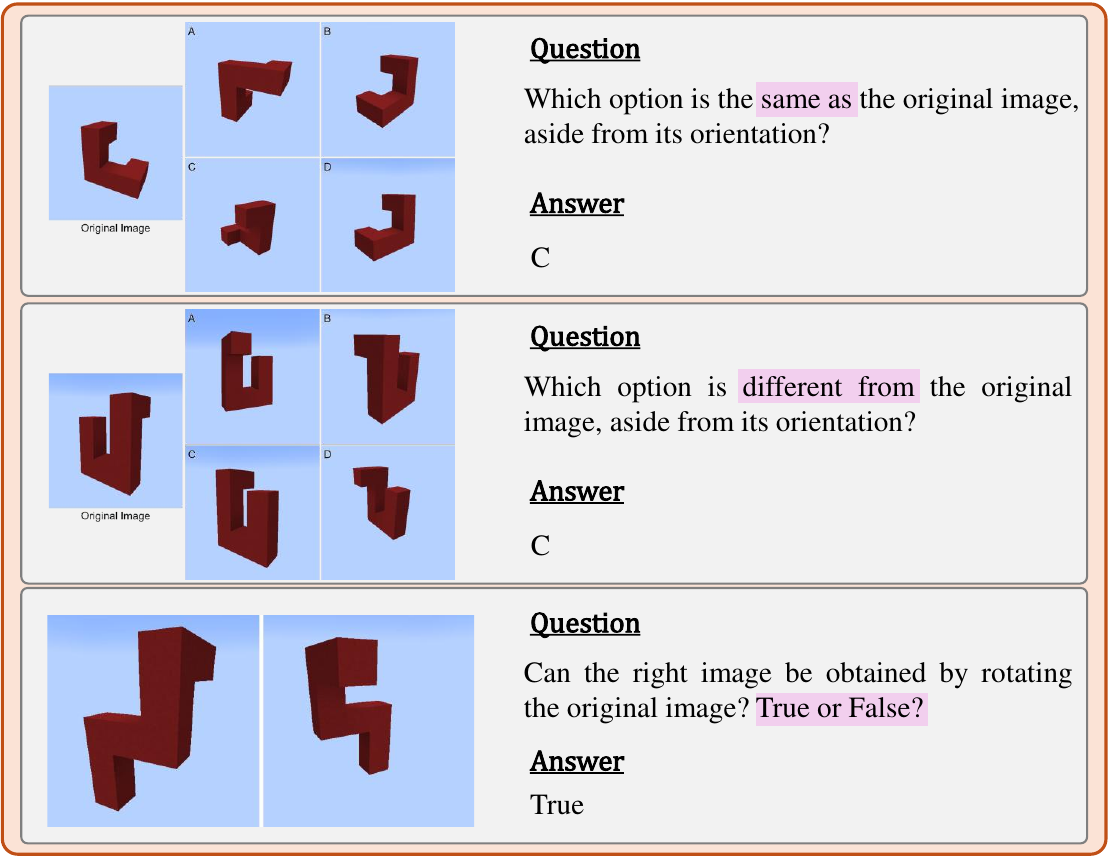}
\caption{Task example visualization of Spatial Reasoning task.}
\label{fig:task_example_spatial_reason}
\vspace{-0.1cm}
\end{figure*}

\begin{figure*}[h!]
\vspace{-0.2cm}
\centering
\includegraphics[width=0.8\linewidth]{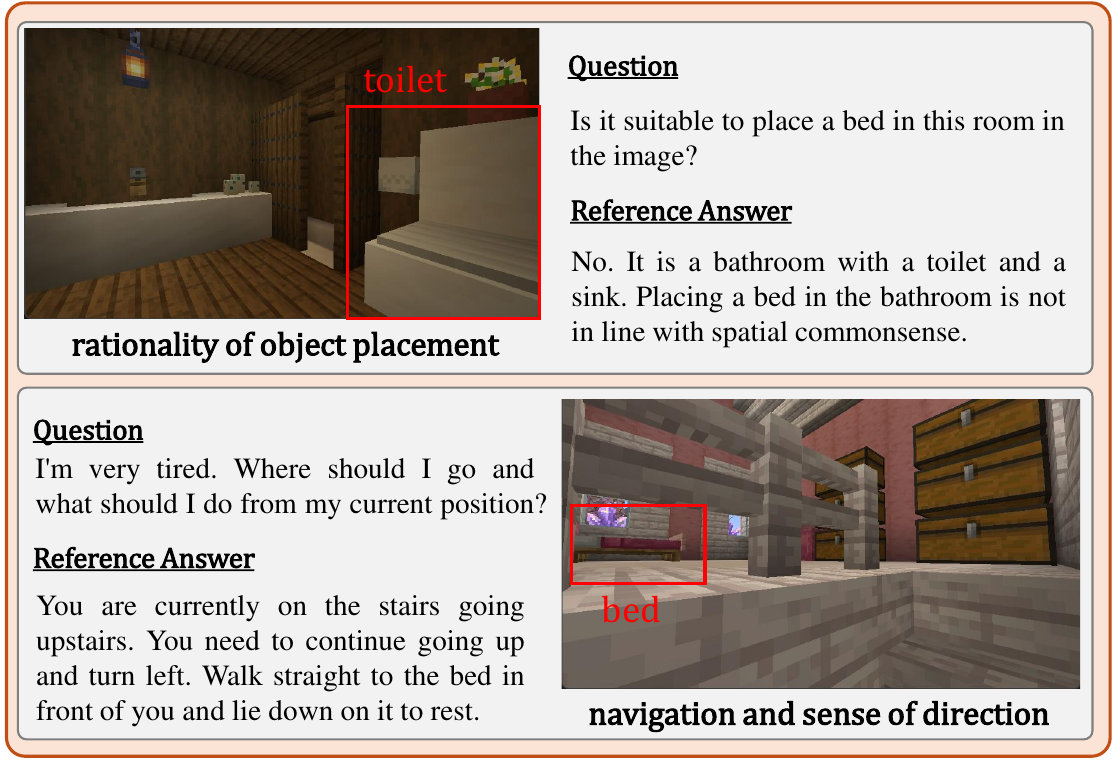}
\caption{Task example visualization of Spatial Commonsense task.}
\label{fig:task_example_spatial_common}
\vspace{-0.2cm}
\end{figure*}

We provide more visualization examples for the Executable Spatial Plan Generation, Spatial Understanding, Creativity, Spatial Reasoning, and Spatial Commonsense tasks, which are presented in Figure~\ref{fig:task_example_spatial_plan}-\ref{fig:task_example_spatial_common}.  

\subsection{Croissant Metadata}
The Croissant file for our MineAnyBuild hosted on Hugging Face is available at \url{https://huggingface.co/api/datasets/SaDil/MineAnyBuild/croissant}.
The file has successfully passed the the croissant-checker\footnote{https://huggingface.co/spaces/JoaquinVanschoren/croissant-checker}.

\section{Details for Benchmark, Tasks and Data Curation}
\label{sec_D}
In this section, we provide some details about our benchmark MineAnyBuild, corresponding tasks and data curation pipeline.

\subsection{Details of Dataset}

Our MineAnyBuild has 4000+ curated tasks covering several critical dimensions for evaluation of AI agents. We have curated 483 diverse architectures/assets from thousands of data candidates for planning tasks. We also collected 10+ large scenes for Spatial Commonsense task and generated 192 stimuli from the original datasets. The quantity and diversity of our datasets can also be infinitely expanded and scaled through our data curation pipeline. The numbers of different tasks in current MineAnyBuild datasets are shown in Table~\ref{tab:task_numbers}. 
There are 22 building types in our proposed dataset, including houses, fictional characters, items, etc., shown in Table~\ref{tab:task_types}.

\begin{table}[h]
\centering
	\caption{The numbers of different tasks in current MineAnyBuild dataset.
	}\label{tab:task_numbers}
		\resizebox{\linewidth}{!}{
	{\renewcommand{\arraystretch}{1.5}	\begin{tabular}{c||ccccc}
				\specialrule{.1em}{.05em}{.05em}	
    Task&Executable Spatial Plan Generation&Spatial Understanding&Spatial Reasoning&Creativity&Spatial Commonsense\\ \hline

Numbers&946&473&1728&1419&50\\
    
	\specialrule{.1em}{.05em}{.05em}	\end{tabular}}}
\vspace{-0.2cm}	
\end{table}

\begin{table}[h]
\centering
	\caption{Building types in current MineAnyBuild datasets.
	}\label{tab:task_types}
		\resizebox{\linewidth}{!}{
	{\renewcommand{\arraystretch}{1.5}	\begin{tabular}{c||l}
				\specialrule{.1em}{.05em}{.05em}	
    Types&\makecell[l]{houses, animals, items, town\_centers, fictional\_characters, cars, military\_buildings, buses,\\emergency\_vehicles, miscellaneous, famous\_films, working\_vehicles, bridges, garden, \\parks, ruins, ships, sightseeing\_buildings, planes, minecraft\_villages, lamps, others}\\
	\specialrule{.1em}{.05em}{.05em}	\end{tabular}}}
\vspace{-0.2cm}	
\end{table}

\subsection{Details of Tasks}

\subsubsection{More Details of Executable Spatial Plan Generation Task}
As we mentioned in the main text, the task input we designed consists of an abstract architecture building instruction accompanied by precise explanations. Agents are required to think about the decomposition of architectural substructures and their corresponding connections to generate executable spatial plans for architecture building, just like completing a jigsaw puzzle.

We also provide Executable Spatial Plan Generation tasks with simpler instructions. The instructions for this task type are consistent with those in Creativity task, only describing the entire building. We require the agent to output its planning based on the visual input and these simple instructions, and then provide the corresponding executable blueprint matrix according to the planning. 

For tasks with given detailed instructions, we provide specific explanations based on the supervision by powerful MLLMs or human, guiding the agent to gradually construct these sub-structures according to the step-by-step explanations. While for simple instruction tasks without these detailed guidance, 
we require agents to output the process of decomposing substructures. In fact, these two types of tasks complement each other rather than being completely overlapping. Overall, they can effectively evaluate the agent's comprehensive capability for spatial planning. We also provide examples of these two types of Executable Spatial Plan Generation task data in Figure \ref{fig:task_example_spatial_plan} for differentiation.


%

\subsubsection{More Details of Creativity Task}
As we mentioned in the Creativity task prompt in Section ~\ref{task_prompts}, we guide MLLM-based agents on how to construct creative buildings. Specifically, the key points are as follows:
\begin{itemize}
    \item[1)] How to best interpret and implement the build specification.\\
    - List key elements from the build specification.\\
    - Brainstorm block combinations for different parts of the structure.\\
    - Outline a rough structure layout.
    \item[2)] Creative ways to use the available blocks to achieve the desired aesthetic.
    \item[3)] How to ensure the mapping correctness in your blueprint matrix.
    \item[4)] Ways to maximize creativity and the dynamic range of possible builds.
    \item[5)] Consider potential challenges and solutions.
\end{itemize}

In this way, we can guide agents to plan in a more creative direction, thereby stimulating their creative capabilities with specific prompts and presenting more distinctive architectural creations.


\subsubsection{More Details of Spatial Understanding Task}
The instructions for the Spatial Understanding task can be automatically generated through templated code, i.e., we only need to associate the values of the matrix with the corresponding indices based on the existing ground-truth blueprint matrix and block materials, and set the block at [0][0][0] as the pivot block.
Then we can obtain the corresponding task instruction data represented by relative positions.

It seems that this task is obvious and easy for the agent, but in fact it is not. During our conducted experiments, we found that the agents' understanding of 3D data is still weak, and they have difficulties in handling the conversion between relative coordinates and world coordinates in practical applications. Therefore, we still keep this task to evaluate the agent's perspective transformation capability, as shown in Figure~\ref{fig:task_example_spatial_ud}.

The main difference between this task and the Executable Spatial Plan Generation task is that the former focuses more on the ``result'' of coordinate perspective transformation, while the latter emphasizes more on the ``process'' of spatial planning. The two tasks are not exactly the same but ultimately complementary.


\subsubsection{More Details of Spatial Reasoning Task}
As shown in Figure~\ref{fig:task_example_spatial_reason}, there are three main types of our Spatial Reasoning tasks in total, where selecting the same one from four stimuli, selecting the different one from four stimuli, and judging whether two stimuli can be obtained only by rotation.


\begin{figure*}[h!]
\vspace{-0.1cm}
\centering
\includegraphics[width=0.75\linewidth]{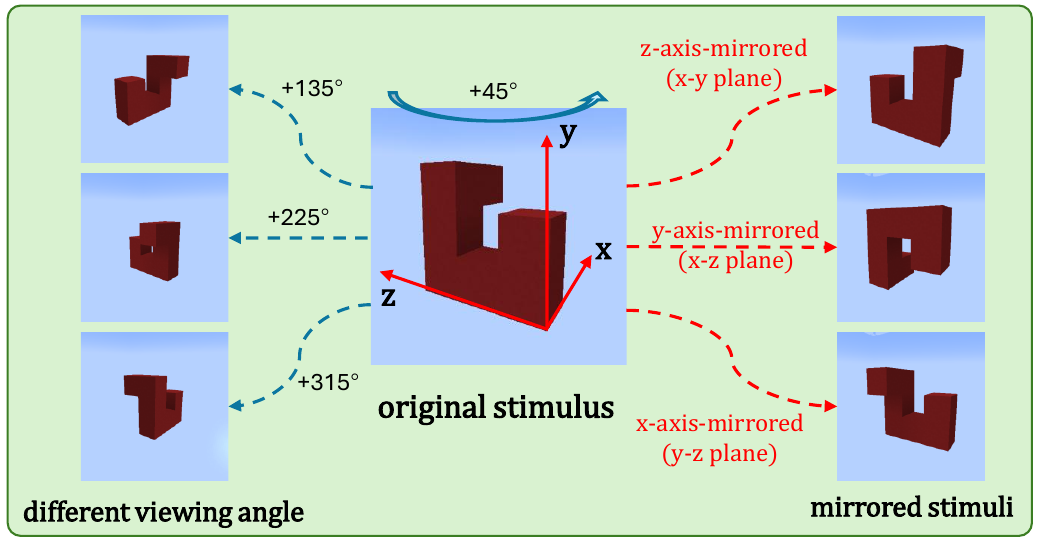}
\caption{Augmentation of stimuli utilized in Spatial Reasoning task.}
\label{fig:task_example_spatial_reason_aug}
\vspace{-0.1cm}
\end{figure*}

In Figure~\ref{fig:task_example_spatial_reason_aug}, we demonstrate how we expand from one stimulus to several stimuli. We mainly focus on randomly generated stimuli in the Minecraft environment and perform mirror symmetry along the X/Y/Z axes. The chiral geometries produced in this way usually cannot be made identical to the original ones just by rotation, unless the original stimulus itself has a certain degree of symmetry, but we require and avoid such geometries with obvious symmetry.

\subsubsection{More Details of Spatial Commonsense Task}
\begin{figure*}[h!]
\vspace{-0.2cm}
\centering
\includegraphics[width=0.7\linewidth]{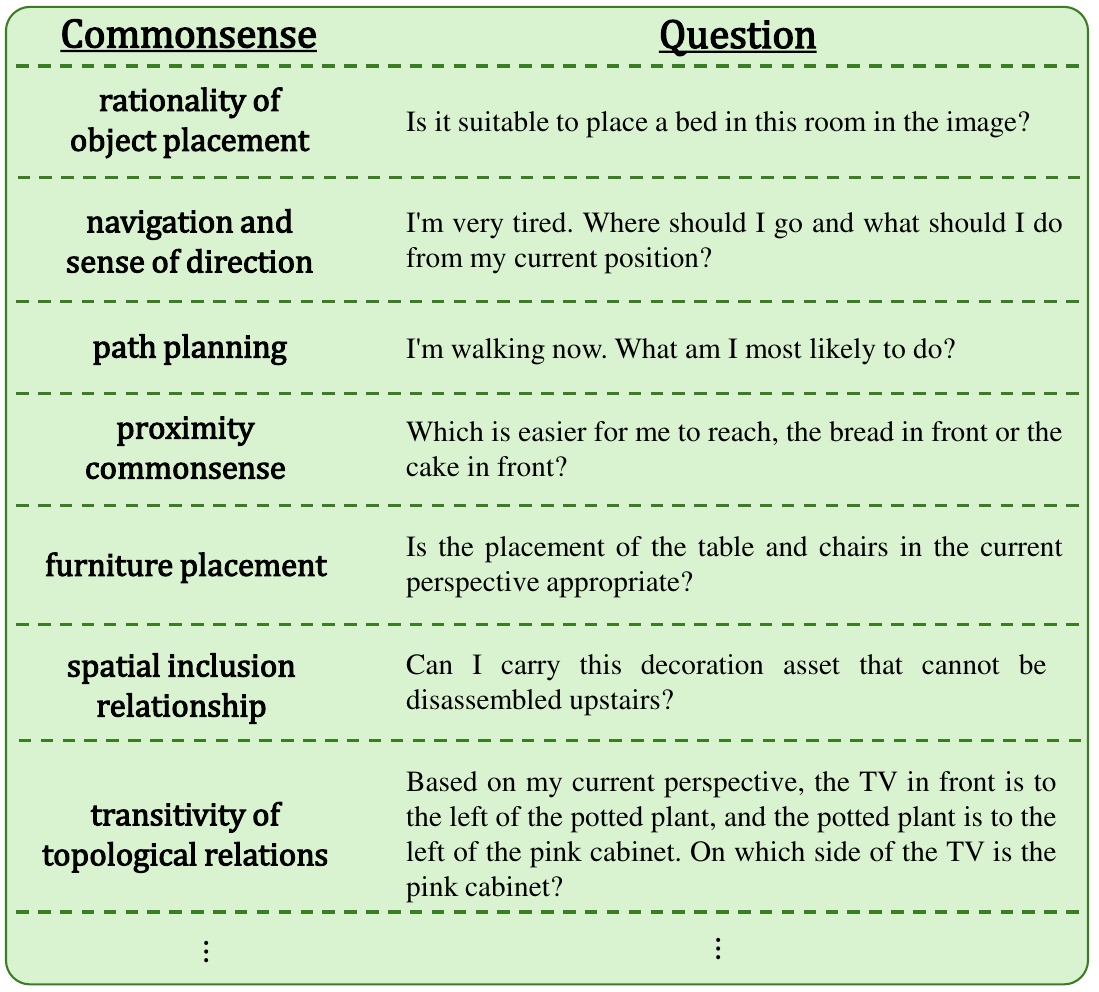}
\caption{Vital types of spatial commonsense in our task data.}
\label{fig:details_spatial_commonsense}
\vspace{-0.2cm}
\end{figure*}

Spatial Commonsense refers to humans' intuitive understanding of spatial attributes such as the position, direction, distance, and shape of objects in the physical world, and it appears in every aspect of daily life.
We list the most important types of spatial commonsense in our task data in Figure~\ref{fig:details_spatial_commonsense}, and provide explanations as follows:

\begin{itemize}
    \item \textbf{Rationality of object placement:}
    To estimate a contradiction between asset function and space and  whether the size of an item is suitable for a certain space (e.g., ``a refrigerator should not be placed in the bathroom'').

    \item \textbf{Navigation and
sense of direction:}
    We require agents to locate directions without previous information, and determine abnormal situations (e.g., ``you need to turn right when going upstairs'' and ``the chair can not be under the table''). 

    \item \textbf{Path planning:}
    We require agents to avoid obstacles and bypass them rather than attempting to pass through a wall when conducting path planning.

    \item \textbf{Proximity commonsense:}
    We require agents to make judgments based on spatial commonsense and intuition rather than precise measurement, just as humans judge distance.

    \item \textbf{Furniture placement:}
    For some common combinations of spatial object relationships, such as the orientation and position of table and chair arrangements, we request agents to make correct judgments. 

    \item \textbf{Spatial inclusion relationship:}
    If object A is completely inside of container B, then the volume of A must be smaller than the internal space of B (e.g., ``a piano cannot fit into a chest''). 

    \item \textbf{Transitivity of
topological relations:}
    If A is to the left of B and B is to the left of C, then A is usually to the left of C. However, agents need to make judgments and verifications based on actual visual observations. Humans can determine this transitivity without logical training, and we evaluate whether the agents can achieve it.
    
\end{itemize}

\subsection{Details of Data Curation Pipeline}

\subsubsection{Details About Infinitely Expandable Paradigm}
As we mentioned in main text, we conduct our data curation pipeline based on three steps: 1) data collection, 2) quality checking, and 3) data annotation. We also propose an infinitely expandable paradigm for our data curation pipeline.  

\begin{figure*}[h!]
\vspace{-0.2cm}
\centering
\includegraphics[width=\linewidth]{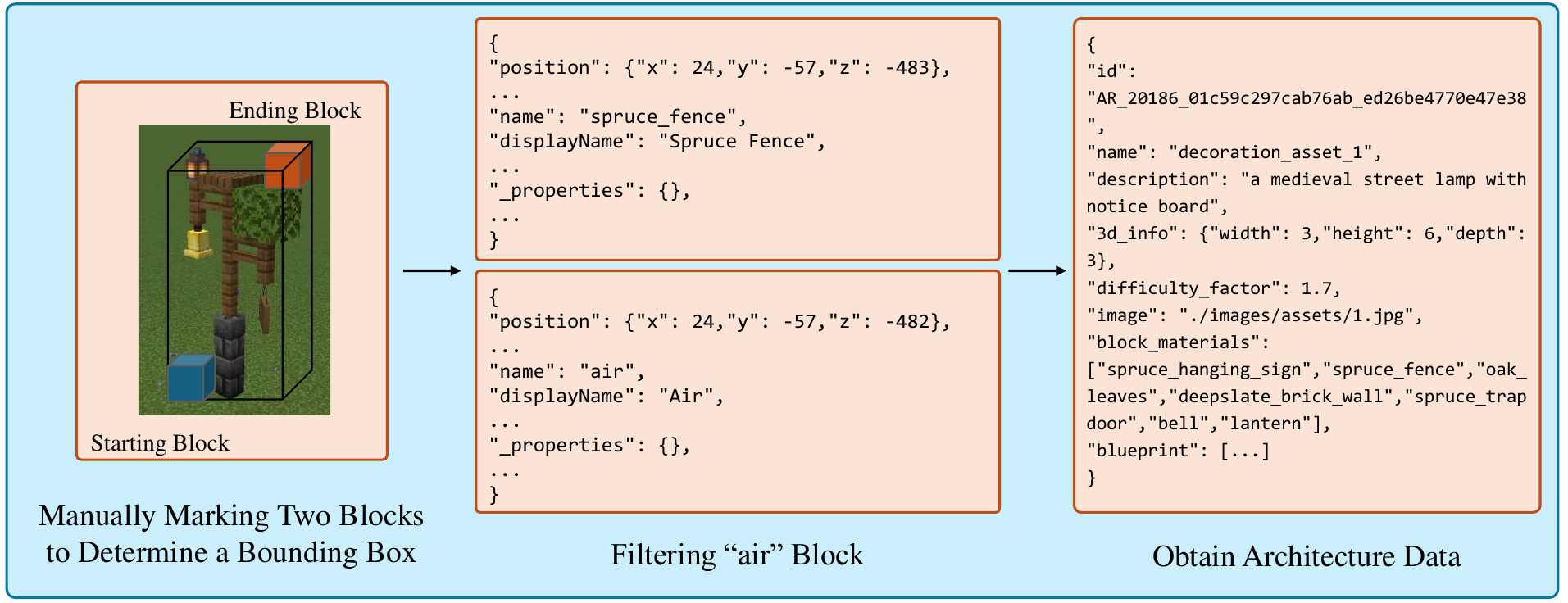}
\caption{A simplified example of data curation pipeline.}
\label{fig:data_curation_pipe}
\vspace{-0.2cm}
\end{figure*}

Specifically,
we manually mark the starting block (the minimum values on the X/Y/Z coordinates) and the ending block (the maximum values on the X/Y/Z coordinates) of the 3D coordinates as the three-dimensional coordinate box of the entire building, and obtain all the block information corresponding to each position through \textit{mineflayer} simulator~\cite{mineflayer}.
After filtering the \textit{``air''} blocks, 
corresponding \textit{three\_d\_info}, \textit{blueprint} and \textit{block\_materials} can be acquired to obtain the architecture data. We visualize a simplified version of this pipeline in Figure~\ref{fig:data_curation_pipe}.

\subsubsection{Estimation Metric of Difficulty Factor}
\label{diff_factor}

To measure the construction difficulty for user-generated structures in Minecraft environment, we propose a novel estimation metric. This metric is designed to combine the perceived effort with complexity in the building process by integrating key architectural parameters. Its formulation is based on the following key considerations:

\begin{figure}[h]
    \centering
\includegraphics[width=\linewidth]{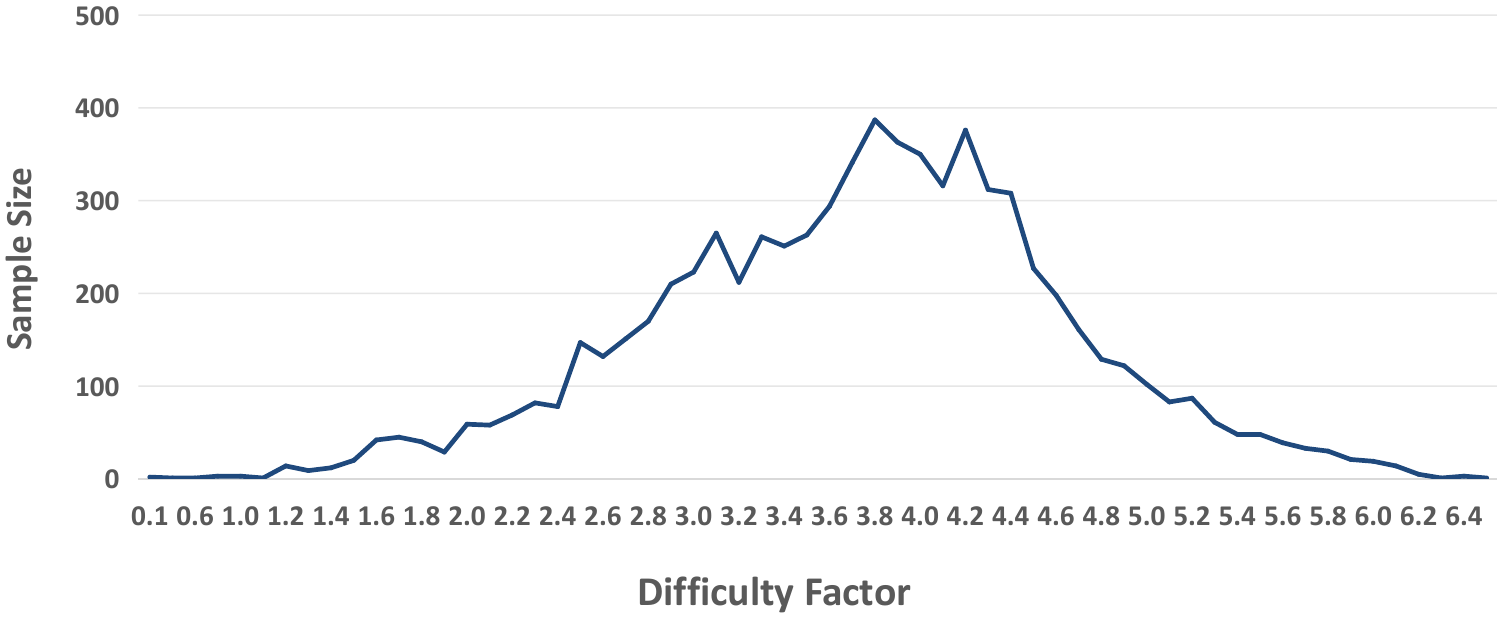}
    \caption{Difficulty Factor Distribution of task samples.}
    \label{fig:diff_factor}
\end{figure}

\begin{itemize}
    \item \textbf{Base Workload}: The foundational construction effort is primarily determined by the total number of blocks ($N$) required for the structure. A larger $N$ typically corresponds to increased labor.
    \item \textbf{Architectural Scale and Volume Complexity}: The spatial extent of a structure, defined by its bounding box dimensions of length ($L$), width ($W$), and height ($H$), significantly influences overall complexity. Larger volumes ($L \cdot W \cdot H$) typically require more intricate planning,
    and attention to balance symmetry and structural integrity.
    \item \textbf{Height Challenges}: Building upward has extra difficulties compared to building outward. Considering in the same way as the commonsense of real-world architecture, factors such as the need for scaffolding, higher risk of falls, and the effort of moving materials vertically. This is captured with an interaction between the total block count ($N$) and the height ($H$), since placing blocks at greater heights is naturally harder.
    
\end{itemize}

Based on these factors, we define the difficulty estimation metric, $D$, as follows:

\[
D = \ln\left(k_1 N + k_2 N H + k_3 L W H \right) - B
\]

Where:

\begin{itemize}
    \item $D$ is the estimated difficulty score, a relative scalar value where higher scores suggest greater theoretical construction complexity.
    \item $N$ represents the total 
    count of blocks used in the construction.
    \item $L$, $W$, and $H$ denote the 
    values for length, width, and height of the structure's bounding box, respectively.
    \item $k_1$, $k_2$, and $k_3$ are 
    weighting coefficients associated with the base workload, height-augmented workload, and volumetric complexity, respectively.
    \item $B$ is a 
    bias term used for calibration.
\end{itemize}

In our initial tests, the coefficients $k_1$, $k_2$, and $k_3$ were all set to 1 (i.e., $k_1 = k_2 = k_3 = 1$), and the bias term $B$ was set to 0.4. Early evaluations on a dataset of Minecraft constructions indicate that the resulting difficulty scores approximate a Gaussian (bell-shaped) distribution. This suggests that the metric can provide a solid basis for initial classification and comparison of construction difficulty.

Furthermore, the use of a logarithmic function, rather than mapping difficulty scores to a predefined fixed range, provides robust scalability to the metric. 
This makes it especially useful for new datasets, as it allows consistent evaluation across different levels of construction complexity and avoids problems like score saturation or low sensitivity at extreme values.

\subsubsection{Difficulty Tiers}
As shown in Figure~\ref{fig:diff_factor}, our difficulty factor distribution is an approximately normal (Gaussian) but right-skewed distribution. Based on the distribution, we define three difficulty tiers using the mean $\mu=2.8$ and standard deviation $\sigma=1.0$. Tasks with difficulty\_factor $\leq1.8$ are considered \textbf{Easy}, those between 1.8 and 3.8 are \textbf{Medium}, and those $\geq3.8$ are \textbf{Hard}, following the rule of $\mu\pm\sigma$. 

Here is a simple experiment that we sample 50 data points and evaluate a baseline model (GPT-4o) in Creativity task with these difficulty tiers. The model achieves a score of 4.35 on Easy tasks, 2.95 on Medium tasks and 0.10 on Hard tasks. The result indicates that for easy tasks, agent can achieve a good score, while for hard tasks (e.g., to build an extremely complex house), it cannot complete correct execution and obtain a very low score.

\subsection{Discussion on Evaluation and Output Format}
\label{discuss_eva}
\subsubsection{Discussion on Evaluation and Output Format}
We ultimately select the output format of a 3D blueprint matrix as the data structure to represent architectures. 
Why didn't we choose the mainstream output formats, such as executable code? We considered that code generation is actually quite complex for architectures building, and this representation method cannot express the metadata form of buildings at the underlying level.

Although code generation aligns well with the motivation of disassembling sub-structures during the building construction process, for some complex buildings, this representation method only repeats the reasoning of agents, thus limiting the planning effect of agents. For example, when constructing a complex house with a roof, the code generation output by agents may be more inclined to express the coordinate values of each position in the matrix (e.g., 
$matrix[1][2][3] = Block\_A$).
For some complex structures, they are more likely to make fundamental errors in the relationships between 3D coordinate points.

We expect agents to analyze and plan building outputs in the form of ``Layers'', i.e., a 3D building is reduced in dimension to several 2D plane data through layering. For these 2D plane data, they can construct the building based on the characteristics of model data and the matrix as the root. If we also use code generation to express levels, we can find that this is consistent with the form of a blueprint. The blueprint method can ensure that agents better understand the relationships between 3D coordinate points, and then combine spatial reasoning at the abstract level to better reflect the spatial intelligence of agents.

The following is the reason why we do not present Spatial Reasoning and Spatial Commonsense tasks in planning forms.

Our MineAnyBuild datasets mainly include three tasks, i.e., Executable Spatial Plan Generation, Spatial Understanding, and Creativity to comprehensively evaluate the agents'  spatial planning ability in the aspects like instruction following or abstract architecture  understanding. Beyond these tasks, we additionally supplement the VQA tasks of Spatial Reasoning and Spatial Commonsense for the intention of assisting the evaluation for the agents' capabilities in commonsense-based spatial reasoning, which significantly impacts its spatial planning accuracy. Defining these two tasks with the VQA form is convenient for capability evaluation, which is also demonstrated in other previous benchmarks.

\subsubsection{Details of Output Format}

\textbf{(1) Output Format of Blueprint Matrix:}

The blueprint matrix has three dimensions. The three dimensions of this matrix are height (y-axis), depth (z-axis) and width (x-axis), respectively. The size of this matrix is determined by the sizes of the three dimensions.

The elements of the blueprint matrix are integers, each of which corresponds to a block. We use integers instead of strings as a representation of a block, to avoid an explosion in the number of tokens caused by a large number of elements. Therefore, we adopt a format similar to sparse matrices and use integers to represent elements of the matrix.

\textbf{(2) Output Post-processing for Architecture Generation:}

We use the list "block\_materials" to represent the types of blocks that will be used in the building task. When querying the MLLM-based agents, we convert this list into a hash (a dictionary in Python) to indicate the mapping relationship between block types and integers. The “air” block is set to -1 by default. The integers corresponding to the block types are calculated by adding 1 to the index of this block type in the list. 
For example, for the list [”grass”, “oak\_wood”], the hash should be {”air”: -1, “grass”: 1, “oak\_wood”: 2}. We also use the reversed hash during the execution process to establish a reverse mapping for further evaluation and visualization.

\section{Prompts}
\label{sec_E}
We provide task prompts, evaluation prompts, and data curation prompts in Section~\ref{task_prompts}, Section~\ref{eva_prompts}, and Section~\ref{Data Curation Prompts}, respectively.

\subsection{Task prompts}
\label{task_prompts}

\noindent\textbf{(1) Executable Spatial Plan Generation:}

\begin{promptbox}
\textbf{\#\# System Prompt:}

You are an expert Minecraft builder in a flat Minecraft Java 1.20.4 server and Python coding. Your goal is to produce a Minecraft architecture, considering aspects such as architecture structure, block variety, symmetry and asymmetry, overall aesthetics, and most importantly, adherence to the platonic ideal of the requested creation.\\

\textbf{\#\# User Prompt:}\\
Build the architecture based on the instruction and reference image. The instruction divides the architecture in the reference image by structure and demands. Please analyze and plan how to build the corresponding sub-structures according to the divided structure and demands, and give the ONLY ONE OVERALL blueprint matrix of the total architecture. The blueprint is a three-dimension list, and the dimensions of the blueprint matrix are in the order of height(positive y), length(positive z) and width(positive x). Fill in "-1" into the blueprint to denote as "air". The elements of the list matrix are integers filled according to the given mapping table.\\

Here is an example.\\
Block materials: \{"oak\_planks": 1\}\\
Instruction: build a 3*3*4 (width, length, height) wooden house. We want it to be a simple boxy house. The roof and floor should be solid and there is some space that player can go inside the house.

Output: \\
Planning: The floor and roof of this wooden house can be made of 3*3 oak\_planks as a square. Make the house hollow with air in the layer 2 and 3 and leave the space for entrance towards west (negative x). The two-layer walls are also made of 7*2=14 oak\_planks. I can build it layer by layer so that I can truly understand the spatial structure of this house. Then, here is the overall blueprint matrix of the whole architeture. I'm sure that the 3-dim list has correct template and fill in -1 as "air" to simulate no block here.\\
Blueprint: 
\begin{lstlisting}
```json [[[1,1,1],[1,1,1],[1,1,1]],[[1,-1,1],[1,-1,1],[1,1,1]],[[1,-1,1],[1,-1,1],[1,1,1]],[[1,1,1],[1,1,1],[1,1,1]]] '''
\end{lstlisting}

IMPORTANT: You must only use blocks from the given block materials dictionary. Make full use of them.\\
IMPORTANT: You must output ONLY ONE BLUEPRINT following the example format (A 3-DIM MATRIX).\\
IMPORTANT: Fill in -1(interger) as "air" into the blueprint matrix if no block is placed in the corresponding position.\\

\textbf{\#\# Input:}\\
Now, take a breath and continue.\\
Block materials: $\{block\_types\}$\\
Instruction: $\{instruction\}$\\
Reference image: <image>\\

\textbf{\#\# Output:}\\
Output:
\end{promptbox}

\noindent\textbf{(2) Spatial Understanding:}
\begin{promptbox}
\textbf{\#\# System Prompt:}

You are an expert Minecraft builder in a flat Minecraft Java 1.20.4 server and Python coding. Your goal is to produce a Minecraft architecture, considering aspects such as architecture structure, block variety, symmetry and asymmetry, overall aesthetics, and most importantly, adherence to the platonic ideal of the requested creation.\\

\textbf{\#\# User Prompt:}\\
Build the architecture based on the instruction and reference image. The instruction provides the relative coordinates of every block in reference image, and please identify the structure and summarize it as an overall blueprint matrix. The blueprint is a three-dimension list, and the dimensions of the blueprint matrix are in the order of height(positive y), length(positive z) and width(positive x). Fill in "-1" into the blueprint to denote as "air". The elements of the list matrix are integers filled according to the given mapping table.\\

Here is an example.\\
Block materials: \{"oak\_planks": 1\}\\
Instruction: build a 3*3*4 (width, length, height) wooden house layer by layer from bottom to top. Layer 1: oak\_planks: [(0,0), (0,1), (0,2), (1,0), (1,1), (1,2), (2,0), (2,1), (2,2)]. Layer 2: oak\_planks: [(0,0), (0,2), (1,0), (1,2), (2,0), (2,1), (2,2)]. Layer 3: oak\_planks: [(0,0), (0,2), (1,0), (1,2), (2,0), (2,1), (2,2)]. Layer 4: oak\_planks: [(0,0), (0,1), (0,2), (1,0), (1,1), (1,2), (2,0), (2,1), (2,2)].

Output: 
\begin{lstlisting}
```json
[[[1,1,1],[1,1,1],[1,1,1]],[[1,-1,1],[1,-1,1],[1,1,1]],[[1,-1,1],[1,-1,1],[1,1,1]],[[1,1,1],[1,1,1],[1,1,1]]]'''
\end{lstlisting}

IMPORTANT: You must only use blocks from the given block materials dictionary. Make full use of them.\\
IMPORTANT: You MUST output ONLY ONE BLUEPRINT following the example format (A 3-DIM MATRIX). You MUST NOT answer your reasons.\\
IMPORTANT: Your results should only follow the format of the example and not be influenced by the content of the examples.\\
IMPORTANT: Fill in -1(interger) as "air" into the blueprint matrix if no block is placed in the corresponding position.\\

\textbf{\#\# Input:}\\
Now, take a breath and continue.\\
Block materials: $\{block\_types\}$\\
Instruction: $\{instruction\}$\\
Reference image: <image>\\

\textbf{\#\# Output:}\\
Output:
\end{promptbox}

\noindent\textbf{(3) Creativity:}
\begin{promptbox}
\textbf{\#\# System Prompt:}

You are an expert Minecraft builder in a flat Minecraft Java 1.20.4 server and Python coding. Your goal is to produce a Minecraft architecture, considering aspects such as architecture structure, block variety, symmetry and asymmetry, overall aesthetics, and most importantly, adherence to the platonic ideal of the requested creation.\\

\textbf{\#\# User Prompt:}\\
Build the architecture based on the instruction and let your imagination run wild and use your creativity to build your best architecture. Before providing your final blueprint matrix, plan your solution considering the following spatial planning perspectives:\\
1. How to best interpret and implement the build specification.\\
- List key elements from the build specification\\
- Brainstorm block combinations for different parts of the structure\\
- Outline a rough structure layout\\
2. Creative ways to use the available blocks to achieve the desired aesthetic.\\
3. How to ensure the mapping correctness in your blueprint matrix.\\
4. Ways to maximize creativity and the dynamic range of possible builds.\\
5. Consider potential challenges and solutions\\
The blueprint is a three-dimension list, and the dimensions of the blueprint matrix are in the order of height(positive y), length(positive z) and width(positive x). Fill in "-1" into the blueprint to denote as "air". The elements of the list matrix are integers filled according to the given mapping table.\\

Here is an example.\\
Block materials = ["oak\_planks", "cobblestone", "red\_wool"]\\
Instruction: build a 3*3*4 (width, length, height) wooden house. \\
Output:\\
Planning Reasons: Let's build a simple wooden house. I use cobblestone as the material for the floor and oak\_planks for the wall and roof. \\
Selected\_block\_materials = {"oak\_planks": 1, "cobblestone": 2}\\
Blueprint:
\begin{lstlisting}
```json
[[[2,2,2],[2,2,2],[2,2,2]],[[1,-1,1],[1,-1,1],[1,1,1]],[[1,-1,1],[1,-1,1],[1,1,1]],[[1,1,1],[1,1,1],[1,1,1]]]'''
\end{lstlisting}

IMPORTANT: You must output ONLY ONE BLUEPRINT following the example format (A 3-DIM MATRIX).\\
IMPORTANT: Fill in -1(interger) as "air" into the blueprint matrix if no block is placed in the corresponding position.\\

\textbf{\#\# Input:}\\
Now, take a breath and continue.\\
Block materials: $\{block\_types\}$\\
Instruction: $\{instruction\}$\\

\textbf{\#\# Output:}\\
Output:
\end{promptbox}

\noindent\textbf{(4) Spatial Reasoning:}
\begin{promptbox}
\textbf{\#\# System Prompt:}\\
You are an expert Minecraft builder and player in a flat Minecraft Java 1.20.4 server.\\

\textbf{\#\# User Prompt and Input:}\\
You need to answer this question with a visual image.\\
Question: $\{instruction\}$ <image>\\
You must output ONLY one option (from True,False) without any reason based on the question. IMPORTANT: You can only answer one word (from A,B,C,D or True,False).\\

\textbf{\#\# Output:}\\
Your answer:
\end{promptbox}

\noindent\textbf{(5) Spatial Commonsense:}
\begin{promptbox}
\textbf{\#\# System Prompt:}\\
You are an expert in Minecraft and interior design, familiar with real-life common sense.\\

\textbf{\#\# User Prompt and Input:}\\
You will receive a question and an image. Answer the question based on the image, focusing on spatial commonsense. Your response must not exceed 70 words. Do not include any additional content or thoughts. Now, take a breath and continue.\\
Instruction: $\{instruction\}$\\
The next image is 
$\{img\_desp\}$. <image>\\

\textbf{\#\# Output:}\\
Your answer:
\end{promptbox}

\subsection{Evaluation Prompts}
\label{eva_prompts}

\noindent\textbf{(1) Executable Spatial Plan Generation:}

\begin{promptbox2}
\textbf{\#\# System Prompt:}\\
You are an expert Minecraft builder in a flat Minecraft Java 1.20.4 server and an expert architecture critic.\\

\textbf{\#\# User Prompt:}\\
Give a grade from 1 to 10 to the following Minecraft architectures from different views. The scores of reference human-annotated architectures are all 8 by default, as a reference for comparison. You should give the grade based on how well they are presented and correspond together to the building instructions in the following aspects:\\
- Completeness(Instruction Following): from *nothing, abandoned*(1) to *partial, incomplete*(5) and *masterfully completed, perfectly realized*(10).\\
- Complexity: from *simplistic, basic*(1) to *straightforward, moderate *(5) and *challenging, hardcore*(10).\\
- Overall Aesthetic, Atmosphere and Fidelity: from *stark, bare*(1) to *appealing, unusual*(5) and *epic, masterpiece*(10).\\

\textbf{\#\# Input:}\\
The following image is the ground-truth human-annotated reference image.\\
<image>\\
Give the grades based on the following image showing the Minecraft architecture in JSON format.\\
<image>\\
Building instructions: $\{instruction\}$\\
You must ONLY return the results in the following JSON format, but do not refer to the grades in this example and just follow the FORMAT:
\begin{lstlisting}
{
"Completeness(Instruction Following)": {
    "grade": 8,
    "comment": "The architecture well follows the building instructions and achieves a good finish, which is similar to the reference architecture."
    },
"Complexity": {
    "grade": 8,
    "comment": "This architecture has several advanced building techniques like using several stairs upside down and half slabs to present some structures with half a block."
    },
"Overall Aesthetic, Atmosphere and Fidelity": {
    "grade": 8,
    "comment": "The selection and placement of blocks have a certain aesthetic sense, reveal the feeling of ancient, in line with the requirements of the given instructions."
    }
}
\end{lstlisting}
You must not output any other reasoning or analysis.\\

\textbf{\#\# Output:}\\
Output:
\end{promptbox2}

\noindent\textbf{(2) Spatial Understanding:}
\begin{promptbox2}
\textbf{\#\# System Prompt:}\\
You are an expert Minecraft builder in a flat Minecraft Java 1.20.4 server and an expert architecture critic.\\

\textbf{\#\# User Prompt:}\\
Give a grade from 1 to 10 to the following Minecraft architectures from different views. The scores of reference human-annotated architectures are all 10 by default, as a reference for comparison. You should give the grade based on how well they are presented and correspond together to the building instructions in the following aspects:\\
- Instruction Following(Completeness): from *nothing, abandoned*(1) to *partial, incomplete*(5) and *masterfully completed, perfectly realized*(10).\\

\textbf{\#\# Input:}\\
The following image is the ground-truth human-annotated reference image.\\
<image>\\
Give the grades based on the following image showing the Minecraft architecture in JSON format.\\
<image>\\
Building instructions: $\{instruction\}$\\
You must ONLY return the results in the following JSON format, but do not refer to the grades in this example and just follow the FORMAT:
\begin{lstlisting}
{
"Completeness(Instruction Following)": {
    "grade": 8,
    "comment": "The architecture well follows the building instructions and achieves a good finish, which is similar to the reference architecture."
    }
}
\end{lstlisting}
You must not output any other reasoning or analysis.\\

\textbf{\#\# Output:}\\
Output:

\end{promptbox2}

\noindent\textbf{(3) Creativity:}
\begin{promptbox2}
\textbf{\#\# System Prompt:}\\
You are an expert Minecraft builder in a flat Minecraft Java 1.20.4 server and an expert architecture critic.\\

\textbf{\#\# User Prompt:}\\
Give a grade from 1 to 10 to the following Minecraft architectures from different views. You should give the grade based on how well they are presented and correspond together to the building instructions in the following aspects:\\
- Creativity: from *boring, dull*(1) to *mediocre, normal*(5) and *blue sky thinking, inspiring*(10).\\
- Completeness: from *nothing, abandoned*(1) to *partial, incomplete*(5) and *masterfully completed, perfectly realized*(10).\\
- Complexity: from *simplistic, basic*(1) to *straightforward, moderate *(5) and *challenging, hardcore*(10).\\
- Architecture Structure: from *boxy, rudimentary*(1) to *intuitive, modest*(5) and *sophisticated, intricate*(10).\\
- Overall Aesthetic, Atmosphere and Fidelity: from *stark, bare*(1) to *appealing, unusual*(5) and *epic, masterpiece*(10).\\

\textbf{\#\# Input:}\\
Give the grades based on the following image showing the Minecraft architecture in JSON format.\\
<image>\\
Building instructions: $\{instruction\}$\\
You must ONLY return the results in the following JSON format, but do not refer to the grades in this example and just follow the FORMAT: 
\begin{lstlisting}
{
"Creativity": {
    "grade": 6,
    "comment": "The building uses the same material but different forms of block types to enrich the architectural design. The design of this building is based on reality but beyond reality."
    },
"Completeness": {
    "grade": 5,
    "comment": "The architecture well follows the building instructions and achieves a good finish. The architecture is not broken and is built completely."
    },
"Complexity": {
    "grade": 6,
    "comment": "This architecture has several advanced building techniques like using several stairs upside down and half slabs to present some structures with half a block."
    },
"Architecture Structure": {
    "grade": 6, 
    "comment": "The design of the whole building is based on the vertical line as the axis and symmetrical in the center, which has the sense of extending upward."
    },
"Overall Aesthetic, Atmosphere and Fidelity": {
    "grade": 5,
    "comment": "The selection and placement of blocks have a certain aesthetic sense, reveal the feeling of ancient, in line with the requirements of the given instructions."
    }
}   
\end{lstlisting}
You must not output any other reasoning or analysis.\\

\textbf{\#\# Output:}\\
Output:
\end{promptbox2}

\noindent\textbf{(4) Spatial Commonsense:}
\begin{promptbox2}
\textbf{\#\# System Prompt:}\\
You are an expert in the field of multi-modal large language models and answer proofreading. You can well compare the differences between the output results of large models and the standard answers and score them.\\

\textbf{\#\# User Prompt:}\\
You will get the output result of a multi-modal large language model and a standard answer. You need to compare the two and score the output result of the MLLM. What you need to note is: \\
1)Evaluate the matching degree between the output result of the large model and the standard answer. It is not necessary for the contents of the two to be completely the same, but the tendency of the answers must be the same to be considered a correct match.\\
2)You need to score the matching degree, with a full score of 10. If it is a correct match, please score at least 8 points or more. If it is a wrong match, please score at least 3 points or less. For example, if the output of the large model is yes, and the standard answer is no, then it is obviously a wrong match.\\
3)You need to carefully check the key information in the standard answer, such as spatial position and direction, action tendency, and spatial common sense reasoning. If the output of the large model meets all of them, please add points; if there are any that are not met, please deduct points as appropriate within the range.

Please output the score and scoring reason (within 70 words) following this JSON format:\\
{"score": 5, "reason": ""}\\

\textbf{\#\# Input:}\\
Standard answer: $\{answer\}$\\
MLLM response: $\{response\}$\\

\textbf{\#\# Output:}\\
Your result in JSON format:
\end{promptbox2}

\subsection{Data Curation Prompts}
\label{Data Curation Prompts}

\begin{promptbox}
\textbf{\#\# System Prompt:}\\
You are an expert in Minecraft architecture, proficient in spatial planning and architectural design.\\

\textbf{\#\# User Prompt:}\\
Ensure that generated content is related solely to construction, excluding any irrelevant content related to user interaction.\\
Describe how to build this object in under 150 words.\\
Based on the input JSON file, generate a textual description of the structure within the file. \\
The output must exclude any coordinates and adhere to natural language norms without structured bullet points or numbered lists.\\
Examples in the required format:\\
{\it Let's build a Ancient Pillar Ruin structure from bottom to top.
First, place 5 stone\_bricks as the shape of a cross twice to build a two-layer foundation.
Then, place 1 stone\_bricks in the center, and place each stone\_brick\_slab on the last 4 stone\_bricks.
Place 2 cobblestone\_wall vertically reaching a height of two blocks, and then 2 chiseled\_stone\_bricks similarly on top of the cobblestone\_wall.
Leaning against the top chiseled\_stone\_bricks, place 4 stone\_brick\_stairs upside down towards different orientation.
Finally, place 1 stone\_bricks in the center, and place each stone\_brick\_slab on the 4 stone\_brick\_stairs.}\\

\textbf{\#\# Input and Output:}\\
\end{promptbox}

\section{Additional Experimental Information, Results and Analyses}
\label{sec_F}

In this section, we first provide additional experimental information such as compute resources and evaluation metrics. Then, we present more quantitative results with corresponding analyses. 


\subsection{Agents}
Besides MLLM-based agents, reinforcement learning (RL)-based agents are also the mainstream research objects in many current related works on Minecraft environments. RL-based agents have a variety of low-level atomic actions and can be further developed in combination with reinforcement learning algorithms. We are planning to develop RL-based agents for addressing our proposed benchmark under popular RL frameworks such as Mineflayer~\cite{mineflayer}, MineDojo~\cite{fan2022minedojo}, and MineRL~\cite{guss2019minerl}. We will synchronize and update these RL-based agents to our Github code repository at \url{https://github.com/MineAnyBuild/MineAnyBuild} in the future.

\subsection{Compute Resources and Cost}
\label{compu_res}
We mainly use MLLM-based agents for the evaluation of our benchmark MineAnyBuild.

For proprietary models, we implement the input/output (I/O) of the open-world AI agents by querying the corresponding APIs of MLLMs. We spent approximately 350 US dollars on the API calls in our current experiments.

For open-source models, we conduct the inference of these models on NVIDIA RTX 3090 GPUs and NVIDIA RTX 4090 GPUs.
Noticing that the average output length of the planning tasks is much longer than that of other tasks, more GPU memories are required to support our benchmark evaluation.
Considering the limited resources, we did not use GPUs with more memory to run open-source models with larger parameters. In the future, we will supplement the evaluation results of large-scale open-source models based on the actual situation.


\subsection{Evaluation Metrics}

\subsubsection{Scores}

\noindent\textbf{1) Evaluation Scores:}\\
During our experiments, we find that if we grade a single task from only one dimension, the scores of the critic model are not entirely reliable, making it hard to distinguish the fundamental differences among some outputs. Therefore, we consider introducing multiple scoring dimensions for evaluation. Inspired by Chain-of-Thought~\cite{wei2022chain}, the multi-dimensional scoring can prompt the critic model to provide more accurate scores. 

We present the composition of evaluation scores and the corresponding scoring criteria (from 1 to 10) in the evaluation prompts in Section~\ref{eva_prompts}.
The evaluation scores are separated and designed to evaluate for each task.

\textbf{(a) Creativity.}
We mainly have the following five scoring dimensions for evaluation: Creativity ($S_{\textrm{Creativity}}$), Completeness ($S_{\textrm{Completeness}}$), Complexity ($S_{\textrm{Complexity}}$), Architecture Structure ($S_{\textrm{AS}}$), Overall Aesthetic, Atmosphere and Fidelity ($S_{\textrm{OAAF}}$).

\begin{equation}
S_{\textrm{Evaluation}} = k_1*S_{\textrm{Creativity}} + k_2*S_{\textrm{Completeness}} + k_3*S_{\textrm{Complexity}} + k_4* S_{\textrm{AS}} + k_5* S_{\textrm{OAAF}}
\end{equation}
where we set $k_1=0.8, k_2=k_3=k_4=k_5=0.05$ to better emphasize the evaluation of creativity.

\textbf{(b) Executable Spatial Plan Generation.}
We mainly have the following three scoring dimensions for evaluation: Completeness(Instruction Following) ($S_{\textrm{CIF}}$), Complexity ($S_{\textrm{Complexity}}$), Overall Aesthetic, Atmosphere and Fidelity ($S_{\textrm{OAAF}}$).

\begin{equation}
S_{\textrm{Evaluation}} = k_1*S_{\textrm{CIF}} + k_2* S_{\textrm{Complexity}} + k_3* S_{\textrm{OAAF}}
\end{equation}
where we set $k_1=0.4, k_2=k_3=0.3$ to balance various dimensions so as to comprehensively evaluate the effectiveness of spatial planning.

\textbf{(c) Spatial Understanding.}
We only have the following scoring dimensions for evaluation: Completeness(Instruction Following) ($S_{\textrm{CIF}}$).

\begin{equation}
S_{\textrm{Evaluation}} = S_{\textrm{CIF}}
\end{equation}

For the evaluations scores above, we assign different importances to each dimension of a specific task based on the following principle: We first select a main scoring dimension based on human preference priors and assign it a relatively high importance. For the remaining dimensions, we evenly distribute the remaining weight proportion to ensure the output buildings have basic structures without obvious errors. 
The assignment of importance can achieve a relatively good balance between the main evaluation dimension of creativity task and other basic scoring dimensions.

\noindent\textbf{2) Matching Scores:}\\
For the tasks which correspond to a specific ground-truth blueprint of the reference images, we calculate the ``Matching Score'' between the results and ground-truths.
We only conduct it on the Spatial Understanding and Executable Spatial Plan Generation tasks. The specific calculation formula for this score is as follows:
\begin{equation}
S_{\textrm{Matching}} = \frac{M}{N}*10
\end{equation}
where $N$ is the block amount of the ground-truth data, and $M$ is the amount of matching blocks in the corresponding 3D bounding space filtering out ``air'' block. Only when the block types match in the relative position can it be considered valid.

We set the weighting coefficient of this score to a relatively small value between 0.05 and 0.1 because we found that most of the outputs could not perfectly reproduce the buildings compared to the original reference images.
We believe that the construction of architectures does not necessarily have to be completely identical to be truly good. Therefore, we have adjusted this weighted value downward.

\noindent\textbf{3) Voting Scores:}\\
For Creativity task, we design a voting ranking scoring algorithm as 
``Voting Score''
referencing the Swiss-round algorithm~\cite{csato2017ranking}.
The specific design of the algorithm is shown in Algorithm~\ref{rank_algorithm}.
We set the weighting coefficient of this score to a relatively small value $\sim$ 0.05.  
We will ultimately convert the rankings from $L$ in Algorithm~\ref{rank_algorithm} into $S_{\textrm{Voting}}$ ranging from 3 to 8 as the final voting score.

\begin{algorithm*}[t]
\caption{Multi-round Agent Battle Ranking Algorithm} \label{rank_algorithm}
\begin{algorithmic}[1]
\REQUIRE (1) Agent Ids $I={id_1,...,id_n}$: The set of agent identifiers to be ranked;  (2) Total Rounds $R$: The total number of battle rounds.
\ENSURE (1) Ranked List $L$: The final ranked list containing (id, score); (2) Compared Log $H$: The complete battle record.
\FOR{each $id_{i}\in I$}
\STATE Create a tuple of agents $a_{i} \gets \{id:id_{i}, score:0, compared:\emptyset\}$
\ENDFOR
\FOR{round $r \gets 1$ \TO $R$}
\STATE Generate battle combinations: 
\STATE Sort the agent set by $(-score, id)$, and then create an empty battle set $P \gets \emptyset$ and a used set $U \gets \emptyset$
\FOR{$i \gets 1$ \TO $n$}
    \IF{$a_i.id \in U$}
    \STATE \textbf{continue}
    \ENDIF
    \FOR{$j \gets i+1$ \TO $n$}
        \IF{$a_j.id \in U$} \STATE \textbf{continue}
        \ENDIF
        \IF{$a_j.id \notin a_i.compared$}
            \STATE $P \gets P \cup \{(a_i, a_j)\}$
            \STATE $U \gets U \cup \{a_i.id, a_j.id\}$
            \STATE \textbf{break}
        \ENDIF
    \ENDFOR
\ENDFOR

\IF{$P = \emptyset$} 
    \STATE \textbf{break}
\ENDIF

\FOR{each pair $(a_a, a_b) \in P$}
    \STATE Get Critic Model's response $choice \in \{1,2\}$, Determine the winner $w$ and the loser $l$:
    \IF{$choice = 1$}
        \STATE $w \gets a_a$, $l \gets a_b$
    \ELSE
        \STATE $w \gets a_b$, $l \gets a_a$
    \ENDIF
    
    \STATE Update scores: $w.score \gets w.score + 1$, $l.score \gets l.score - 1$
    \STATE Record the battle: $w.compared \gets w.compared \cup \{l.id\}$, $l.compared \gets l.compared \cup \{w.id\}$
\ENDFOR

\ENDFOR
\STATE $L \gets$ The agent set sorted by the compare function
\RETURN $(L, H)$
\end{algorithmic}
\end{algorithm*}

\begin{equation}
    S_{\textrm{Voting}} = 3 + 5 \times \frac{\text{score} - \min(\text{score})}{\max(\text{score}) - \min(\text{score})}
\end{equation}

We will also decide whether to use this score based on the specific scale of the evaluation.


\subsubsection{Formula for Comprehensive Score}

For different tasks, we obtain a comprehensive score based on these three kinds of scores, denoted as \textbf{``Score'' (out of 10)} shown in Table \redbox{1} of the main text, through corresponding weighting to indicate the performance of agents in each task. The specific calculation formulas for each task are as follows:

\begin{align}
S_{\textrm{CR}} &= k_1*S_{\textrm{Evaluation}} + k_2*S_{\textrm{Voting}}\\
S_{\textrm{SP}} &= k_3*S_{\textrm{Evaluation}} + k_4*S_{\textrm{Matching}}\\
S_{\textrm{SU}} &= k_5*S_{\textrm{Evaluation}} + k_6*S_{\textrm{Matching}}
\end{align}
where $S_{\textrm{CR}}$, $S_{\textrm{SP}}$ and $S_{\textrm{SU}}$ are the comprehensive scores of Creativity, Executable Spatial Plan Generation, Spatial Understanding tasks, respectively. We set $k_2=k_4=k_6=0.05$ with relatively low weights, and $k_1=1-k_2, k_3=1-k_4, k_5=1-k_6$ to emphasize the evaluation score.

For most cases, we only consider Evaluation Scores as the Comprehensive Score to simplify the evaluation.
We will also dynamically adjust the proportions of these scores in the future and determine an optimal weighting coefficient based on experiments and manual supervision.

\subsubsection{Explanations on ``Overall'' Score}
The ``Overall'' Score in Table 1 of the main text is the average score of these five tasks, with a maximum value of 100. The calculation formula is:
\begin{align}
S_{\textrm{Overall}}=\frac{(S_{\textrm{ESPG}}+S_{\textrm{SU}}+Acc_{\textrm{SR}}/10+S_{\textrm{C}}+S_{\textrm{SC}})}{5}*10
\end{align}
where $S_{\textrm{ESPG}}$, $S_{\textrm{SU}}$, $Acc_{\textrm{SR}}$, $S_{\textrm{C}}$ and $S_{\textrm{SC}}$ are the Score/Accuracy of Executable Spatial Plan Generation, Spatial Understanding, Spatial Reasoning, Creativity and Spatial Commonsense tasks.

\subsubsection{Error Bar Chart of Scoring by Critic Models}
\begin{figure*}[h!]
\vspace{-0.2cm}
\centering
\includegraphics[width=\linewidth]{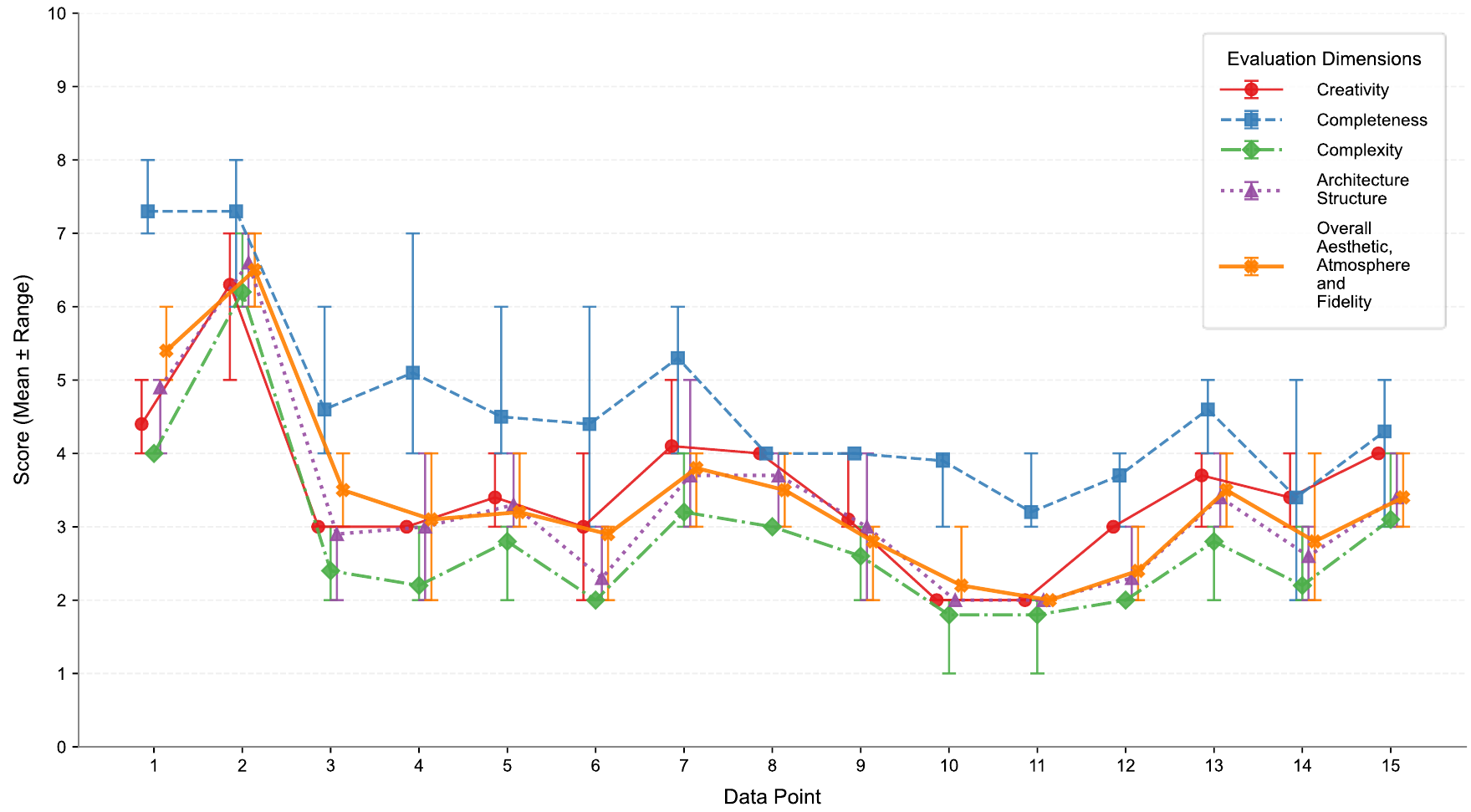}
\caption{Error bar chart for the scoring of critic models.}
\label{fig:errorbar}
\vspace{-0.2cm}
\end{figure*}

We provide an error bar chart for the scoring of critic models, as shown in Figure~\ref{fig:errorbar}. Specifically, we set 15 data points for the Creativity tasks, with different tasks for each data point. We conduct 10 API calls, using GPT-4.1 as critic model, with consistent inputs for each data point to obtain different outputs. 
For each score of the 10 outputs, we calculate the average, maximum, and minimum values to determine the upper and lower bounds of the error for that score. We set the upper bound of the error as the maximum value and the lower bound of the error as the minimum value, and then the values of the data point correspond to the average scores.

We can observe that for most data points, the upper and lower error of the scores is concentrated within 1. The scores given by the critic model are relatively consistent, which also lays a foundation for the reliability of our benchmark.

\subsection{Assessments for Reliability Evaluation}

We provide some assessments to present the reliability and reasonability of the critic model in our benchmark.

\textbf{(1) Spearman Correlation:}\\
To evaluate the reliability of our MLLM-based critic model, we randomly sample 50 task outputs from our benchmark. We rate these outputs from 1 to 10 just like the critic model does. We compute the Spearman correlation between human and model scores, obtaining $\rho=0.76 (p<0.01)$. The value shows that the reliability and reasonability of the critic model.

\textbf{(2) Intra-model Consistency:}\\
As shown in Figure~\ref{fig:errorbar}, we can observe that for most data points, the upper and lower error of the scores is concentrated within 1. The scores given by the critic model are relatively consistent, which also lays a foundation for the reliability of our benchmark.

\subsection{Full Results of Output Success Rate}
We provide the detailed data corresponding to Figure~\redbox{4} in the main text, as shown in Table~\ref{tab:osr}.

\begin{table}[h]
	\caption{Detailed data of Output Success Rate (OSR) in Figure~\redbox{4} of the main text.}	
	\label{tab:osr}
	\centering
			\resizebox{0.9\linewidth}{!}{
	{\renewcommand{\arraystretch}{1.5}\begin{tabular}{l||ccc}
			 \specialrule{.1em}{.05em}{.05em}
		 
    Models&\cellcolor{lightcornflowerblue!40}Spatial Understanding&\cellcolor{lightgreen!40}Executable Spatial Plan Generation&\cellcolor{lightcornflowerblue!40}Creativity
    
    \cr\hline
    \rowcolor{lightgoldenrodyellow}\multicolumn{4}{l}{\textbf{\textit{Proprietary}}}\\
    Claude-3.5-Sonnet&97.41&98.91&100.00\\
    Claude-3.7-Sonnet&94.59&96.36&98.70\\
    Gemini-1.5-Flash&73.81&93.25&98.27\\
    Gemini-1.5-Pro&94.48&99.56&93.66\\
    Gemini-2.0-Flash&91.23&86.37&94.48\\
    GPT-4o&88.09&92.44&92.03\\
    GPT-4o-mini&95.73&95.10&86.94\\
    \hline\rowcolor{lightgoldenrodyellow}\multicolumn{4}{l}{\textbf{\textit{Open-source}}}\\
    InternVL2.5-2B&34.95&27.60&28.18\\
    InternVL2.5-4B&37.90&32.71&42.18\\
    InternVL2.5-8B&74.33&68.82&66.54\\
    Qwen2.5VL-3B&50.88&50.84&44.65\\
    Qwen2.5VL-7B&76.46&77.63&79.42\\
    LLava-Onevision-7B&42.87&52.87&60.38\\
    \hline\rowcolor{lightgoldenrodyellow}\multicolumn{4}{l}{\textbf{\textit{Abandoned}}}\\
    InternVL2.5-1B&16.61&21.20&22.19\\
  LLava-Onevision-0.5B&13.21&14.76&15.23\\
        
          \specialrule{.1em}{.05em}{.05em}
		\end{tabular}}}
\end{table}

\subsection{Accuracy Evaluation Results on Spatial Reasoning Task}
We present the results of more MLLM-based agents on Spatial Reasoning task, as shown in Table~\ref{tab:sr_others}.
We can observe that even some state-of-the-art (SOTA) MLLMs, e.g., GPT-4.1 and Gemini-2.5-Pro, do not perform optimally on our spatial reasoning tasks. Although the accuracy of these SOTA models on these tasks is not high, the accuracy on human evaluation (where task difficulty is 
simpler than ours) is only 60-90\%~\cite{bersier2025cognitive}, which also indicates the high difficulty of the task itself. This result also reveals that MLLM-based agents are required to be trained on tasks related to spatial intelligence to further improve their intelligence level in the future.

\begin{table}[h]
	 \fontsize{10}{10}\selectfont
    \caption{Accuracy evaluation results of more MLLM-based agents on the Spatial Reasoning task.}
	\label{tab:sr_others}
	\centering
			\resizebox{0.30\linewidth}{!}{
	{\renewcommand{\arraystretch}{1.5}\begin{tabular}{l||c}
			 \specialrule{.1em}{.05em}{.05em}
		 
    Models&\cellcolor{lightgreen!40}Accuracy $\uparrow$
    
    \cr\hline
    \rowcolor{lightgoldenrodyellow}\multicolumn{2}{l}{\textbf{\textit{Proprietary}}}\\
    Gemini-2.5-Pro&24.02\\
    GPT-4.1&20.25\\
    GPT-4.1-mini&22.57\\
    \hline\rowcolor{lightgoldenrodyellow}\multicolumn{2}{l}{\textbf{\textit{Open-source}}}\\
    InternVL2.5-26B&26.16\\

          \specialrule{.1em}{.05em}{.05em}
		\end{tabular}}}
 \vspace{-0.4cm}
\end{table}

\subsection{Discussion on Human Evaluation}
We conducted a simple experiment to examine the differences between human evaluation and the evaluation by critic models.
We asked 5 volunteers to score 10 or 30 architectures (with 13 results for each architecture) via a questionnaire.
We collect the following findings in the process of organizing the questionnaire results:\\
1) Human evaluation is prone to inertia and fatigue during multiple evaluations, which can affect judgment. Especially when we changed the number of architectures to be evaluated from 10 to 30, most volunteers were fatigued by the questionnaire's length,
which can easily lead to evaluation bias.

Human evaluation is intuitively more convincing than critic models. 
However, considering the large-scale evaluation needs in the future, such a heavy workload is more suitable to be handled by the highly intelligent critic model.

2) Human evaluation usually requires a reference. People usually select one result from multiple similar ones as a reference and score others with minor variations based on this reference.

3) Humans are not very accurate in scoring, and each person has different standards. But when the quantity and scale are sufficient, human evaluation remains highly reliable.

\subsection{Discussion on Possible Noise Brought by Critic Model}
For analysis of the possible noise, we find that there are mainly two probable aspects. 

\textbf{(1) Different Scoring Comments:}
We require the critic model to output the comments while giving the evaluation scores. As shown in the evaluation prompts in Section E.2 of Supplementary Material, we manually provide the output examples and the comments generated by critic model are possible sources of noise. 

\textbf{(2) Manually Customized Scoring Tiers:}
As shown in our evaluation prompts, we provided some words for ratings of 1, 5 and 10 respectively, as scoring references for the critic model. The manually-written prompts in this part may introduce some possible noise. 

But overall, the impact of these two possible sources of noise on our evaluation is marginal, which can be supported by the provided error bar chart.

\subsection{Discussion on Abnormal Result on GPT-4o-mini in Spatial Reasoning Task}
The evaluation results in Spatial Reasoning task of other models (e.g., GPT-4o) are normal, while the output results of GPT-4o-mini are abnormally uniform. Due to the smaller parameters and weaker capabilities of GPT-4o-mini, most of the outputs are the frequent answers (e.g., B/C/False with 765/640/288 times), and A/D/True (with 10/24/0 times) are rarely seen. It shows that GPT-4o-mini is more likely to “guess” a relatively higher accuracy in this test but the absolute score is not high, though our data is evenly and randomly distributed across all options. In contrast, stronger models may attempt to reason more deeply, which may not always get higher scores on difficult questions.

\subsection{Information and Metrics about Quality Control Processes in Data Curation Pipeline}
We reanalyze our data generated before and provide the lacking information and metrics including rejection rates and inter-annotator agreement about quality control processes for Executable Spatial Plan Generation task as follows. 

\textbf{(1) For rejection rates:} 
We randomly sample 100 instructions from this task and observe a rejection rate of 11\%, suggesting that most machine-generated instructions are broadly acceptable with minimal invalid cases. In most of the rejected cases, information of the instructions is either too redundant or contains some extraneous information. Thanks to the powerful instruction generation ability of GPT-4.1, the quality of most generated instructions is already quite high. Our manual processing is just to ensure that they meet our expectations.

\textbf{(2) For inter-annotator agreement (IAA):}
We randomly select 50 instructions from this task and ask two annotators to judge whether each instruction is acceptable/rejected. We calculate Cohen’s Kappa to evaluate IAA and obtain a score of $\kappa=0.63$, indicating a strong level of agreement and the reasonably high quality of machine-generated task instructions.

\subsection{Discussion on Other Baseline Evaluation}
Our MineAnyBuild mainly focuses on evaluating MLLM-based agents, which is highly representative for the evaluation of AI agents. The baselines including human experts, domain-specific and reinforcement learning-based agents are well worthy of being discussed and researched in the future. Due to several engineering challenges of code implementation and tight time constraints, we provide some transferring insights below for future research: 

\textbf{(1) For human experts:}
We plan to recruit $\sim$10 human participants with experience in Minecraft building by referencing current work~\cite{jiang2025behavior}, to perform the same planning tasks to build the architectures. We will utilize the same evaluator and evaluation criteria to assess human experts and agents, so as to provide more comprehensive evaluation results. 

\textbf{(2) For domain-specific agents:}
Domain-specific agents, such as Voyager~\cite{wang2023voyager}, could potentially perform well by leveraging task-specific knowledge or predefined strategies on Minecraft tech-tree tasks. The input/output of these agents cannot be directly transferred to our benchmark, therefore we need to do some adaptation works for further evaluation. Constructing some in-domain agents for our building data leaves some future works to better improve the intelligence of AI agents.

\textbf{(3) For reinforcement learning-based agents:}
In Section F.1 and the GitHub repository, we have stated that we plan to develop RL-based agents under popular RL frameworks (e.g., Mineflayer~\cite{mineflayer}, MineDojo~\cite{fan2022minedojo}, and MineRL~\cite{guss2019minerl}) in the future. These agents will adopt an architecture similar to Vision-Language-Action (VLA) models. These VLA models will focus on how to perform action modeling on vision and language inputs and embeddings, which will bring certain help to the research of AI agents other than language modeling. We believe that this research will also be helpful for the research in Embodied AI in the future.

\subsection{Performance of Small-parameter Open-source MLLMs}
For some models with small parameters (e.g., 0.5B and 1B), we found that their performance was particularly poor and errors often occurred. Therefore, we finally decided not to include these results in the main text. For the sake of fairness, we placed these results in Table~\ref{abandoned_table}.

\begin{table*}[h!]
 \centering

\caption{Evaluation results of small-parameter open-source MLLMs on \textbf{MineAnyBuild}.}
	\label{abandoned_table}
	\resizebox{\linewidth}{!}{
	{\renewcommand{\arraystretch}{1.3}
		\begin{tabular}{l||ccccc|c}			\specialrule{.1em}{.05em}{.05em}
    \multirow{2}{*}{Models}&\cellcolor{lightcornflowerblue!40}Executable Spatial Plan Generation&\cellcolor{lightgreen!40}Spatial Understanding&\cellcolor{lightcornflowerblue!40}Spatial Reasoning&\cellcolor{lightgreen!40}Creativity&\cellcolor{lightcornflowerblue!40}Spatial Commonsense
    &\multirow{2}{*}{Overall}

    \cr\cline{2-6}
    &Score $\uparrow$&Score $\uparrow$&Accuracy $\uparrow$&Score $\uparrow$&Score $\uparrow$&\cr\hline

\rowcolor{lightgoldenrodyellow}\multicolumn{7}{l}{\textbf{\textit{Open-source}}}\\
        InternVL2.5-1B&0.11&0.08&28.5&0.12&4.18&14.68\\

LLava-Onevision-0.5B&0.09&0.00&27.8&0.37&3.26&13.00\\
 \specialrule{.1em}{.05em}{.05em}

			\end{tabular}}}
\end{table*}

Regarding the performance of these models on the Spatial Reasoning task, we find that in most cases, their outputs are the same option (e.g., ``A'' or ``False''), and only in a few cases they output other options. This results in a probability for them to achieve an accuracy slightly higher than 25\%, which is not worthy of reference when compared with other models.

\section{Additional Visualization Results}
\label{sec_G}

In this section, we provide additional visualization results with analyses about the executable planning output and failure cases. 


\subsection{Executable Planning Output}

\begin{figure*}[h!]
\vspace{-0.2cm}
\centering
\includegraphics[width=\linewidth]{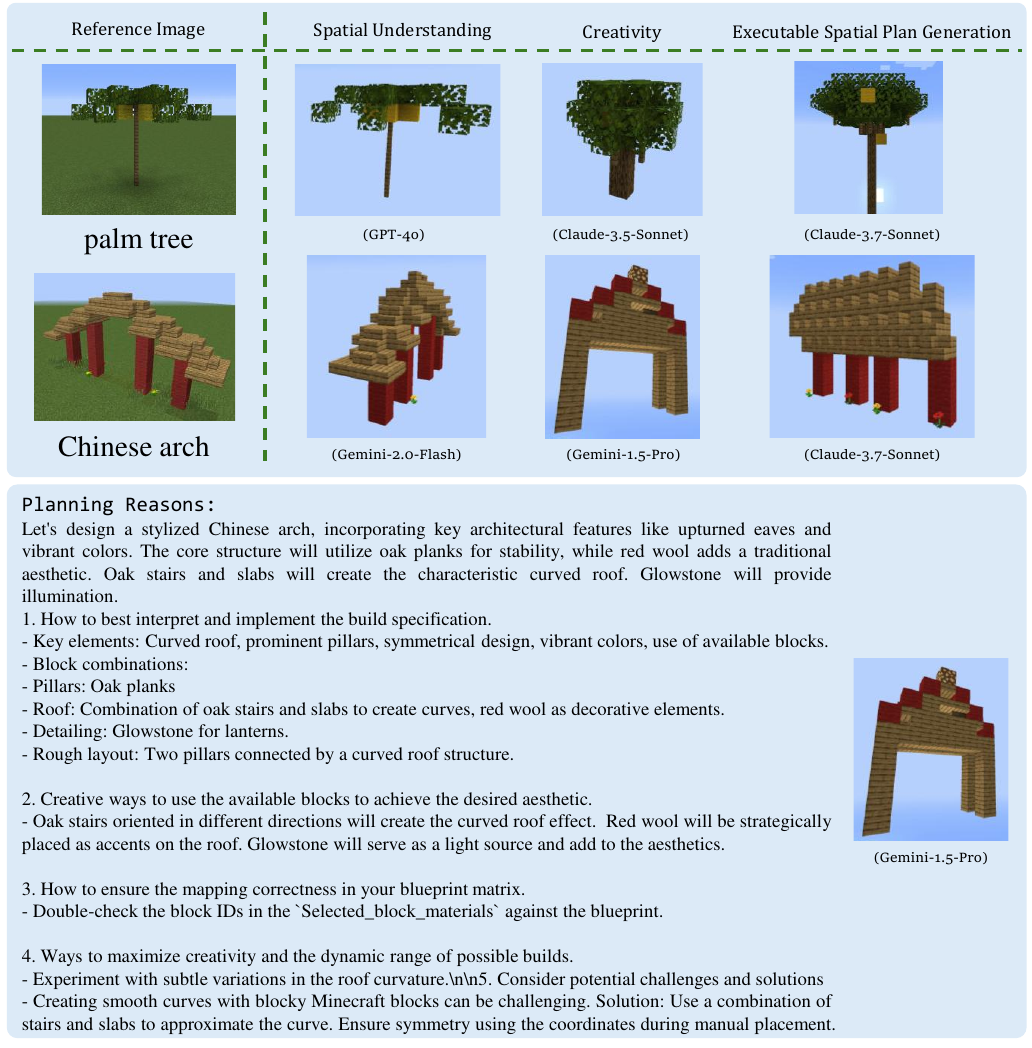}
\caption{More visualization results on executable planning output.}
\label{fig:visual_success}
\end{figure*}

We provide more visualization results on executable planning output in Figure~\ref{fig:visual_success}.
We can see that proprietary models perform well in spatial planning tasks of complex architectures. For example, for \textit{palm tree} and \textit{Chinese arch} in Figure~\ref{fig:visual_success}, both GPT-4o and Gemini-2.5-Flash can well convert relative coordinates to world coordinates in Spatial Understanding task. For Creativity task, the agents exhibit capabilities to manifest architectural structures without ground-truth visual supervision. As shown in the Figure~\ref{fig:visual_success} below, the agent proactively utilize the block materials to construct the desired architecture by spatial planning and unleashing creativity.

\subsection{Failure Cases}

\begin{figure*}[h!]
\vspace{-0.2cm}
\centering
\includegraphics[width=0.92\linewidth]{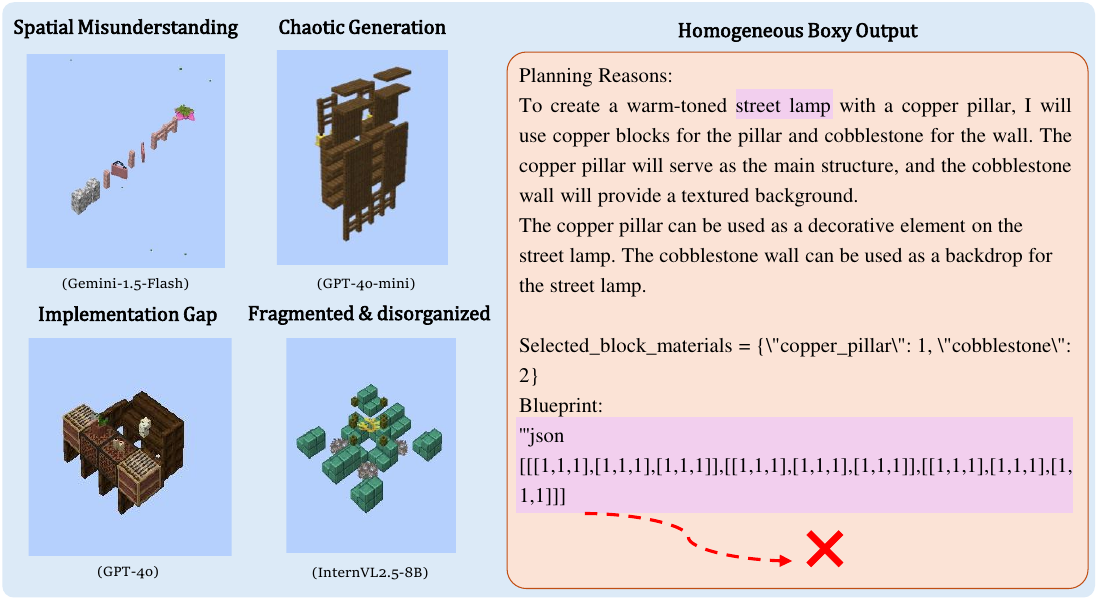}
\caption{More visualization results on failure cases.}
\label{fig:visual_fail}
\vspace{-0.2cm}
\end{figure*}

We provide more visualization results on failure cases in Figure~\ref{fig:visual_fail}, revealing that there are still some challenges for agents, e.g., spatial misunderstanding, chaotic/fragmented/disorganized generation, and homogeneous boxy structure output. 
We can observe that, when the block amount exceeds a certain threshold or facing non-cubic architectural structures, agents struggle to understand precise space localization and relationships among blocks, particularly in sub-structure composition and design.
Furthermore, agents present limited capability to interpret relationships between block orientation (e.g., {\it stairs} as a part of the house roof) and global structural patterns.
It is worth emphasizing that the fundamental issue lies in agents' abilities to ``translate'' textual planning into executable architectural outputs (i.e., blueprint matrix), namely implementation gap. 
These identified limitations provide future directions for spatial intelligence research.

Moreover, we provide the deep analyses of the failure reasons for the cases in the main text and above with a summarization as follows: 

\textbf{(1) Spatial Misunderstanding:}
Agents frequently misinterpret 3D positional relationships or fail to maintain the correct spatial arrangements, which highlights a persistent weakness in spatial grounding and planning.

\textbf{(2) Implementation Gap:}
The agents have a central issue that they can not transform high-level textual plans into precise and executable blueprint matrices. The integration of substructures often fails due to incorrect block indexing, orientation errors or inconsistent spatial logic, leading to blueprint parsing or execution failures. This is essentially because the model's understanding of data structures such as codes or DSL and 3D matrices from a numerical perspective is still limited. If these models strengthen the training of spatial data, it may enhance the capabilities of these agents.

\textbf{(3) Structural Degeneration under Complexity:}
When the tasks demand non-cubic, asymmetric or creative designs, the agents tend to collapse into simple and box-like outputs or disorganized results. This indicates that their limited ability to scale from basic patterns to more abstract and complex architectural concepts.

These failure modes reflect deeper limitations in MLLM's capabilities to perform hierarchical spatial planning, maintain geometric consistency and ground language into manipulable 3D structures. They also provide more research directions for MLLMs, e.g., to improve multi-modal spatial understanding, align linguistic abstraction with executable plans or enhance agent’s ability for structural composition in open-ended 3D environments.

\section{Declaration of LLM Usage}
\label{sec_H}

We primarily use LLMs for data filtering and the scoring system in our critic model (Multi-modal Large Language Models).
Our implementation involves no ethical breaches, with all outputs conducting human review to ensure safety and compliance. The usage of LLMs in our work does not impact the scientific rigorousness and originality of the research.

\end{document}